\documentclass[Afour,sageh,times]{sagej}

\usepackage{moreverb,url}
\usepackage[colorlinks,bookmarksopen,bookmarksnumbered,citecolor=red,urlcolor=red]{hyperref}
\usepackage{algorithm}
\usepackage{algpseudocode}
\usepackage{color,soul}
\usepackage[caption=false,font=footnotesize]{subfig}
\newcommand\BibTeX{{\rmfamily B\kern-.05em \textsc{i\kern-.025em b}\kern-.08em
T\kern-.1667em\lower.7ex\hbox{E}\kern-.125emX}}

\def\volumeyear{2018}
\begin{document}

\runninghead{Le Goff, Mukhtar, Coninx and Doncieux}

\title{Bootstrapping Robotic Ecological Perception from a Limited Set of Hypotheses Through Interactive Perception}

\author{L\'eni K. Le Goff\affilnum{1}, Ghanim Mukhtar\affilnum{1}, Alexandre Coninx\affilnum{1} and St\'ephane Doncieux\affilnum{1}}
\affiliation{\affilnum{1}Sorbonne Universit\'e, CNRS, Institut des Syst\`emes Intelligents et de Robotique, ISIR, F-75005 Paris, France}



\begin{abstract}
To solve its task, a robot needs to have the ability to interpret its perceptions. In the case of vision, this interpretation is particularly difficult and relies on the understanding of the structure of the scene, at least to the extent of its task and sensorimotor abilities. A robot with the ability to build and adapt this interpretation process according to its own tasks and capabilities would push away the limits of what robots can achieve in a non controlled environment. A solution is to provide the robot with processes to build such representations that are not specific to an environment or a situation.
A lot of works focus on objects segmentation, recognition and manipulation.
Defining an object solely on the basis of its visual appearance is challenging given the wide range of possible objects and environments. Therefore, current works make simplifying assumptions about the structure of a scene. An example of such an assumption is the tabletop hypothesis, which posits that objects lay on a flat surface. Such assumptions reduce the adaptivity of the object extraction process to the environments in which the assumption holds. To limit such assumptions, we introduce an exploration method aimed at identifying moveable elements in a scene without considering the concept of object. By using the interactive perception framework, we aim at bootstrapping the acquisition process of a representation of the environment with a minimum of context specific assumptions. The robotic system builds a perceptual map called relevance map which indicates the moveable parts of the current scene. A classifier is trained online to predict the category of each region (moveable or non-moveable). It is also used to select a region with which to interact, with the goal of minimizing the uncertainty of the classification. A specific classifier is introduced to fit these needs: the collaborative mixture models classifier. The method is tested on a set of scenarios of increasing complexity, using both simulations and a PR2 robot.
\end{abstract}


\maketitle

\section{Introduction}\label{sec:intro}

\begin{figure}[h]
\centering
\includegraphics[height=60mm,width=\linewidth,keepaspectratio]{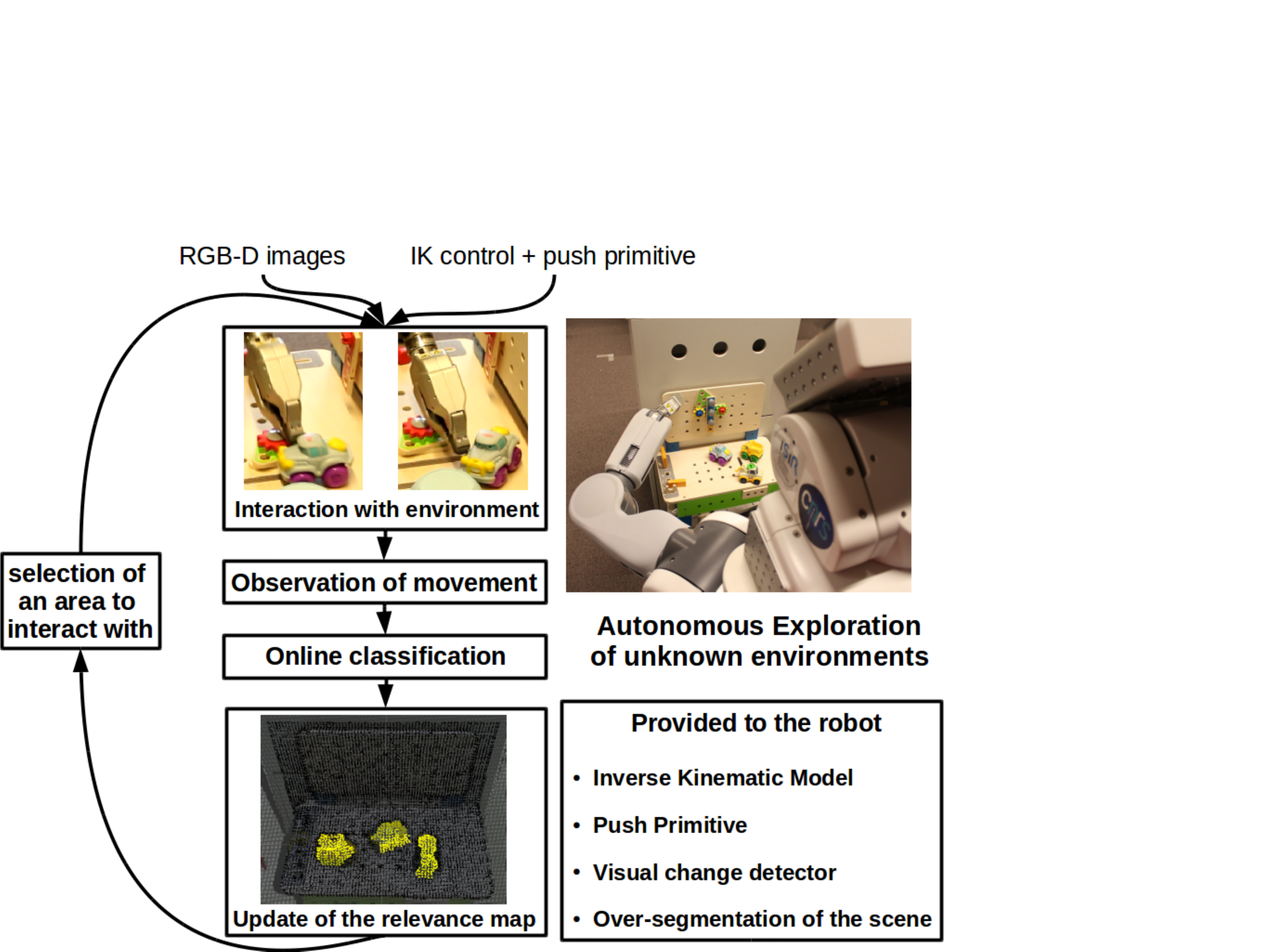}
\caption{Overview of the proposed babbling approach. The robot chooses a point in the environment with which to interact and then observes whether an object has moved as a result of its action. On the basis of such interactions, a classifier is trained online using the information gathered. Finally, a relevance map is computed to guide exploration.}
\label{fig:schema_gen}
\end{figure}

Beyond a preprogrammed scenario, building robots that are able to act and fulfill a mission in uncontrolled and unstructured environments remains a challenge. Even everyday environments, which tend to be highly structured, are hard to deal with given their variability. To achieve tasks in such environments, a robotic system needs a robust and adaptive perception of the world. The focus of this work is to bootstrap a world's representation learning process for a robot, i.e. a robotic ecological perception.   


Vision is a rich modality that carries a dense set of information reflecting the complexity of realistic environments. To understand a visual scene, a robot must first be able to focus on important components of its visual field according to its embodiment, skills, and current goal. To select an appropriate action, it must simplify this sensor flow. Certain hypotheses can be formulated related to the structure of an environment, e.g. the tabletop hypothesis, or to the shapes of objects; however, these hypotheses limit the ability of the robot to adapt to new environments. A way to avoid making such assumptions is to allow the robot to explore its surrounding via a direct interaction. In this way, by observing the effect of its action on the environment, the robot can gain novel sensory signals and learn from regularities in the sensorimotor space. This domain of research is known as interactive perception : action to enhance perception, and perception to enhance interaction \citep{bohg2017interactive}. 

Most previous studies on interactive perception have object segmentation, recognition and manipulation as goals \citep{bohg2017interactive}. To achieve these complex objectives, researchers use a passive image processing bootstrap step to produce object hypotheses. Then, through interactions with the environment, the robot confirms or rejects these hypotheses. Certain assumptions must be introduced to implement this preliminary step, e.g. objects are on a planar surface \citep{VanHoof2014,Gupta,Chang2012,Bersch2012,Metta2002,Fitzpatrick,Fitzpatrick2003,schiebener2011segmentation,hermans2012guided} or object shapes are close to predefined primitives (spheres, cubes, cylinders, etc.) \citep{Schiebener2014,kuzmivc2010object}. The present work aims at removing the need for these hypotheses. To this end, we propose to build a useful segmentation of a scene with an interactive perception approach.  

To deal with visual scenes complexity, it has been shown that humans do not consider all parts of the scene as equivalent. They possess a visual attention process that lowers energy consumption during the analysis of a scene \citep{carrasco2011visual} by focusing the attention on elements of the environment that are considered as salient \citep{Itti2001}. 

In computer vision, the study of visual saliency in human attention has led to salient object detection within an image \citep{Borji2014}. Works in this field aim at producing a saliency map of images, i.e. a binary map representing an accurate segmentation of an object in an image. 
With a saliency map, a robot could directly focus its attention on elements important to its task, for instance part of the environment that can move as a result of its own action, and thus decrease computational time. This represents a starting point for developing an autonomous capacity with which a robot can deal with realistic environments prior to having any representation of objects. 

The goal of this work is to define a method that associates the concepts of interactive perception and salient object detection \textit{to autonomously learn a perceptual map which represent moveable parts in an environment}. 
 The perceptual map is built on the basis of an autonomous exploration involving interactions of a robotic arm with the environment. It represents relevant areas according to the robot capabilities and to the task being undertaken. Saliency maps are focused on what is considered as salient for a human. A robot has different goals and sensorimotor capabilities and may thus not consider the same areas as relevant. To avoid any ambiguity, we thus term this map a \textit{relevance map}. We assume that \textit{something moveable by a robot's end-effector is potentially an object}. The robot learns to identify the relevant features of components that are moveable as a result of its actions. Relevant areas shown in the map could be one object, group of objects, part of an objects or an articulated object. By learning a simpler representation than object models for recognition or manipulation, our method requires less assumptions specific to the environment or to the objects. So, scenarios that are different from tabletop and object with complex shapes and textures are considered in this study.

The main contribution of this work is \textit{a segmentation process to identify relevant parts of the visual scene using the interactive perception paradigm with minimum a priori knowledge about the environment structure.}

This work is not focused on objects extraction for recognition or manipulation; rather, it constitutes a previous step, allowing an initial identification prior to a more targeted exploration. With minimum a priori knowledge on the environment, this work represents the very first step of a developmental process which could lead to a robust and adaptive extraction and identification of objects that are completely unknown to the robot designer.

\section{Related Work}\label{sec:rela_works}

\subsection{Saliency Map}\label{sec:SalM}

Saliency is directly linked to the study of human attention. A saliency map shows the distribution of salient components in the visual field, i.e. parts of the visual field that attract the gaze \citep{Itti2001}. It assesses which object in a picture will most attract the visual attention of a human \citep{Borji2014}.

Saliency maps can be built by different methods. The three main methods are salient object detection (SOD), fixation prediction (FP) and object proposal generation (OPG) \citep{Borji2014}. These methods have the same goal, i.e. \textit{detecting objects based on the study of human attention.} In other words, they aim to determine which object in a picture would most attract the visual attention of a human. The aim of SOD is to generate a saliency map that represents with high accuracy the most salient object in a picture. The saliency map is composed of regions in the picture that represent salient objects. The aim of FP is to produce a saliency map that represents possible fixation points for human attention; the corresponding saliency map is a set of points. Finally, the aim of OPG is to propose bounding boxes that might include an object. The result of OPG is not exactly a saliency map, but it shares the same properties.

In the following, the focus will be on SOD methods, which are the closest methods to the one introduced later herein.

SOD is used to detect the most salient object in an image and then produce a clean segmentation of the object boundaries. Most methods focus on detecting one salient object (the most salient), but some attempt to detect several objects \citep{Borji2014}. SOD methods have two components. First, they detect the salient parts in the image to yield a grayscale map, with a white color indicating the most salient parts. Second, they build an accurate segmentation of the object boundaries by applying a threshold and generating a binary map in which white areas represent the most salient object. 

SOD is divided into two main categories: approaches using heuristics and approaches using machine learning algorithms. In both cases, strong assumptions are made in relation to what object in  a picture will be the most attractive to a human:
\begin{itemize}
\item Center prior: salient objects are more likely to be in the center of the picture \citep{liu2014adaptive,jiang2013submodular,peng2013salient}
\item Background prior: the narrow borders of the image is part of the background \citep{liu2014adaptive,li2013saliency,jiang2013saliency}
\item Focusness prior: the camera often focuses on a salient object to attract attention; this can be defined as the degree of focus blur \citep{jiang2013salient}
\item Boundary connectivity prior: salient objects are less connected to the image borders \citep{zou2013segmentation,zhu2014saliency}
\item Color prior: certain colors seem to be more attractive to humans, e.g. salient objects are more likely to contain warm colors such as red or yellow to be in salient objects \citep{shen2012unified,liu2014adaptive,jiang2013submodular,peng2013salient}
\item Semantic prior: humans pay more attention to certain objects such as faces, cars, dogs, etc. \citep{shen2012unified}
\end{itemize}
The boundary connectivity, background, and center priors suggest that salient areas always exist around the center of the image. The focusness prior assumes that the image has been taken by a human who knows where to focus. These priors cannot be used to detect objects as this would imply that a robots knows where to center its camera and therefore knows where the object is located. The color and semantic priors are specific to humans and may not be relevant in certain situations.

Other priors are heuristics in relation to what an object may look like in a 2D image:
\begin{itemize}
\item Objectness prior defines a measure of "objectness" based on a provided definition of what an object is and then relies on this measure to compute saliency \citep{jia2013category,jiang2013salient}
\item Spatial distribution prior: if a color is widely distributed in an image, the salient object will likely not contain this color \citep{jiang2013submodular}
\end{itemize}
These priors can be useful, but they are not essential and can reduce the generality of the method. In the method proposed herein, most of these priors are replaced by the interaction of the robot with the environment.

In addition to these priors, saliency map methods decompose the visual scene in either blocks or regions. Blocks are rectangles on the image used to compute the visual features. Regions can have an arbitrary shape. They rely on superpixels, which are clusters of similar pixels based on color and contrast.

According to \citet{Borji2015}, the best performing methods have three features in common:
\begin{itemize}
\item Superpixels: contrary to block-based approaches, superpixels produce an accurate object boundary segmentation.
\item Background prior is used. This contrasts with the location prior, which assumes a specific location for a salient object in an image; usually, this is the center of the image. This assumption is strong and restricts the method to single object detection. Moreover, an autonomous robot with no concept of object will not be able to center the image around the area of interest.
\item Machine learning algorithms is used to train a model of saliency. Discriminative regional feature integration \citep{jiang2013discri} can be used to train a regression model based on a 93-dimensional features vector. This allows the method to be adaptable and scalable to more complex scenarios. 
\end{itemize}

These methods are used to build models of what humans would consider as salient. Furthermore, they are focused on static 2D pictures, rather than on the stream of images that a robot can collect while interacting with its environment. The focus here is not on building human-like saliency estimation. The question addressed is \textit{what are the most relevant areas of a real scene for an agent with given capabilities and with a certain goal ?} According to \citet{Borji2015}, a region-based approach (i.e. superpixel) with supervised learning is an efficient method for building a saliency map.  In this paper, we thus propose a new region-based method to detect relevant objects based on self-supervised learning. Relevance is a similar concept to saliency, but depends on task and robot features instead of human features.

\subsection{Object Segmentation by Interactive Perception} \label{sec:ip}

\begin{table*}[t!]
\begin{center}
\begin{tabular}{| l | l | l | l |}
\hline
Ref & Goal & Priors & Initial Segmentation \\
\hline
\citet{Kenney2009} & OS & RB, PM & - \\
\hline
\citet{Fitzpatrick2003} & OS & TS, RB, PM, OD & - \\
\hline
\citet{VanHoof2014} & OS & TS, RB, AP & pixel clustering, PS \\
\hline
\citet{ude2008making} & OS, OR & OH, HA & - \\
\hline
\citet{Gupta} & OS, OR & TS, RB, AP & color-based clustering, PS \\
\hline
\citet{Chang2012} & OS, OR & TS, RB, AP & pixel clustering, PS \\
\hline
\citet{hermans2012guided} & OS & TS, RB, AP & PS \\
\hline
\citet{Hausman2013} & OS, OR & TS, RB, AP & RANSAC (shape primitives) \\
\hline
\citet{Bersch2012} & OS, OR & TS, RB, AP & PS \\
\hline
\citet{kuzmivc2010object} & OS, OR & RB, AP, SP & SIFT, RANSAC \\
\hline
\citet{schiebener2011segmentation} & OS, OR & RB, AP, SP, TO & Harris Corner, RANSAC, PS  \\
\hline
\citet{Schiebener2014} & OS, OR & TS, RB, AP, SP & saliency map, difference of gaussian \\
\hline
\citet{Bergstrom2011} & OS & RB, PM, AP, TS & HSV Histograms, 3D Ellipsoids \\
\hline
\citet{Xu2014} & OS & RB, PM, AP, TS & Supervoxels \\
\hline
\citet{eitel2017learning} & OS & RB, PM, AP, TS & surface-based \\
\hline
\hline
Our Approach & Relevance Map & AP & Supervoxels \\
\hline
\end{tabular}
\end{center}
\caption{Summary of methods in Interactive Perception to segment objects. The goals are object segmentation (OS) and object recognition (OR). The priors are the following : tabletop scenario (TS), rigid body (RB), action primitives (AP), planar motion of the objects (PM), object database (OD), textured objects (TO), object in hand (OH); shape primitives (SP), and human assistance (HA). In object hypothesis generation, PS stands for plane segmentation} 
\label{tab:ip}
\end{table*}

Learning to perceive the world via interaction is known in robotics as interactive perception \citep{bohg2017interactive}. By interacting with its surrounding, a robot learns a representation of the environment through the relation between its capabilities and the effect of its actions.

Early works on this topic have been conducted by \citet{tsikos1988segmentation} in which they proposed a method to separate stacked and heaped objects thanks to interactions (like push, pick and shake) executed by a robotic system. This approach makes further image processing easier. Fifteen years later, interactive perception defined as above was studied by \citet{Metta2002,Fitzpatrick,Fitzpatrick2003}. In their works, a humanoid robot learns to segment a single object on a table and to recognize its arm. The robot interacts with its surrounding and uses optical flow extracted from 2D images to detect motions. By observing the motion as a consequence of its actions the system is able to segment the object from the background. In those studies, motions are restricted to planes and the experiments are tabletop scenarios in order to simplify the problem.
 
\begin{figure}[h]
\centering
\includegraphics[height=60mm,width=1\linewidth,keepaspectratio]{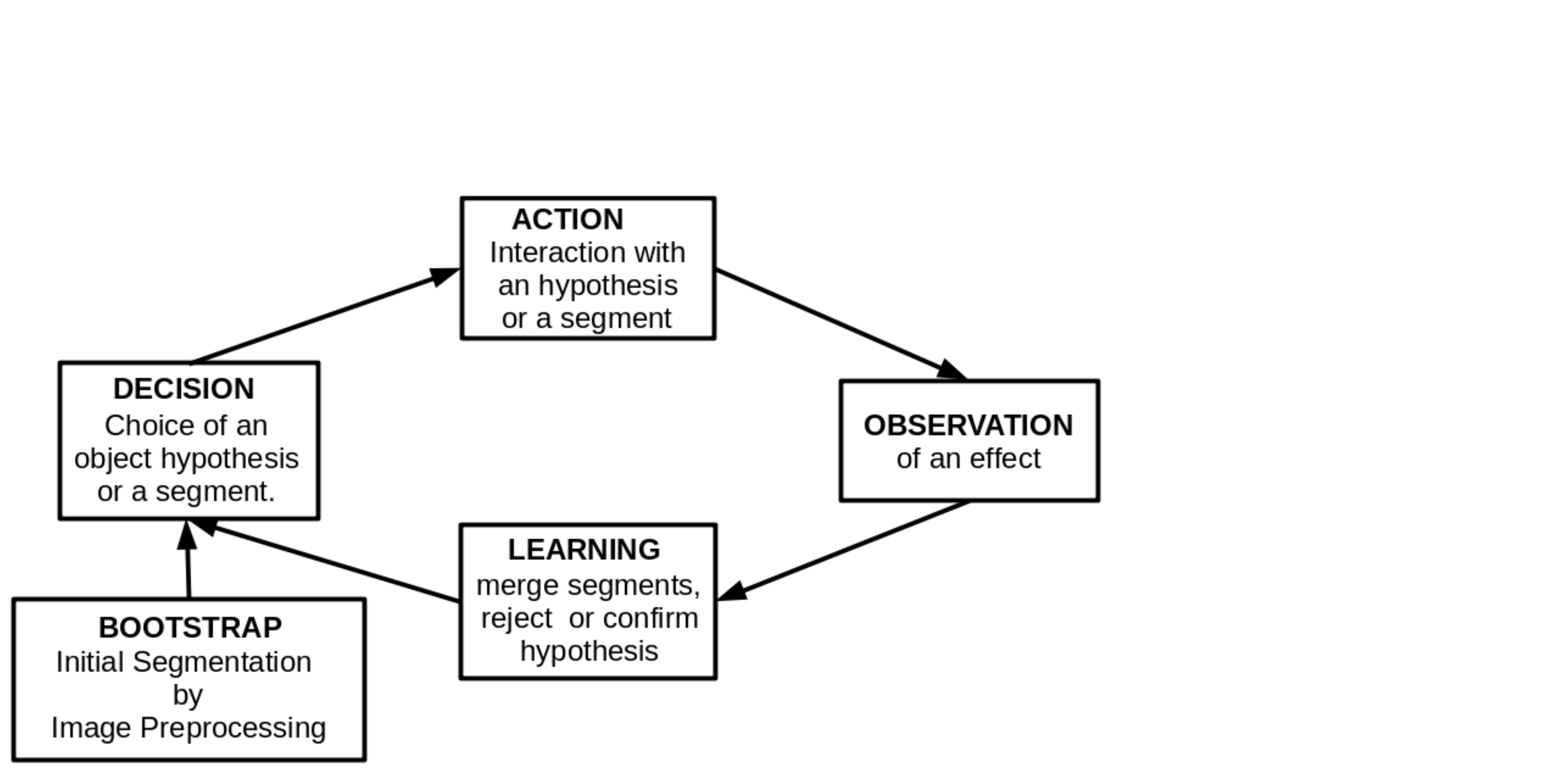}
\caption{General workflow of interactive perception methods.}
\label{fig:IP}
\end{figure}

Most works on interactive perception start with a passive image processing step in which a first segmentation is done. This segmentation could be an oversegmentation with segments that are smaller than the objects \citep{VanHoof2014,schiebener2011segmentation,patten2018action}. In this case, assumptions are used to maximize the probability to interact with segments that are part of objects. In other approaches, segments are object candidates \citep{Gupta,Chang2012,Bergstrom2011,hermans2012guided}. Object candidates are clusters of pixels in 2D images or clusters of points in pointclouds. The actions of a robot are then designed to reject or confirm these hypotheses. The complete interactive perception method generally follows the cycle depicted in figure \ref{fig:IP}, i.e. choice of a segment or an object candidate with which to interact, application of an action to chosen part of the environment, observation of the effect, and update of the perception by merging segments, confirming or rejecting of the chosen hypothesis. Some methods use the interaction to collect data and train a model with machine learning algorithms. 

The bootstrap step often uses passive computer vision methods and no interactions. This allows assumptions about objects and environments to bootstrap the system. Table \ref{tab:ip} summarizes the priors and techniques used for the bootstrap step. \citet{schiebener2011segmentation} used random sample consensus (RANSAC) to find planes and cylinders in a picture and generate object hypotheses. As they used 2D images, they needed highly textured objects to estimate 3D shapes. Other methods exploit depth information from 3D sensors, such as stereoscopic cameras or IR laser cameras, to detect planes \citep{VanHoof2014,Gupta,Chang2012, Bergstrom2011,hermans2012guided,eitel2017learning}. Indeed, the environments used in these studies were limited to tabletop scenarios to allow segmentation of the table from the objects on top of the table. It simplified the initial segmentation, which relied on a clustering method based on color. Their methods also aimed to separate objects via interaction. 

To handle textureless objects, an alternative approach is to make assumptions about the shapes of the objects. Candidate object generation then relies on comparisons of objects with primitive shapes, such as cylinders, planes, or spheres \citep{Bersch2012,Hausman2013,Schiebener2014}. \citet{Bersch2012} also segment the table, and \citet{kuzmivc2010object} consider only objects containing planes.

In another study \citep{ude2008making}, no assumptions about object type are made; the method does not rely on candidate object hypotheses. The robot starts with an unknown object in its hand and analyzes it in front of a complex, cluttered environment. This allows the robot to learn a model of the background and of the object. This learning exercise cannot be achieved autonomously by the robot as it needs to start with the unknown object in hand.

These approaches aim at both discovering and separating objects. They require a predefinition of what an object is and also require a dedicated candidate object generation method. Scenarios are typically restricted to tabletop scenarios, in which objects are on a flat surface, to easily distinguish objects from the background. This is a significant limitation to the range of environments that can be handled by the robot. An alternative is to make assumptions about objects, e.g. about their shapes or their textures, but it also requires an a priori definition of the possible shapes of all objects with which the robot may interact. Finally, to prevent any assumptions about the environment or the objects, a human teacher or helper can be involved; however, it also reduces the autonomy of the robot.

In the present study, the goal is to reduce the number of assumptions related to the objects and the environment in order to pave the way to more adaptive robot behaviors \citep{2016ACLN3729}. Provided that the perception system actually sees the objects, a single assumption is used: objects are parts of the environment that the robot can move. The robot uses interactive perception to learn to distinguish relevant objects from the background. An important feature of the proposed method is that \textit{the concept of object does not need to be defined}; it relies only on the concept of relevance.

\subsection{Interactive Perception to Build a Saliency Map}

Little work has focused on making a robot build a saliency map using an interactive perception process. Saliency map building is considered as a pure computer vision problem \citep{Borji2014,Borji2015}, and interactive perception aims to produce detailed representations of objects \citep{bohg2017interactive}. However, several studies make the association between these two fields, albeit indirectly. \citet{Craye2015} build a saliency map of an environment using a mobile robot. \citet{kim2015interactive} and \citet{Ugur2007} build a map that represents where in the environment a robot can apply certain actions. In these studies, the concepts of saliency or of relevance are not mentioned, but this map can be called a \textit{relevance} map as the robot focuses on regions with which it can interact. 

\citet{Craye2015} look at how to build a saliency map of their experimental rooms using mobile robots. The robots navigates the laboratory and builds a saliency map of objects. In that study, salient objects are defined as elements protruding from a flat surface. The robot could then build saliency maps of the objects on the floor (chairs, tables, etc.) or of objects on a table or desk. The saliency map is built using robot exploration, but this is not exactly interactive perception as no change is made in the environment as a result of the actions performed by the robot. Also, the assumption that object are on a flat surface limits the range of objects that can be recognized as salient.

\citet{Ugur2007} proposed a method for learning "traversability" affordance with a wheeled mobile robot which explores a simulated environment. The robot tries to go through different obstacles: laying down cylinders, upright cylinders, rectangular boxes, and spheres. The laying down cylinders and spheres are traversable while boxes and upright cylinders are not. The robot is equipped with a 3D sensor and collects data after each action labeled with the success of going through the objects. The sample data are extracted thanks to a simulated RGB-D camera. Then, an online SVM (\citet{bordes2005fast}) is trained based on the collected data. The resulting model predicts the "traversability" of objects based on local features. To drive the exploration, an uncertainty measure is computed based on the soft margin of the model decision hyperplane. Finally, they tested their method on a navigation problem, on real robots and in a realistic environment. They demonstrate, by using the model learned in simulation, that the robot was able to navigate through a room full of boxes, spherical objects and cylindrical objects like trash bins without colliding with non-traversable objects.

\citet{kim2015interactive} in the same idea seeks to learn pushable objects in a simulated environment using a PR2 with an RGB-D camera. The objects are blocks of the size of the robots. They are either pushable in one or two directions, or not pushable. The PR2 uses its two arms to try to push the blocks. The learning process relies on a logistic regression classifier and a Markov random field is used to smooth spatially the predictions. The robot explores then the environment and collects data by trying to push the blocks. The outcome of the framework is what they called an affordance map indicating the probability of pushability of a block. When in \citet{Ugur2007} the learning is made on continuous space, in \citet{kim2015interactive} the environment is discretized in a grid with the cells of the size of a block, thus, the learning space is discrete. Finally, they use an exploration strategy based on uncertainty reduction to select the next block to interact with.

%

Kim et al. and U\v{g}ur et al. study how, by interaction, a robot could build a \textit{relevance} map according to a task or capability. They do not mention relevance map explicitly but the affordance map of \citet{kim2015interactive} is close from our relevance map by the way they both segment interesting elements for the agent. Exploration and learning were conducted in simulation only and in simple environments. 
The present work aimed to consider more realistic environments in experiments performed with a real robot equipped with an arm. 

\section{Method}

\subsection{Overview}

The goal of our method is to produce a \textit{relevance map} through an autonomous exploration driven by a robotic arm. The robot explores an unknown, dynamic\endnote{In this work, "dynamic" means that the state of the environment is not reinitialized at the beginning of each iteration.}, environment. This exploration is driven by a relevance map of the environment, which is built online. Our approach could actually bootstrap most of the methods described in section  \ref{sec:ip}. We follow the main principles of interactive perception and SOD. The system first oversegments the scene into regions and then classifies them to generate a grayscale map representing the relevance of the regions. It then chooses an area to explore, interacts with it, observes the effects on the environment, and updates the classifier and the relevance map (see Figure \ref{fig:schema_gen}). The perception relies on a RGB-D camera (Microsoft Kinect 2\endnote{Other kind of 3D cameras could be used such as stereoscopic cameras}) to retrieve a 3D pointcloud of the scene. This camera is an active depth camera that has troubles to perceive dark and reflective surfaces \citep{lachat2015first}. This limitation comes from the perception device, not from the proposed method.

\begin{figure}[h]
\centering
\includegraphics[height=60mm,width=.9\linewidth,keepaspectratio]{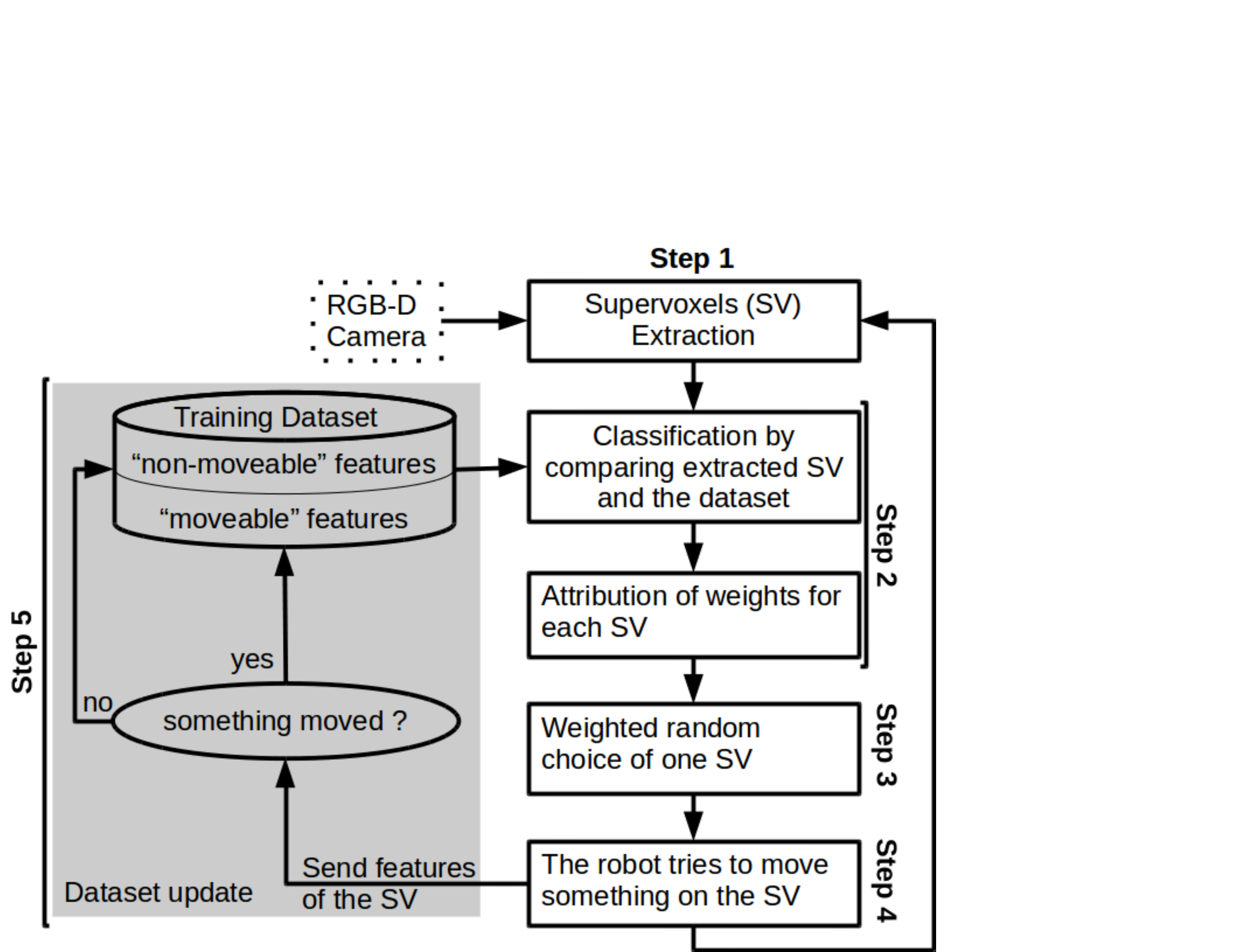}
\caption{General workflow of the approach.}
\label{fig:schema_exp}
\end{figure}

Figure \ref{fig:schema_exp} presents the general workflow of the method. The exploration is sequential with each iteration structured into 5 steps :
\begin{itemize}
\item[\textbf{Step 1}] Oversegmentation of the pointcloud into regions of the same size, called supervoxels. The oversegmentation is described in section \ref{sec:vccs}. Visual features are extracted to characterize each region (see section \ref{sec:feat}).
\item[\textbf{Step 2}] Computation of the \textit{relevance map} based on the oversegmentation and according to the prediction of the classifier. This step is described in section \ref{sec:sm}.
\item[\textbf{Step 3}] Choice of a supervoxel with which to interact. This choice is driven by a sampling process that relies on the relevance map (see section \ref{sec:choice}).
\item[\textbf{Step 4}] The robot interacts with the selected supervoxel with its push primitive (see section \ref{sec:contr}).
\item[\textbf{Step 5}] Observation of a possible effect. A basic change detection method is applied locally on the chosen supervoxel. The features of the supervoxel are stored in the database as samples. A label of 1 is used if a change is detected, and a label of 0 is used otherwise (see section \ref{sec:motdet}).
\end{itemize}

At the beginning of the exploration, all relevance scores are initialized to 0.5. Without any information about the environment, all supervoxels are assumed to be uncertain and must be explored.

\subsection{Supervoxels}\label{sec:vccs}

\begin{figure}[h]
\centering
\includegraphics[height=60mm,width=.5\linewidth,keepaspectratio]{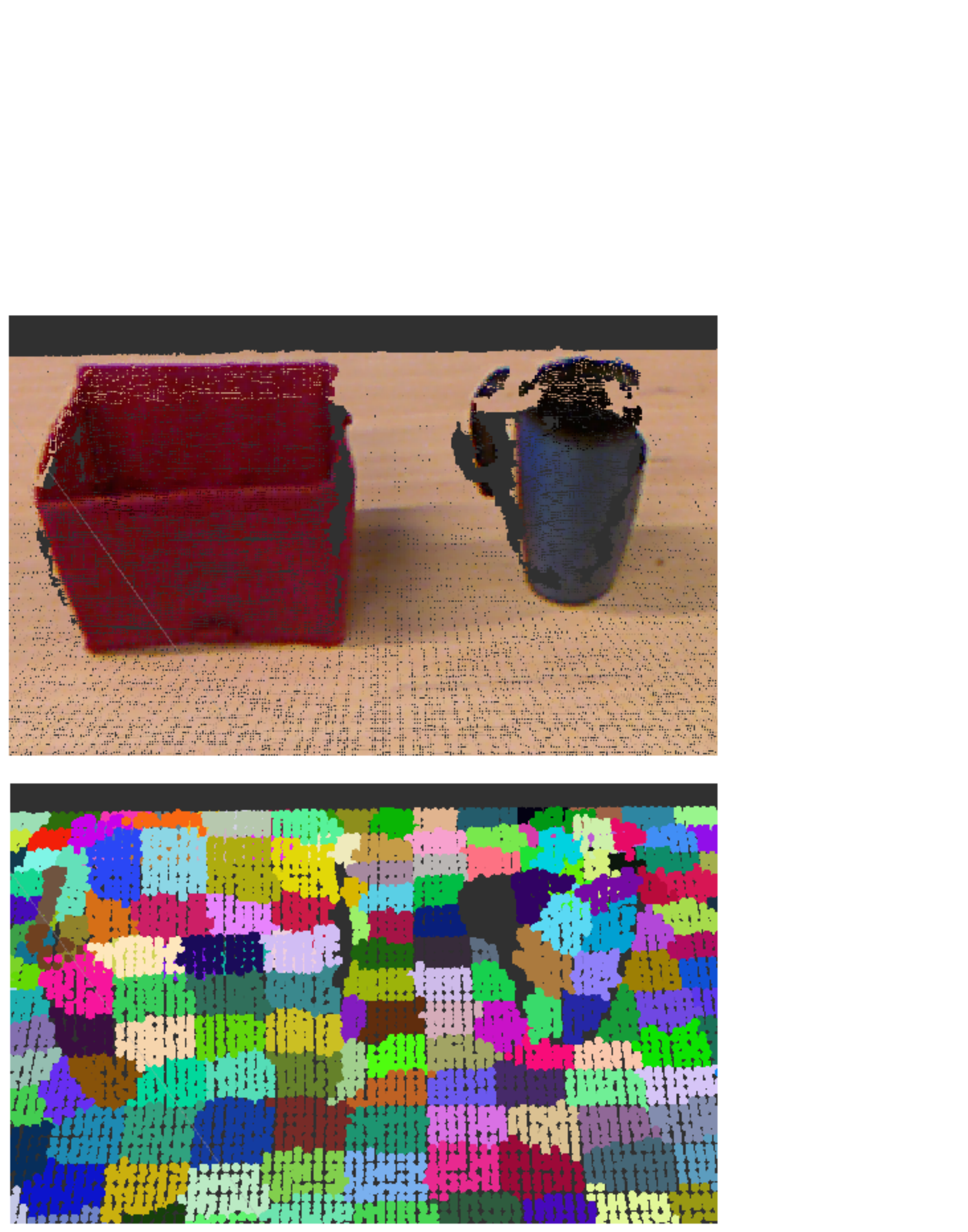}
\caption{Examples of supervoxels segmentation. These images represent pointclouds, the top one is a pointcloud extracted from a kinect, the bottom one is a pointcloud representing supervoxels extracted on the top pointcloud. In these pictures, the voxels are points as their are points of the pointclouds. 
}
\label{fig:supervox}
\end{figure}

The relevance map is based on an oversegmentation into supervoxels using voxel cloud connectivity segmentation (VCCS) \citep{Papon2013}. As illustrated in Figure \ref{fig:supervox}, supervoxels are clusters of voxels. Voxels are the smallest unit of a 3D image as pixels are for 2D images.

VCCS is similar to the superpixel methods such as SLIC superpixels \citep{Achanta2010} and turbopixel \citep{levinshtein2009turbopixels}. It builds clusters of voxels directly on 3D pointclouds. The use of depth information to build supervoxels is a significant enhancement compared with superpixels methods because this segmentation respects object boundaries. The samples stored to update the classification are thus more likely to be associated with a single object; thus, it removes a significant source of noise in the classification. Also, VCCS works on all kinds of environments because the algorithm uses low-level features such as color, normals, and geometric descriptors (fast point feature histograms (FPFH) proposed by \citet{rusu2009fast}). Therefore, VCCS produces a meaningful oversegmentation of RGB-D images. 

VCCS is based on a region growing algorithm. At the beginning, voxel seeds are evenly distributed on the pointcloud. Then, a local nearest neighbors algorithm is applied to each seed controlled by a distance which combines color, spatial and shape distances (see equation \ref{eq:vccsdist}). Color distance is computed in CIELab color space and the shape distance is computed with FPFH.  
	
	\begin{equation}\label{eq:vccsdist}
		D = \sqrt{\frac{\lambda D_c^2}{m^2} + \frac{\mu D_s^2}{3 R_{seed}^2} + \epsilon D^2_{f}}
	\end{equation}	  
	
	Where $D_c$ is the distance on color CIELab space divided by a constant $m$, $D_s$ is the spatial distance divided by $R_{seed}$ which is the maximal distance considered for the clustering and $D_{f}$ is the distance on FPFH. For each of these distances, hyperparameters $\lambda$, $\mu$ and $\epsilon$ control the importance of color, spatial distance, and geometric similarity. This equation shows the four important hyperparameters of VCCS which control the size ($R_{seed}$) and the shape ($\lambda$, $\mu$, $\epsilon$) of the supervoxels. These last hyperparameters do not seem to be a limitation of the variety of environment on which our methods can be applied as they are not environment-specific. For the experiments presented in this article, the parameters were fixed to the values proposed by \citet{Papon2013}. However, the size of supervoxels is more critical as objects must be at least larger than a supervoxel.  
	
	Another drawback of VCCS methods is when extracted from a video stream on a static scene, the segmentation will change at each frame. This inconsistency of the segmentation over a video prevents to apply supervoxels tracking methods. 

In this work, supervoxels are, the smallest visual unit considered. All visual features are extracted from a supervoxel and also the robot interacts with the supervoxels which have at least the size of its end-effector. We use the implementation of VCCS available in the PointCloud Library (PCL) (\citet{Rusu_ICRA2011_PCL}).

	VCCS algorithm output, like implemented in PCL is a set of supervoxels, with a centroïd point for each supervoxels. The centroid point is at the average position and have the average color and the average normal of the points part of its supervoxel. The output includes also an adjacency map representing the graph of geographical proximity of the supervoxels. Thus, to access to the neighbors of a supervoxel, it is enough to use the adjacency map.

\subsection{Features Extraction}\label{sec:feat}

To estimate the relevance of a supervoxel, a classifier is trained on features based on the color and shape of the supervoxels. Each feature characterizes one supervoxel. The features used in this paper are the following:

\begin{itemize}
\item Color CIELab histogram: A five-bin histogram is computed for each dimension of the CIELab\endnote{CIELab is an international standard of colorimetry decided during the International Commission on Illumination (CIE) of 1978} color domain on the colored pointcloud of a supervoxel. Then, these three histograms are concatenated in a 15-bins histogram.
\item Fast Point Feature Histogram (FPFH) is a common descriptor that characterizes shape based on a pointcloud of normals \citep{rusu2009fast}. Among others, it is used in the computation of VCCS for oversegmentation in supervoxels. It is used generally for 3D registration thanks to its high capacity for shape discrimination. In this paper, FPFH is extracted from the pointcloud including the targeted supervoxel and its neighbors. The average descriptor is computed to finally obtain a 33-dimensional feature of the supervoxel.
\end{itemize}

\begin{figure}[h]
\centering
\includegraphics[width=.9\linewidth]{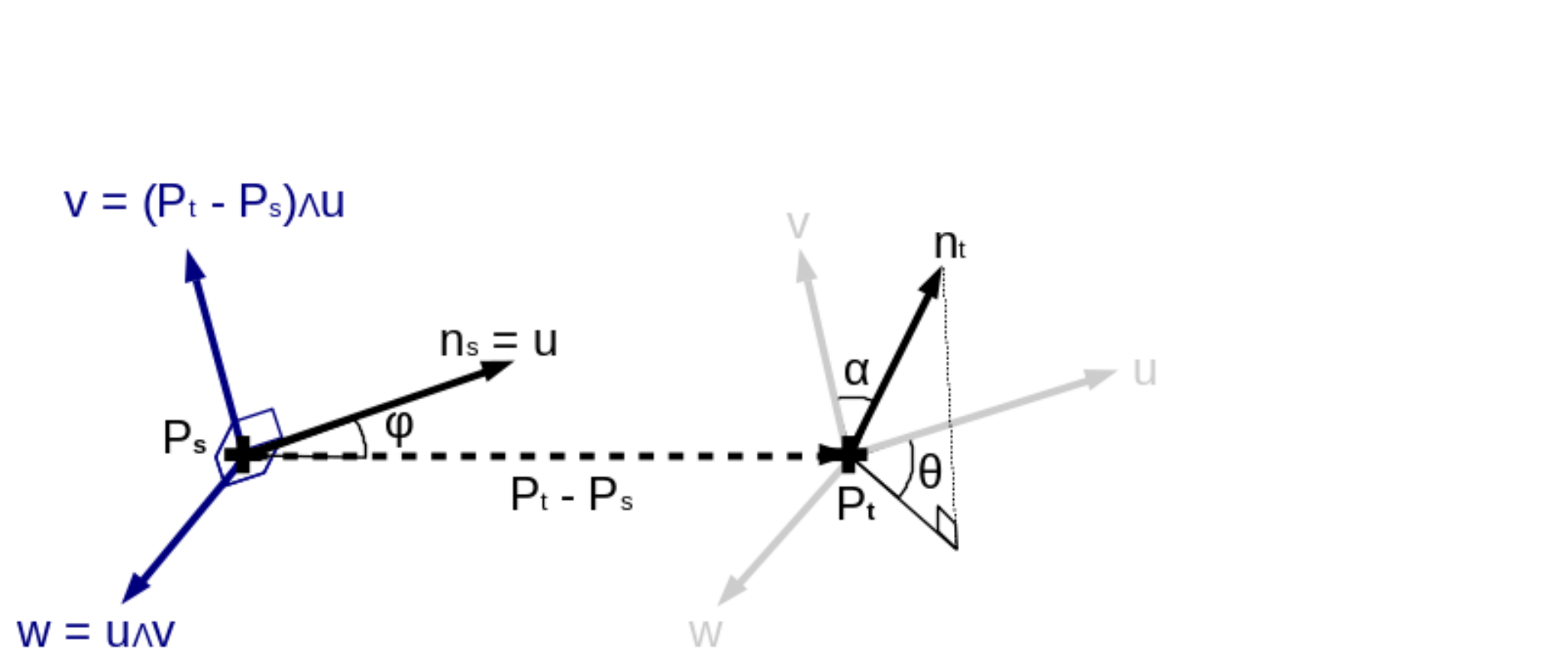}
\caption{Schema of how the orthogonal frame $(u,v,w)$ is defined on which the computation of SPFH is based. Figure reproduced from \citet{rusu2009fast}.} 
\label{fig:spfh}
\end{figure}

FPFH is a simplification of PFH (point feature histogram) aimed at being faster to compute. FPFH is a combination of simplified PFH (SPFH) which are histograms of triplet of value $(\alpha,\phi,\theta)$ computed with equations \ref{eq:spfh}.
\begin{equation}\label{eq:spfh}
\begin{split}
\alpha & = v*n_t \\
\phi & = u*\frac{p_t-p_s}{\Vert p_t - p_s \Vert} \\
\theta & = arctan(w*n_t,u*n_t) 
\end{split}
\end{equation}
Where $(. * .)$ is the scalar product, $(u,v,w)$ is an orthogonal frame defined in equation \ref{eq:spfhframe} and represented in figure \ref{fig:spfh}, $n_t$ and $n_s$ the normals to the surface at points $p_t$ and $p_s$. 
		
To apply these equations, a coordinate frame is defined at one of the points like shown in figure \ref{fig:spfh} and written in equation \ref{eq:spfhframe}.  
			
\begin{equation}\label{eq:spfhframe}
\begin{split}
u & = n_s \\
v & = u \wedge \frac{p_t-p_s}{\Vert p_t - p_s \Vert} \\
w & = u \wedge v
\end{split}
\end{equation}	
Where $(. \wedge .)$ is the vectorial product.
	
An SPFH is computed for the query point $p_q$ for its neighbors. Thus, the computation of FPFH involves the neighbors of $p_q$ and also the neighbors of the neighbors of $p_q$.

The concatenation of the CIELab histogram and FPFH features characterizes a supervoxel. It is a vector of size 48.

\subsection{Building the Relevance Map}\label{sec:sm}

Each supervoxel is weighted with a value between 0 and 1. These values represent the relevance of the supervoxel. These relevances are predictions of the classifier trained online during the exploration. They represent the probability of a supervoxel to be part of "something" moveable by the robot, i.e. an object. The relevance map is represented in this article as a grayscale map on a 3D pointcloud segmented in supervoxels. Each supervoxel is colored between yellow (for maximum relevance) and black (for minimum relevance). The classifier is described in detail in section \ref{sec:method}.

\subsection{Choice of the Next Area to Explore}\label{sec:choice}

After computation of the relevance map, the process must select the next region of the environment to explore. Therefore, a \textit{choice distribution map} is computed. This \textit{choice distribution map} represents the probability of a region to be chosen. The computation of these probabilities is based on the \textit{uncertainty} and \textit{confidence} of the classifier. The exploration is motivated by a reduction of uncertainty or an increase in the level of information (i.e. the entropy) with respect to the environment. The computation of the \textit{choice distribution map} is described in detail in section \ref{sec:samplproc}.

\subsection{Push Primitive}\label{sec:contr}

To enable interaction with the chosen supervoxel, a push primitive is used. This primitive is divided into three steps: approach movement, straight line motion towards the supervoxel center for interaction, and reverse motion.
A planning algorithm with obstacle avoidance is used for the approach phase \citep{sucan2012ompl,moveit}. In this phase, the end-effector of the robot moves to an approach position near the target position, which is the supervoxel center. An approach position is randomly chosen among those associated with a valid motion plan, i.e. a motion plan without self-collisions and collisions with the scene. At the end of this phase, the end-effector is at an approach point and positioned towards the target. Then, the end-effector moves to the target following a straight line of 5 centimeters and attempts to pursue this trajectory for a further few centimeters\endnote{This control sequence is open-loop, the system updates its perception of the environment only after the execution of the control program.}. In other words, the robotic arm tries to push the target. Thereby, pushing motions can take different orientations of the end-effector relative to the target before pushing it. Finally, the robot arm returns to its home position by following the reverse trajectory.  

\subsection{Change Detection} \label{sec:motdet}

As the exploration is sequential, the change detection is simply a comparison between the pointcloud before and after the interaction. Thus, the detection is based only on vision. The comparison follows these steps:
\begin{itemize}
\item The \textit{octree pointCloud change detector} method provided in PCL \citep{Rusu_ICRA2011_PCL} is used to subtract the initial pointcloud (before the interaction) from the current pointcloud. This operation produces a pointcloud limited to the difference between both pointclouds.
\item With a \textit{statistical outlier removal} provided in PCL , the noise in the difference pointcloud is reduced.
\item Finally, points \textit{only in the selected supervoxel} are compared with the difference pointcloud, using an \textit{iterative closest point} algorithm implemented in PCL, to determine if this group of points is in the difference pointcloud. If this is the case, it is considered that the action of the robot produced a movement, and the feature of the supervoxel is given a "moveable" label; otherwise, it is given a "non-moveable" label.   
\end{itemize}

\begin{figure}[!h]
\centering
\includegraphics[width=.9\linewidth]{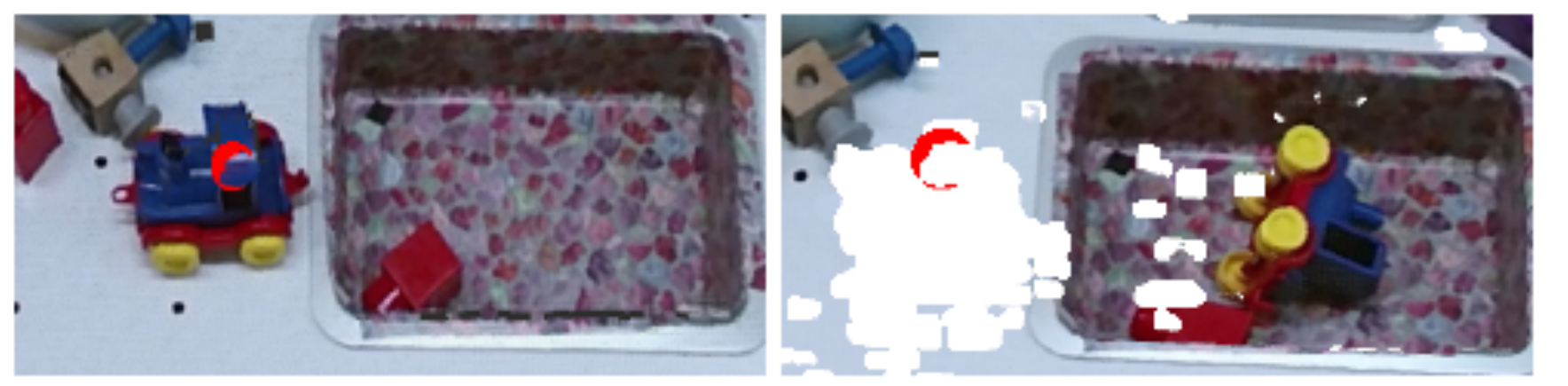}
\caption{Visualisation of the change detector. Right picture represents a part of a scene before a push and the left picture after a push. The red dot on both pictures represents the target of the push primitive which is here the upper part of the blue toy. This target correspond to the center of a supervoxel. The white areas represent the parts detected as different between both images.}
\label{fig:chdet}
\end{figure}

Figure \ref{fig:chdet} shows a visualisation of how the change detector works. The right picture represents a part of a scene before a push and the left picture after a push. The red dot represents the center of the supervoxel targeted by the system. The white areas represent the differences detected between both images. If the target supervoxel is included in the white areas, then a change is detected. The change detector considers only the targeted supervoxel, i.e. a small area around the target. 

\section{Collaborative Mixture Models}\label{sec:method}

To classify the collected samples, a new classification method is introduced: the collaborative mixture models (CMMs).
The proposed method consists in classifying the samples extracted from the supervoxels in two categories: samples from a relevant object and samples from the background. CMMs predict the class of the newly perceived supervoxels and use this information to build a relevance map. This classification is supervised as the system labels the gathered samples owing to (i) the interactions of the robot with the environment and (ii) the change detector. These classes may be nonconvex and the collected samples may be nonlinearly separable as we make no assumptions about the feature space structure and about the environment. Both categories are represented as a set of clusters of the already seen samples. A multivariate normal distribution is associated with each cluster, and these are summed to form a mixture model;  there is thus a Gaussian mixture model (GMM) for each category. The parameters of the multivariate normal distributions are statistically estimated using sample mean and sample covariance estimators. Each Gaussian of the mixture models is associated with a cluster of samples, which is called, a \textit{component}. The number of components in each model is not given \textit{a priori} and is adapted to the training set. Both models start with one component and add or remove new components with \textit{merge} and \textit{split} operations. These operations adapt the number of components to the data. 


\begin{table*}[t!]
\centering
\resizebox{2\columnwidth}{!}{
\begin{tabular}{|l|l|l|l|l|l|l|}
\hline
Ref & Type & non-convex & non-linear & uncertainty & environment specific hyperparameters & supervised \\ \hline
\citet{cauwenberghs2001incremental} & SVM & yes & yes & no & kernel, soft margin & yes \\ \hline
\citet{bordes2005fast} & SVM &  yes & yes & yes & kernel, soft margin & yes \\ \hline
\citet{bordes2005huller} & SVM & yes & yes & no & kernel, soft margin & yes \\ \hline
\citet{tax2003online} & SVM & yes & yes & no & kernel, soft margin & no \\ \hline
\citet{saffari2009line} & random forest & yes & yes & no & -- & yes \\ \hline
\citet{saffari2010online} & boosting & no & yes & no & -- & yes \\ \hline
\citet{cappe2009line} & mixture model & yes & yes & yes & number of components & no \\ \hline 
\citet{kristan2008incremental,kristan2011multivariate} & mixture model & yes & yes & yes & level of compression & no \\ \hline \hline
Our Approach & GMM & yes & yes & yes & tolerance ellipse size ($\alpha$) & yes \\ \hline
\end{tabular}
}
\caption{Online learning algorithms review}
\label{tab:online}
\end{table*}

\subsection{Choice of Classifier}\label{sec:choiceclassif}

The main goal of this work was to develop a method that can handle a large variety of environments. Therefore, the classifier must be able to adapt to different environments. The values of its hyperparameters should not be specific to a particular environment.  In a complex  environment,  the classes may not be convex, and the training dataset extracted from it may not be linearly separable. Furthermore, the classifier must provide a measure of uncertainty, which will be used for the exploration process. The output of the classifier then needs to be a probability rather than a single net value. The choice of the classification algorithm must then fulfill the following criteria:

\begin{itemize}
\item[\textbf{1}] Handle nonconvex/nonlinearly separable datasets and feature spaces
\item[\textbf{2}] Provide uncertainty measurement
\item[\textbf{3}] Have hyperparameters that are not specific to a particular environment
\item[\textbf{4}] Be supervised and adapted for classification
\item[\textbf{5}] Be trained online
\end{itemize} 

We chose to use GMM as a basis for the classifier. This model offers a good classification approach that meets the first and second criteria. The GMM is a regression method capable of approximating a large set of probabilistic distributions. By encoding each category with a GMM, we have a supervised learned classifier that gives as its output the probability of membership to a certain category (criterion 2). Also, the GMM can be seen as an ensemble learning method in which weak classifiers are combined to form a strong classifier. It is thus a classifier able to handle nonconvex classes or nonlinearly separable data (criterion 1).  

The number of components (i.e. multivariate normal distributions) is generally a hyperparameter fixed before the training of the GMM \citep{mclachlan2004finite}. In the proposed method, this parameter depends on the environment because the more complex the scene in the feature space, the more components are needed (see section \ref{sec:res}). It is also trained with the expectation-maximization algorithm, which is, in its classical form, an offline algorithm. The expectation-maximization algorithm can estimate the latent parameters of a probabilistic distribution; however, in the case of mixture models, the number of components must be known beforehand. This is problematic with respect to the third and fifth criteria. 

\citet{cappe2009line} proposed an online expectation-maximisation algorithm in which E-step is replaced by a stochastic approximation of expectation and let M-step unchanged. GMMs with an unknown number of components were studied by \citet{richardson1997bayesian} and \citet{rasmussen2000infinite}, who used a Markov-chain Monte Carlo approach to estimate the parameters and the number of components of the mixture. In these studies, GMM is trained offline.  The study of \citet{cappe2009line} and the studies of  \citet{richardson1997bayesian} and \citet{rasmussen2000infinite} are not intercompatible because the first study use expectation-maximisation and the other two used Markov-Chain Monte Carlo. We could either extend the work of Capp\'{e} and Moulines with an unknown number of components or extend the work of Richardson and Green or Rasmussen to include online training. Richardson and Green chose a Bayesian formalism and Markov-Chain Monte Carlo because they argued that it is more suitable for estimating the number of components than expectation-maximisation.   

An alternative approach is to build a mixture model with kernel density estimator (KDE) as in the studies of 	
\citet{kristan2008incremental,kristan2011multivariate}. KDE algorithms proposed by \citet{kristan2008incremental,kristan2011multivariate} consist, in each training iteration, to add a new component with, as center, the last sample arrived, and then to reduce the mixture with a compression operation. This algorithm has two advantages : it is an online training and the number of components does not have to be specified beforehand. But the compression step could be computationally heavy especially in the multivariate case. Also, the frequency of the compression step must be well chosen as a compromise between the precision of the model in relation to the training data and its generalization ability. If too few compression steps are applied, the model will over-fit the data because there will be too much components, but if too much compression is done, the model will be inaccurate on the training data. 

In a similar idea, \citet{declercq2008online} proposed a regression method based on bivariate GMM. The model is trained online and the number of component is estimated thanks to split and merge operations. As in \citet{kristan2008incremental, kristan2011multivariate}, the new data is added to the model as a Gaussian component with a prior covariance representing the observation noise. This new component is merged to the most uncertain component in the existing mixture. The uncertainty of a component is estimated thanks to an uncertain gaussian model which provides a quantitative estimate of the ability to describe data that follows a gaussian distribution. This measure is called \textit{fidelity}. If the fidelity of the newly merged component is below a certain threshold, then this component is split into two new components. The parameters of the new components are estimated via expectation-maximization. 

The proposed methods of \citet{kristan2008incremental, kristan2011multivariate,declercq2008online} are interesting as they associate online learning with an estimation of the number of components, but they do not address the question of the sampling process. They also consider regression problems whereas in this study the problem is formulated as a classification problem. In a classification problem the output space is discrete, i.e. the labels,  while in regression the output space is continuous. In our opinion, considering labels suit more to our problem which is discriminating features between those characterizing moveable elements and those characterizing non-moveable elements. When, for instance, regression is practical for estimating trajectories or continuous signals.  
 
In our approach, we chose a Bayesian formalism and a statistical estimation of the parameters with geometrical criteria exploiting the labels of the training dataset to estimate the number of components. These choices allow us to have less latent parameters than did Richardson and Green or Rasmussen; thus, we avoid choosing a priori probability distributions for the parameters. However, other online classification methods are proposed in the literature. Next section gives a short review of some of these algorithms.

\subsection{Short Review of Online Classification Algorithms}

In offline learning, a set of training data is available beforehand, and the training can thus be performed in batch. In online learning, the data arrive progressively while the system is learning. As such, the classification algorithm must make predictions while is learning, on the basis of the history of previously encountered data. Online learning is of interest when not all data are available beforehand. In a first case, the data arrive in a stream like with financial prediction or real-time video analysis, thus the system has no control on the order of arrival of the samples. In a second case, the system has an exploration process which allows it to choose the next sample to pick and to this end, it needs to train a model online. This is the case of the present study. It is widely studied for application to several kinds of problems, e.g. speech or pattern recognition, failure prediction, and medical imaging \citep{obermaier2001hidden,salfner2010survey,saffari2010online}.

Table \ref{tab:online} summarizes the features of some state-of-the-art methods for online classification according to the criteria listed in section \ref{sec:choiceclassif}. These methods have an efficiency close to that of batch learning but they do not fulfill all of our criteria. 

Saffari et al. proposed an online random forest \citep{saffari2009line} and an online boosting forest \citep{saffari2010online}. Random forest is an efficient algorithm for classifying nonconvex and nonlinear problems, with boosting the method is faster but less efficient on nonconvex problems. These algorithms are used to track faces in videos in real time. In this task, exploration is not needed; thus, estimation of the uncertainty or confidence of the classification to drive a sampling process is not required. As such, this issue is not addressed. 

From random forest algorithm, an option for measuring uncertainty is to use the confidence of classification of a tree. This confidence is used to estimate which tree will give the best prediction for a given sample. And so, the algorithm estimates the area of classification in the features space of each weak classifier.

 \citet{Craye2015} used online random forest as  a classifier and intelligent and adaptive curiosity (IAC) \citep{oudeyer2004intelligent} as the exploration process. IAC does not involve uncertainty estimation, but it is based on learning progress maximization; which requires to segment the environment into areas  to compute learning progress, i.e. entropy gain. In the study of Craye et al. these areas are defined by a human.

Online versions support vector machines (SVM) are also available \citep{tax2003online,bordes2005fast,bordes2005huller,cauwenberghs2001incremental}. By definition, SVM methods fulfill the first and the forth criteria. SVMs are highly dependent on the type of kernel used to separate the data. To have an efficient classification with SVMs, the kernel and its hyperparameters must be well chosen based on the problem \citep{burges1998tutorial,smits2002improved}; this does not meet our third criterion. However, an uncertainty measurement (2nd criterion) could be easily defined on the basis of the soft-margin. The soft-margin could define an uncertain classification area (\citet{bordes2005fast}). Of course the parameter of the soft-margin must be well chosen between precise classification and large exploration areas.  

Random forest may be a good choice as it is efficient and general; however, it is not an algorithm designed to drive the exploration of an unknown feature space. A multivariate normal distribution gives a better estimation of uncertainty as it is a statistical approximation of a sample distribution. 
 Our problem therefore requires a new classification algorithm that can draw inspiration from previous algorithms: in particular, online mixture models and mixture models with an unknown number of components \citep{cappe2009line,richardson1997bayesian,kristan2008incremental, kristan2011multivariate,declercq2008online}. 

\subsection{Definition of the Classifier}

CMMs are formalized in a Bayesian framework with conditional probabilities. 

The classifier has the following parameters :
\begin{itemize}
	\item $K_l$ : number of components of the mixture model of class $l$, with $l \in \{0,1\}$.
	\item $S = \{s_i,l_i\}_{i<I}$: database of samples and their corresponding label.
	\item $\Theta_l = \{\mu_k,\Sigma_k\}_{k<K_l}$: multivariate normal distribution parameters of both models with mean $\mu_k$ and covariance matrix $\Sigma_k$.
	\item $W_l = \{w_k\}_{k<K_l}$ : weights of the mixture model of class $l \in \{0,1\}$.
	\item $L = \{0;1\}$ : label asked to the classifier to be predicted.
\end{itemize}

\paragraph*{Class Definition} A class is a subspace of the feature space pointed out by a label. Equation \ref{eq:class_est} gives the probability of a sample X to be part of  class 1.

\begin{equation}\label{eq:class_est}
P(L = 1 | W, \Theta, X) = \frac{1 + \Gamma(W_1,\Theta_1,X)}{2 + \Gamma(W_1,\Theta_1,X)+\Gamma(W_0,\Theta_0,X)}
\end{equation}
Where $\Gamma(W_i,\Theta_i,.)$ is the GMM of label i, $W = W_0 \cup W_1$ and $\Theta = \Theta_0 \cup \Theta_1$.

Equation \ref{eq:class_est} generates the output of the classifier.

\paragraph*{Component Definition} A component is a set of points of the feature space statistically represented by a multivariate normal distribution. A component is part of a class, i.e. all parts of a component have the same label or are members of the same class. Equation \ref{eq:comp_est} gives the probability for sample X to be part of a given component $i$. 

\begin{equation}\label{eq:comp_est}
P(k=i|X,\Theta,l) = \frac{w_i*G(\mu_i,\Sigma_i,X)}{\sum^{K-1}_{k=0}{w_k*G(\mu_k,\Sigma_k,X)}}
\end{equation}

Let $C_k(X) = (w_k,G(\mu_k,\Sigma_k,X)),S_k,l)$ be a component and let $M_l = \{C_k\}_{k<K_l}$ be the set of $K_l$ components of class $l$, where $S_k$ is the set of samples used to estimate $\mu_k$, $\Sigma_k$, and $w_k$.

The weights of the mixture models are computed using equation \ref{eq:weights}. 
\begin{equation}\label{eq:weights}
w_k = \frac{\vert C_k \vert}{\sum_i^K \vert C_i \vert} 
\end{equation}

\begin{algorithm*}[!t]
\caption{IAGMM algorithm}\label{algo:genalgo}
\begin{algorithmic}[1] 
\Procedure{IAGMM}{$(s,l),M_0, M_1$}
 \For{iter = $1 \rightarrow N$}
  \State Add new sample $(s,l)$ to the model :
   \If{$M_l = \emptyset$}
    \State $C \leftarrow \{w,G(s,cI,X),\{s\}\}$ \Comment{Create a new component}
    \State $M_l \leftarrow M_l \cup C$ \Comment{Add the new component is the model of label l}
   \Else
    \State $C \leftarrow closest\_component(s,l)$ \Comment{Find the closest component from $s$ with label $l$}
    \State Update the parameters of $C$ with the new sample $s$
    \If{SPLIT($C$,l,$M_0, M_1$) is not successful}
     \State MERGE($C$,l,$M_0, M_1$)
    \EndIf
   \EndIf
  \State Choose randomly one gaussian per class model :
   \For{$i = {0,1}$}
    \State $C \leftarrow random\_choice(M_i)$ \Comment{Randomly choose a component from $M_i$}
    \If{SPLIT($C$,l,$M_0, M_1$) is not successful} \Comment{Apply the split operation thanks to algorithm \ref{algo:split}}
     \State MERGE($C$,l,$M_0, M_1$) \Comment{Apply the merge operation thanks to algorithm \ref{algo:merge}}
    \EndIf
   \EndFor
 \EndFor
 \State \Return $M = M_0 \cup M_1$ \Comment{Return the whole model}
\EndProcedure
\end{algorithmic}
\end{algorithm*}

\subsection{Algorithm}\label{sec:alg}

CMMs rely on a supervised learning algorithm, which is used here to solve a binary classification problem (relevant vs not relevant). The algorithm builds two GMMs, one for each class. Each GMM is made up of several Gaussian distributions, each associated with a weight. Each distribution supports a cluster of samples, called a component (see the previous section). At each iteration, the robot arm interacts with a supervoxel and stores the corresponding sample with the label deduced from the observation of the interaction result. Thus, each iteration consists of adding one sample with its label to the dataset.

Adding a sample consists of three main steps (Algorithm \ref{algo:genalgo}):
\begin{itemize}
\item[1.] If there is no component yet in the class of the new sample, create a new one; otherwise, find the closest component and add the sample to this component. Finally, update the parameters of the component.
\item[2.] A \textit{split} operation is applied to the updated component. If it is not successful, the \textit{merge} operation is then applied.
\item[3.] One component per class is randomly chosen, and the \textit{split} operation is applied to each. If a selected component is not split then the \textit{merge} operation is applied.
\end{itemize}

\begin{figure}[h]
\centering
\includegraphics[height=60mm,width=.8\linewidth,keepaspectratio]{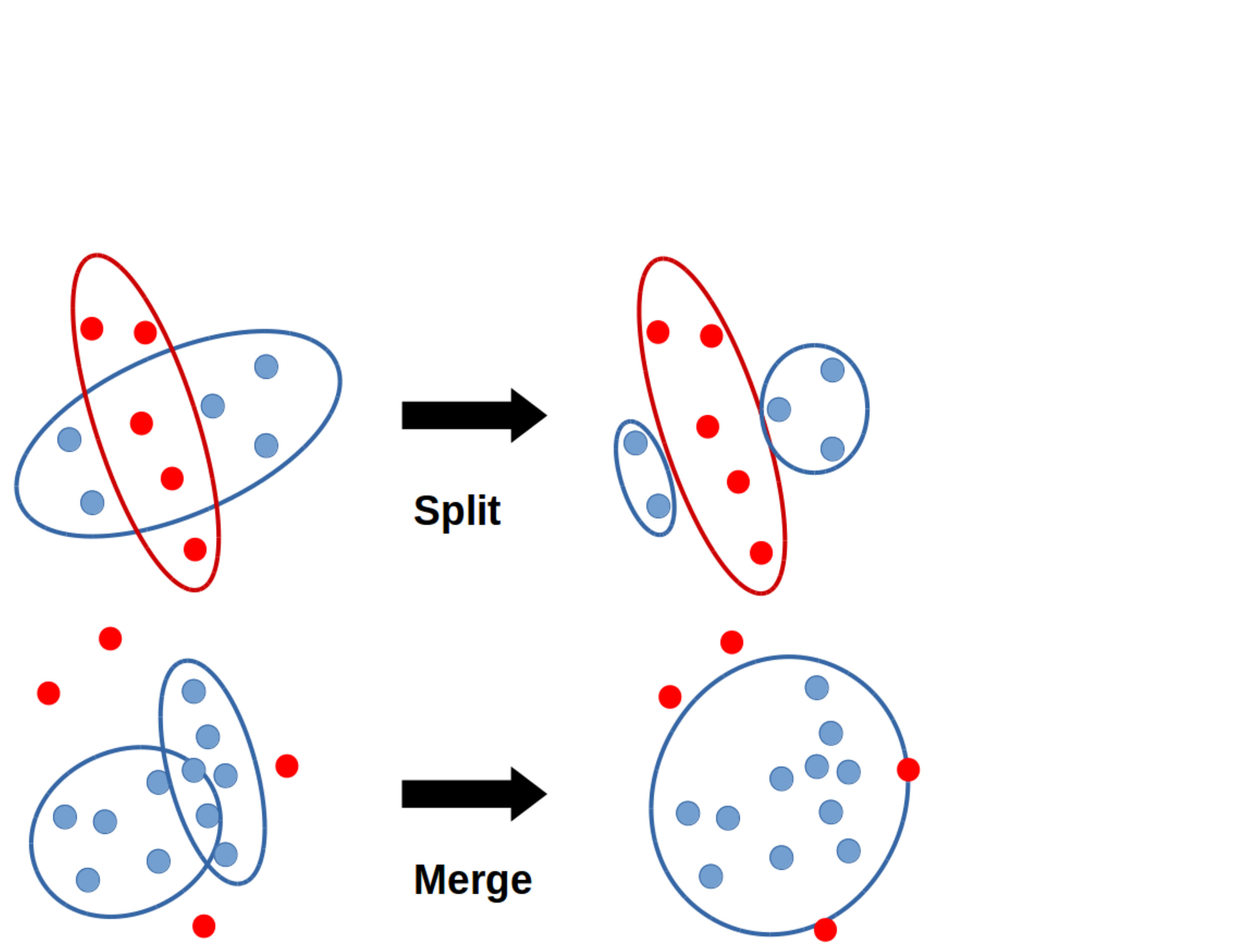}
\caption{Schema showing the cases when split and merge operations are applied. Top : If two components of opposite classes intersect themselves, apply the split operation. Bottom : If two components of same classes intersect themselves, apply the merge operation.}
\label{fig:split_merge}
\end{figure}

The goal of the \textit{split} operations is to keep only convex components. Indeed, multivariate normal distributions are only relevant to model convex spaces or sets of data as they can be represented as hyperellipses. The \textit{merge} operation aims at reducing the number of components to avoid overfitting and reduce computational cost. Figure \ref{fig:split_merge} illustrates the cases where \textit{split} and \textit{merge} operations are applied :
\begin{itemize}
\item \textbf{Split Case:} When two components of different categories cross or intersect, one of these components is nonconvex. This component must be split into two new components.
\item \textbf{Merge Case:} When two components of the same category cross or intersect, they can be merged to form a larger convex component.
\end{itemize}

\paragraph*{Components intersection} The \textit{split} and \textit{merge} operations are based on a geometrical interpretation of the components.  A multivariate normal distribution can be represented geometrically as an hyperellipsoid. The proposed geometrical interpretation relies on the region of tolerance of the distribution. A region of tolerance represents the area in which all the points have a probability greater than $1-\alpha$ of being in the corresponding component. 
The axes of the hyperellipsoid are the eigenvectors of the inverse of the covariance matrix. In both cases (merge and split), there are two components that intersect each other. The parameters of the hyperellipsoid of tolerance can thus be computed as:

\begin{equation}\label{eq:elltole}
(X - \mu)^{T}\Sigma^{-1}(X - \mu) = \frac{(n-1)p}{n-p}\frac{n+1}{n}F_{1-\alpha}(p,n-p) 
\end{equation}
Where $n$ is the number of samples used to estimate $\mu$ and $\Sigma$; $p$ is the dimension of the features space; and $F_{1-\alpha}$ is the quantile function of the Fisher distribution.
 
Equation \ref{eq:intercond} is used to determine whether component $C_1$ intersects with another component $C_2$, where $C_1$ is the component candidate to be split or merged with $C_2$:
\begin{equation}\label{eq:intercond}
(\rho - \mu)^{T}\Sigma^{-1}(\rho - \mu) <= \frac{(n-1)p}{n-p}\frac{n+1}{n}F_{1-\alpha}(p,n-p) 
\end{equation}
Where $\mu$ and $\Sigma$ are the mean and covariance matrices of $C_1$; $\rho$ is the mean of $C_2$; and n is the number of sample in $C_1$.

It is worth noting that $n$ must be strictly greater than $p$ because both arguments of $F_{1-\alpha}$ must be strictly positive. Then, the candidate component must have more samples ($n$) than the number of dimensions of the feature space ($p$). In our case $n$ must be greater than $p = 48$. 

For all experiments, the parameter $\alpha$ is fixed at $0.25$.

\paragraph*{Split operation}
Algorithm \ref{algo:split} describes the split operation. Let C be a component. If C intersects a component of the other class, a model candidate is computed with C split into two new components. If this candidate model has a greater \textit{loglikelihood} than the current model, the candidate is kept. 

\begin{algorithm}[h]
\caption{SPLIT algorithm}\label{algo:split}
\begin{algorithmic}[1]
\Procedure {SPLIT}{$C$,l,$M_0$,$M_1$}
 \State $lbl \leftarrow |l - 1|$
 \For{Each $C' \in M_{lbl}$}
   \If{$C' \cap C \neq \emptyset$} 
    \State $C_1, C_2$ = $split(C)$ 
    \State $\tilde{M}_l \leftarrow (M_l \setminus \{C\}) \cup \{C_1,C_2\}$ 
    \If{$\tilde{L} > L$} \Comment{L is the log-likelihood of M}
     \State $M_l \leftarrow \tilde{M}_l$
    \EndIf
   \EndIf
 \EndFor
\State \Return $M_0 \cup M_1$
\EndProcedure
\end{algorithmic}
\end{algorithm}

\begin{figure}[h]
\centering
\subfloat[Step 1]{\label{fig:split1}
\includegraphics[height=60mm,width=.3\linewidth,keepaspectratio]{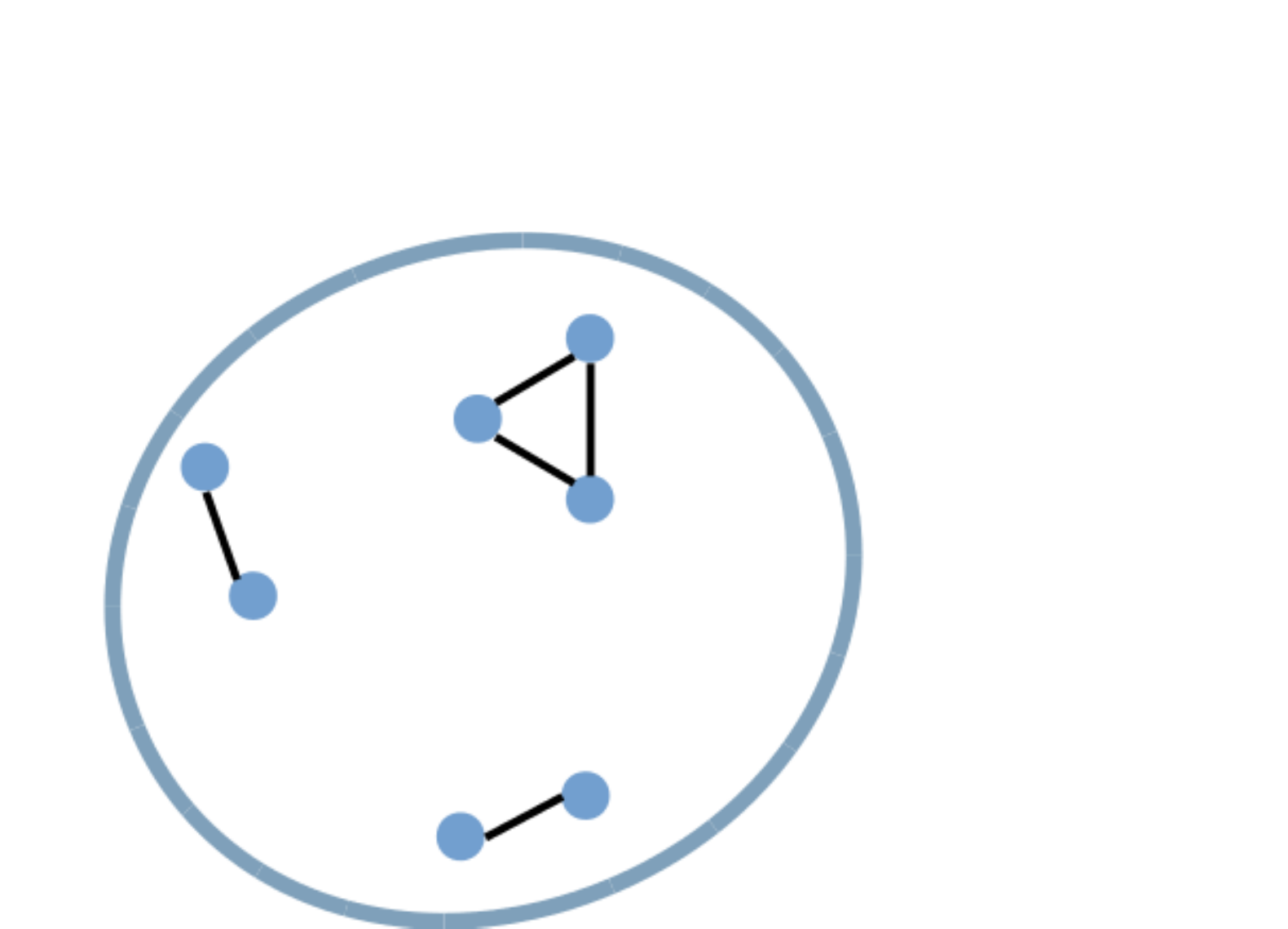}
} 
\subfloat[Step 2]{\label{fig:split2}
\includegraphics[height=60mm,width=.3\linewidth,keepaspectratio]{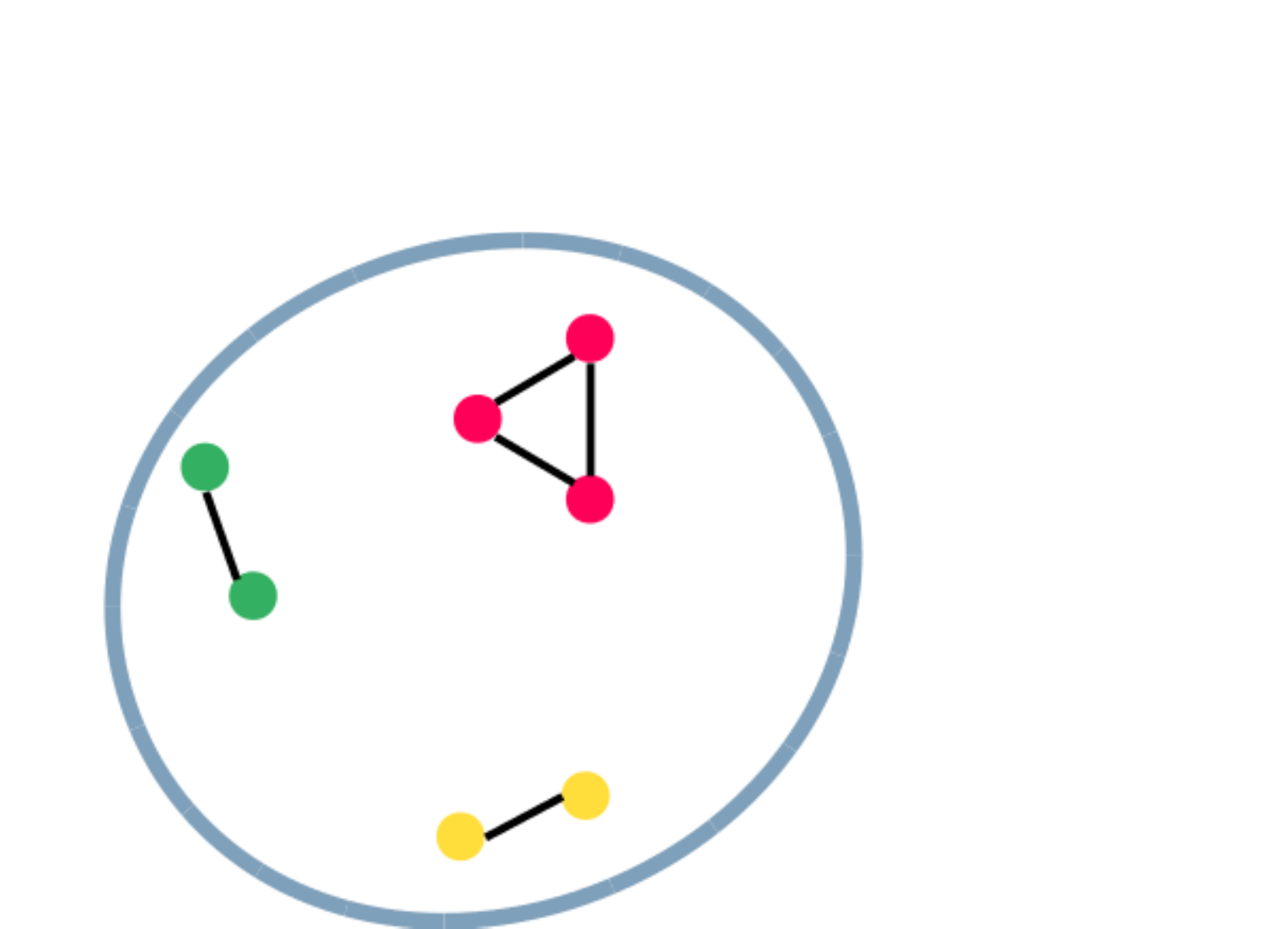}
} 
\subfloat[Step 3]{\label{fig:split3}
\includegraphics[height=60mm,width=.3\linewidth,keepaspectratio]{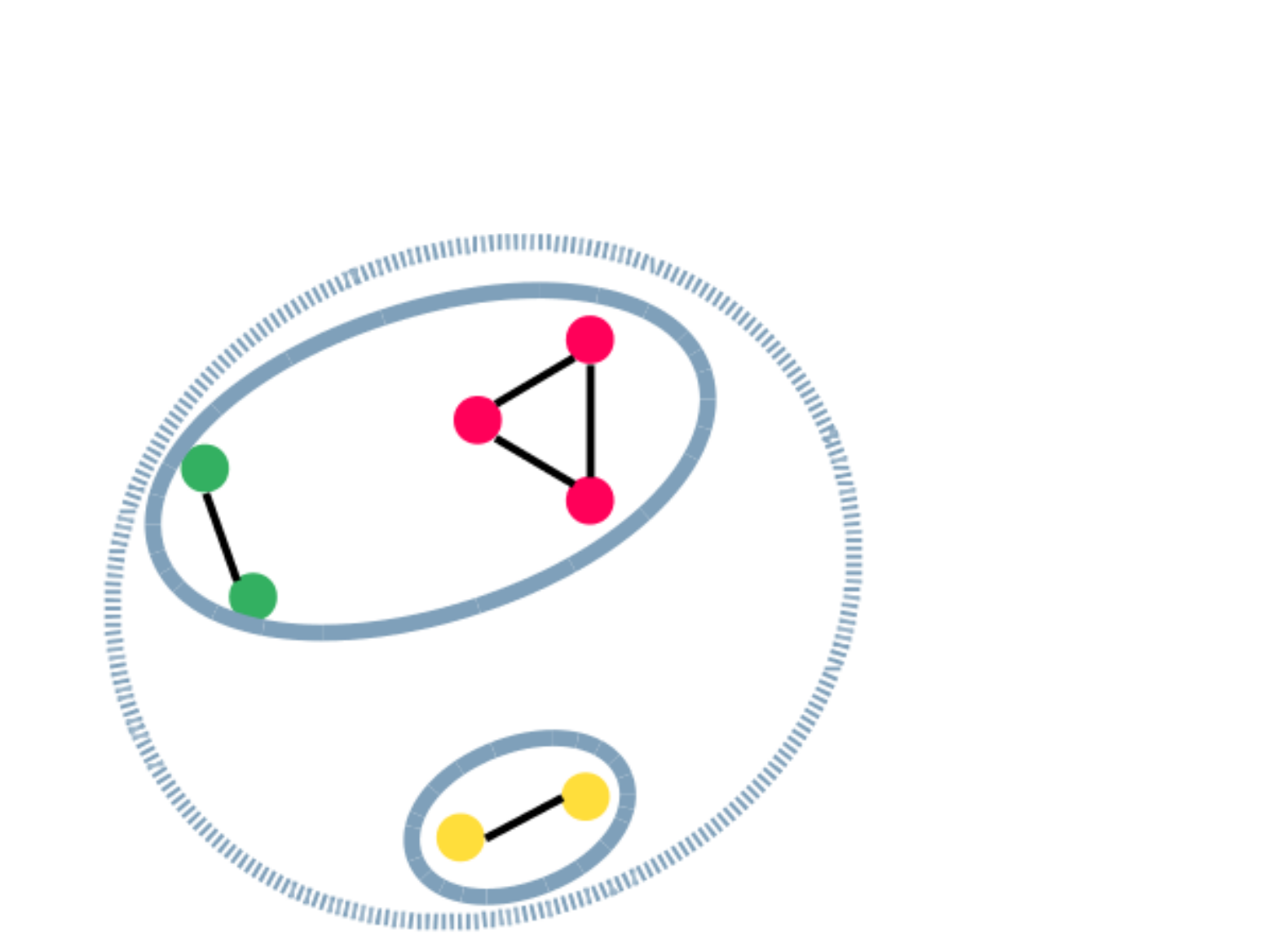}
} 
\caption{Illustration of how the samples are shared between two new components during a split.}
\label{fig:split_schema}
\end{figure}

The split algorithm used to share the samples of the query component between two new components is the following:
\begin{itemize}
\item[Step 1:] Build a graph of minimal distances between the samples of the components
\item[Step 2:] Build a set of samples per sub-graph in which all vertices are connected.
\begin{itemize}
\item If there is only one set then cancel the split.
\item If there are 2 sets then go to step 3.
\item If there are more than 2 sets then merge the closest sets together by average distance until having 2 sets remaining and go to step 3 
\end{itemize}
\item[Step 3:] Make two new components based on the two sets of samples, by computing the sample covariance and the sample mean.
\end{itemize}
This algorithm is illustrated in figure \ref{fig:split_schema}.

\paragraph*{Merge operation}

Algorithm \ref{algo:merge} describes the merge operation. Let C be a component. if C intersects a component C' of the same class, a candidate model is computed with C and C' merged. As for the split operation, if this model candidate has a greater \textit{loglikelihood} than the current model, the candidate is kept.

\begin{algorithm}[h]
\caption{MERGE algorithm}\label{algo:merge}
\begin{algorithmic}[1]
\Procedure {MERGE}{$C$,l,$M_0$,$M_1$}
 \For{Each $C' \in M_l$}
  \If{$C \cap C' \neq \emptyset$} 
   \State $\tilde{C} \leftarrow C \cup C'$ 
   \State $\tilde{M}_l \leftarrow (M_l \setminus \ {C,C'}) \cup \tilde{C}$
   \If{$\tilde{L} > L$} \Comment{L is the log-likelihood of M}
    \State $M_l \leftarrow  \tilde{M}_l$
   \EndIf
  \EndIf
 \EndFor
\State \Return $M_0 \cup M_1$ 
\EndProcedure
\end{algorithmic}
\end{algorithm}

\subsection{Sampling Process}\label{sec:samplproc}

As the classifier is trained sample by sample, i.e. online, the choice of the next sample is critical. A process is used to choose the next sample to explore. This process generates a distribution choice map over the supervoxels based on the prediction of the classifier.

For a pointcloud with N extracted supervoxels and $\{X_i\}_{i<N}$ as the set of features of the supervoxels, the choice probability $P_c(X_i)$ of the feature $X_i$ of the $i^{th}$ supervoxel is defined as follows:  

\begin{equation}\label{eq:samplproc}
P_c(X_i) = u(X_i)*(1-c(X_i))
\end{equation}
Where $u()$ is the classification uncertainty and $c()$ is the classification confidence.

\paragraph*{Uncertainty} As the classification is probabilistic, the output of the classifier can provide information on how certain the classification is. According to equation \ref{eq:class_est}, the closer to $\frac{1}{2}$ the probability of a sample to be part of a class, the higher the \textit{uncertainty} is. The following equation describes how the uncertainty is computed:

\begin{equation}\label{eq:u}
u(X_i) = 
\begin{cases}
f(p) & \vert S_1 \vert <= \vert S_1 \vert \\
f(1-p) & \vert S_1 \vert > \vert S_0 \vert
\end{cases}
\end{equation}
where $p = P(L = 1 | W, \Theta, X)$ and $f$ is the following function:

\begin{equation}\label{eq:xlogx}
f(x) =
\begin{cases}
 -2x(log(2x)-1) & x >= 0.5 \\
 -4x^2(log(4x^2)-1) & x < 0.5
\end{cases}
\end{equation}

\begin{figure}[h!]
\centering
\includegraphics[width=.8\linewidth]{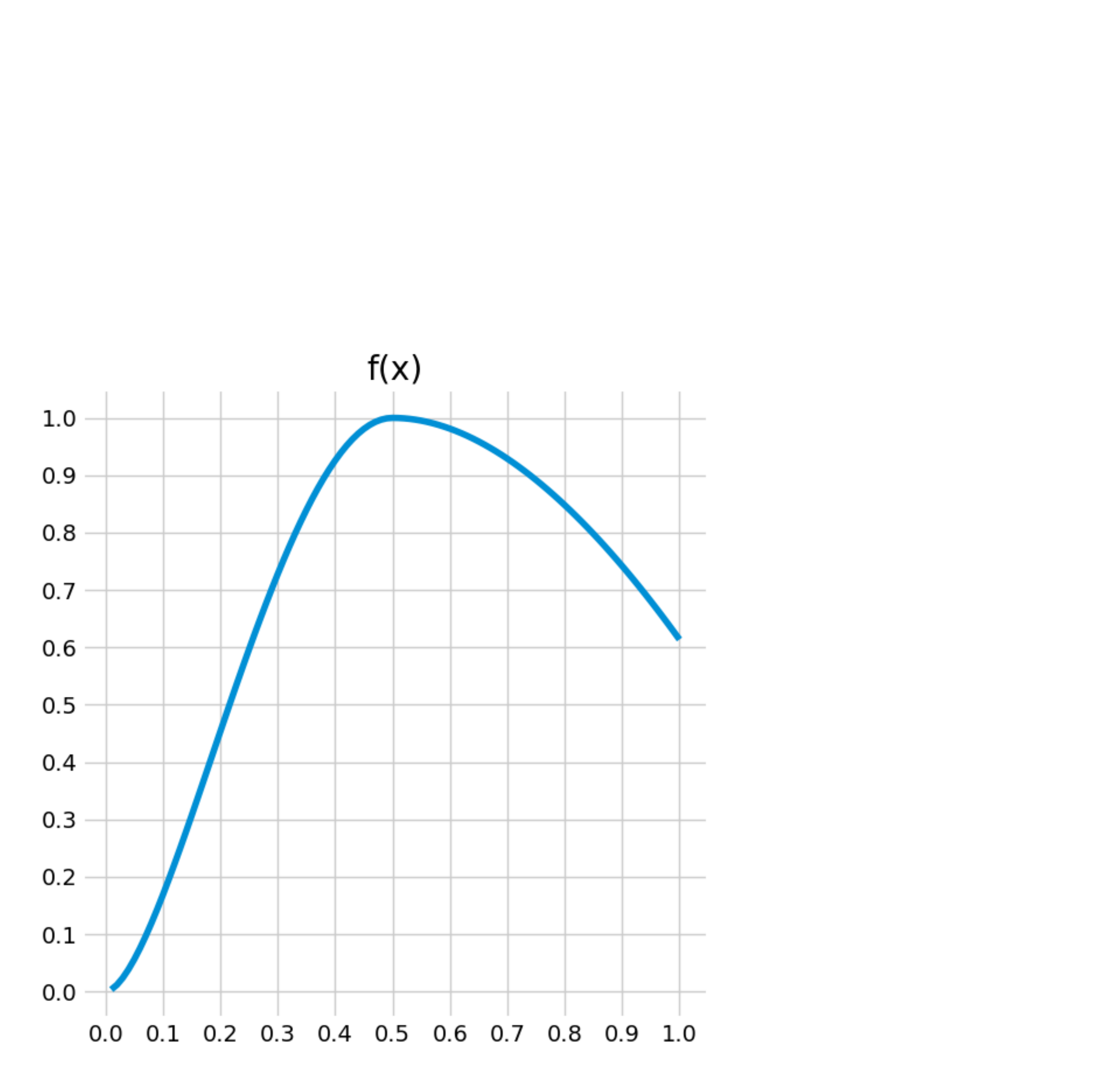}
\caption{Function used for uncertainty estimation. This function gives a higher probability of choice to uncertain classification, but also to certain classification to the chosen class, i.e the one with fewest samples.}
\label{fig:xlogx}
\end{figure}

Theoretically, with this definition of uncertainty, the exploration focuses in priority on uncertain areas to be part of each class (equation \ref{eq:xlogx} and figure \ref{fig:xlogx}). It also tries to keep the same number of samples for each class by choosing the class with the fewest number of samples gathered (equation \ref{eq:u}).

 This last feature is motivated by (i) the fact that, in most supervised learning problems, it is better to have a balanced number of samples, and (ii) the assumption that a balanced number of samples in each class better represents the environment. Issues related to the exploration process are discussed in section \ref{sec:disc}.

\paragraph*{Confidence} A GMM is the sum of Gaussian distributions. The classification is supported by a mapping of the feature space made by the Gaussians functions. Thus, the probability given by a multivariate normal distribution may provide useful information about the structure of the dataset. Given a sample X, its classification confidence is the probability of membership to the closest component defined in equation \ref{eq:comp_est}. The confidence gives a measure of the dataset density. In this way, the exploration focuses on areas with less information; therefore, confidence could be interpreted as an approximation of entropy. 

\section{Experiments}\label{sec:exp}

\subsection{Protocol}\label{sec:proto}

The experiments performed to validate the method are of two types: the ones in a simplified setup and the ones on a real robotic platform. All experiments have a fixed budget of interactions. Each experiment has a single fixed background and a set of mobile objects.

An expert is used to evaluate the quality of the classifier trained during an experiment. The expert is built by saving the pointcloud of the background without the objects at the beginning of the experiment; thus, the objects are easily separated from the background. To determine whether a supervoxel is part of the background, the points of the supervoxel are simply compared with the saved background pointcloud. This defines the ground truth of the classification, which is of course not known by the classifier. 


\begin{figure}[t]
\centering
\subfloat[\textbf{MoveableBalls:} Table with moving balls]{\label{fig:setup0}
\includegraphics[height=60mm,width=.7\linewidth,keepaspectratio]{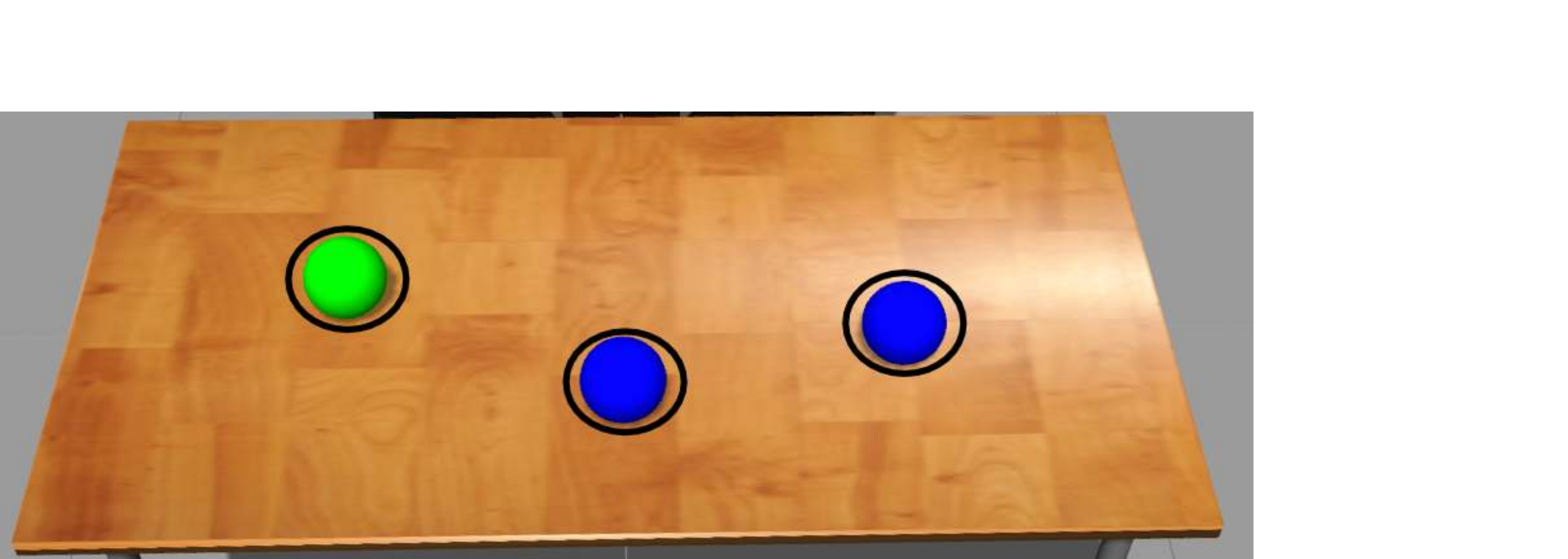}
} \\
\subfloat[\textbf{MoveableBricks1:} Table with moving bricks]{\label{fig:setup1}
\includegraphics[height=60mm,width=.7\linewidth,keepaspectratio]{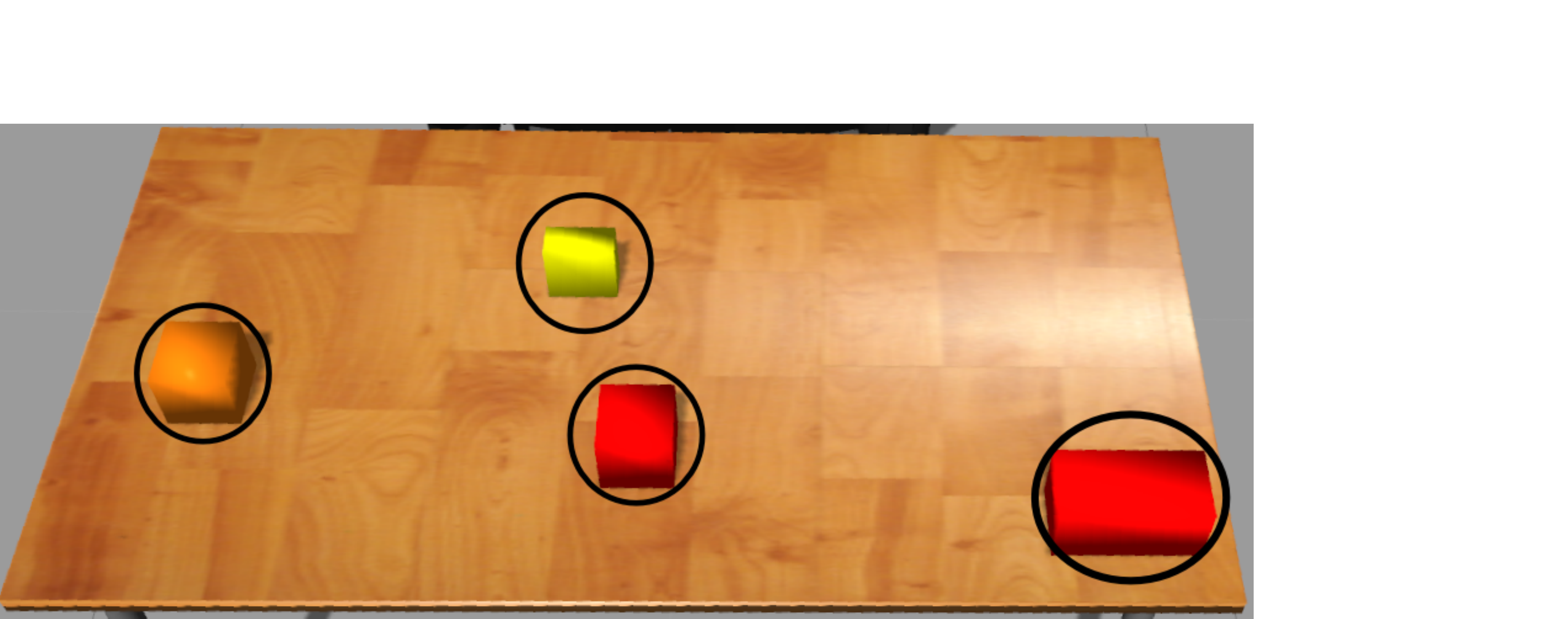}
} \\
\subfloat[\textbf{MoveableBricks2:} Table with fixed spheres and moveable bricks ]{\label{fig:setup3}
\includegraphics[height=60mm,width=.7\linewidth,keepaspectratio]{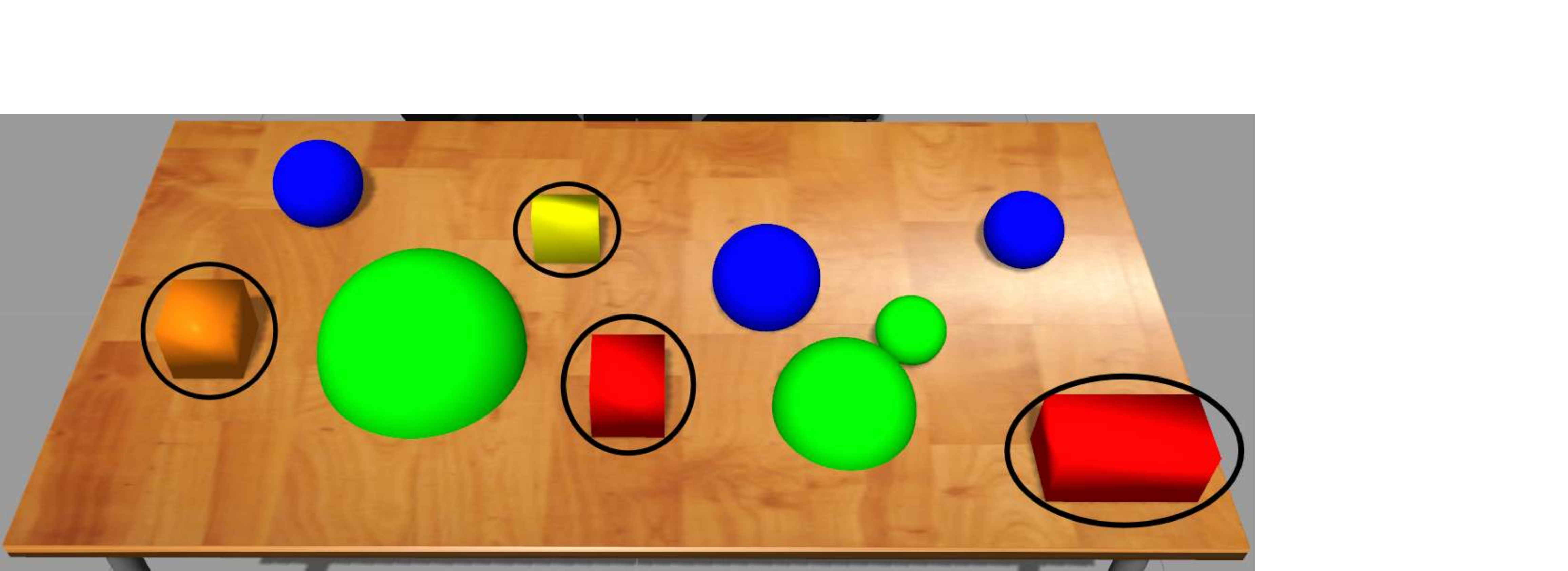}
} \\
\subfloat[\textbf{WhiteMoveableBalls:} Table with fixed spheres and moveable bricks ]{\label{fig:setup4}
\includegraphics[height=60mm,width=.7\linewidth,keepaspectratio]{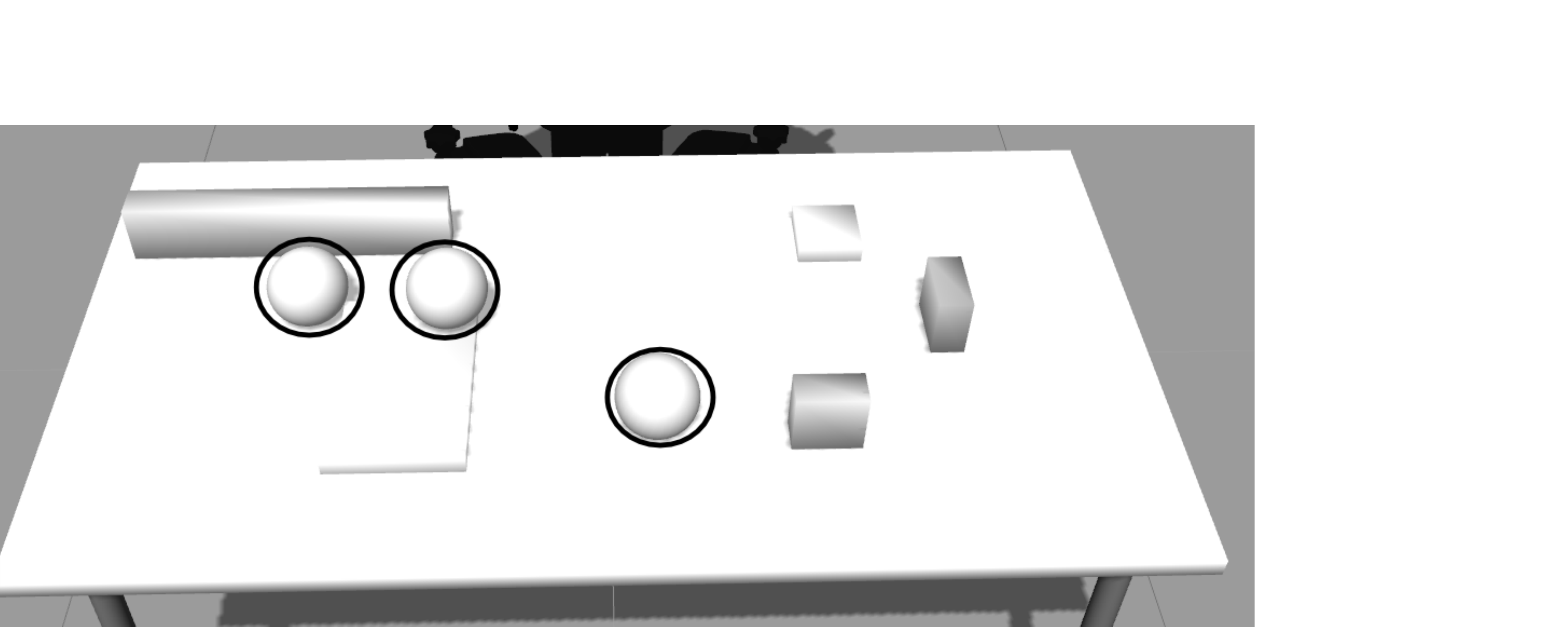}
} \\
\subfloat[\textbf{WhiteMoveableBricks:} Table with fixed spheres and moveable bricks all white]{\label{fig:setup5}
\includegraphics[height=60mm,width=.7\linewidth,keepaspectratio]{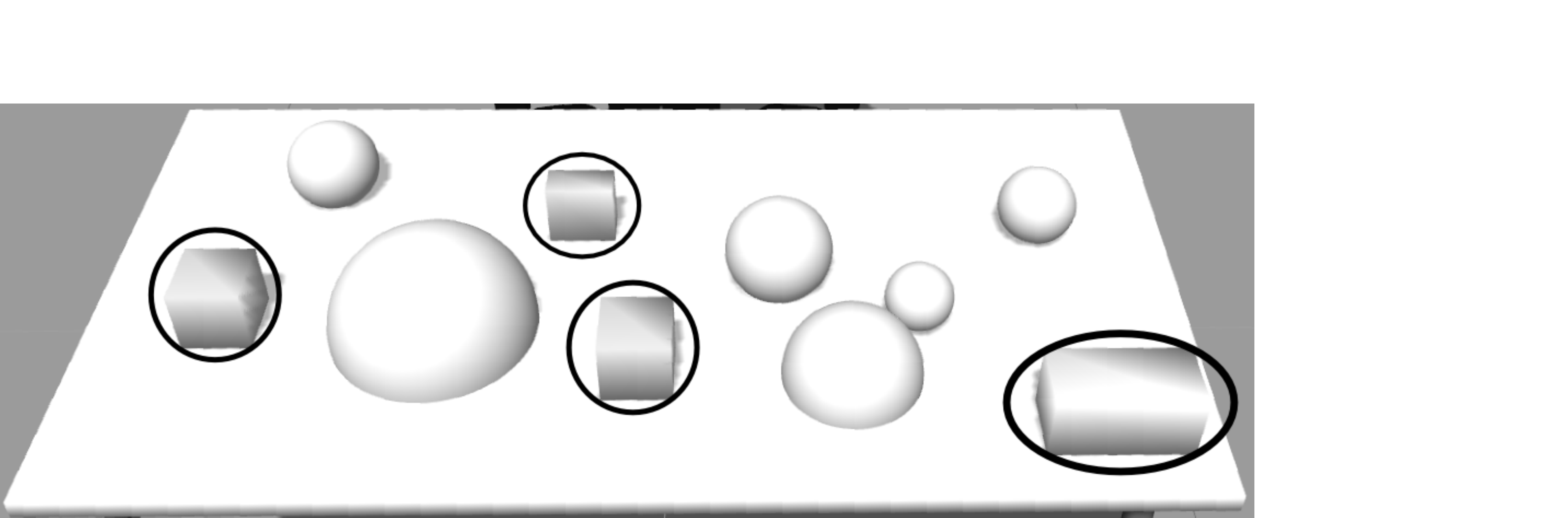}
} \\
\subfloat[\textbf{SimKitchen:} A kitchen with a teapot, a bawl, a cup and a spray cleaner]{\label{fig:simkitchen} 
\includegraphics[height=60mm,width=.7\linewidth,keepaspectratio]{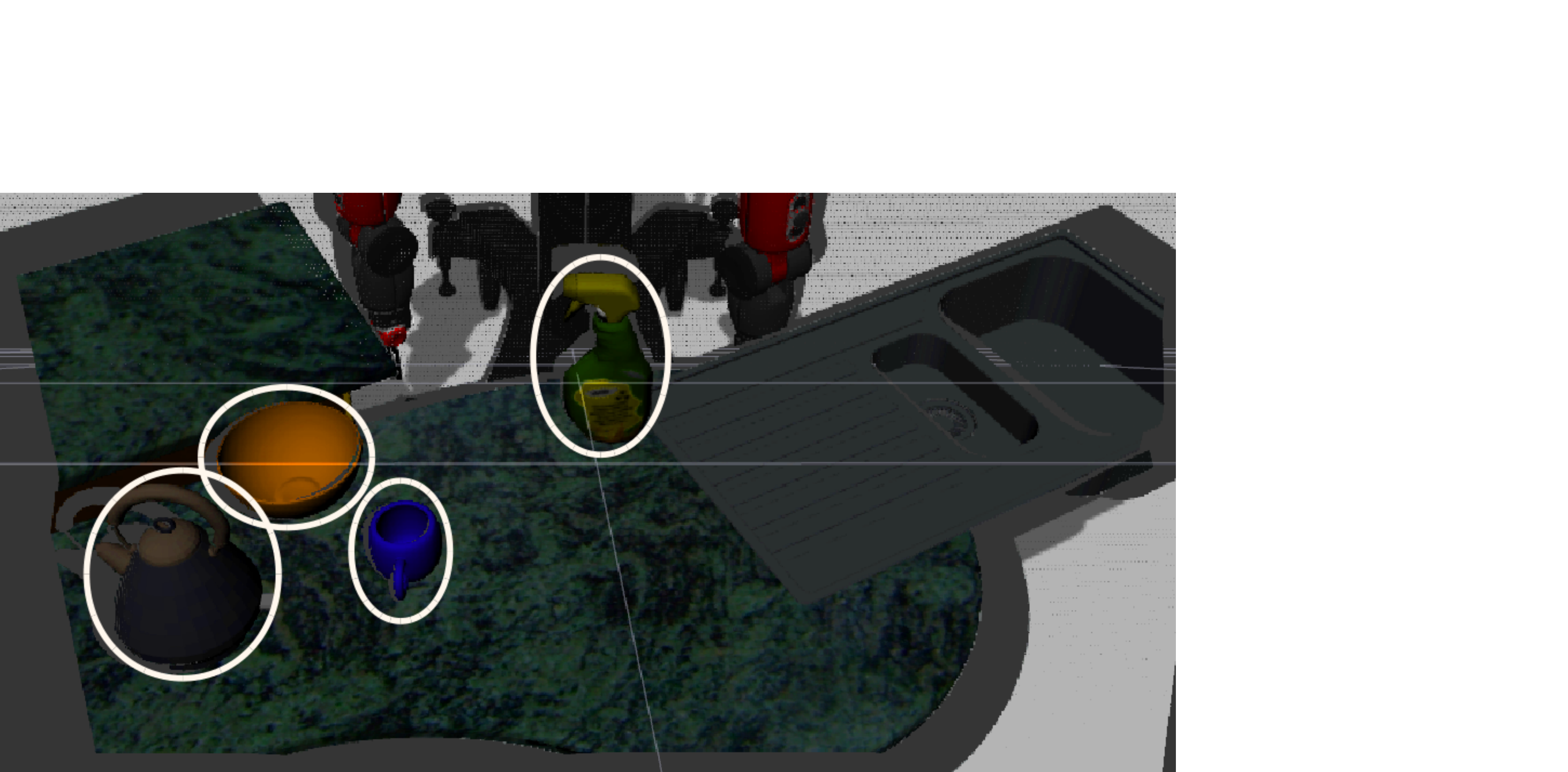}
} 
\caption{Experimental simplified setups ordered by increasing difficulty. The moveable objects are marked by black circles.}
\label{fig:setups}
\end{figure}

\paragraph*{Simplified setup}
The experiments conducted using the simplified setup are used to evaluate the method in an ideal case. It is ideal because we use the gazebo simulator without any robot and with a simulated kinect. Thus, the interactions are fake, i.e. there is no robot that interacts with the environment. The category (moveable or not) of the explored supervoxel is assessed by an expert that knows in advance which supervoxels are part of an object and which supervoxels are part of the background (described in the previous paragraph). In the simplified setup, there is then no noise and no mislabelled sample. The results in this case are an upper bound of what can be expected in reality.

The experiments are conducted in several environments (see figure \ref{fig:setups}). They are designed to have an increasing difficulty with an increasing number of shared features between the objects and the background. For instance, the setup \textbf{MoveableBalls} (\ref{fig:setup0}) is very simple because the background is a wooden table, a flat surface with colors between orange and yellow, whereas the moving objects are blue and green spheres. Thus, the feature space is very easy to split. At each iteration during the experiment, each object is spawned in a random position and with a random orientation.

The objects chosen for the 5 first setups are very simple (cubes and spheres) to simplify the analysis of results according to the feature space. The setup \textbf{SimKitchen} (\ref{fig:simkitchen}) is a more realistic setup. The experiments performed using the real robot use more complex object shapes.

\begin{figure}[h]
\centering
\subfloat[ \textbf{Workbench1:} Simple toy workbench with three moveable toy cars.]{\label{fig:real_setup0}
\includegraphics[height=60mm,width=.45\linewidth,keepaspectratio]{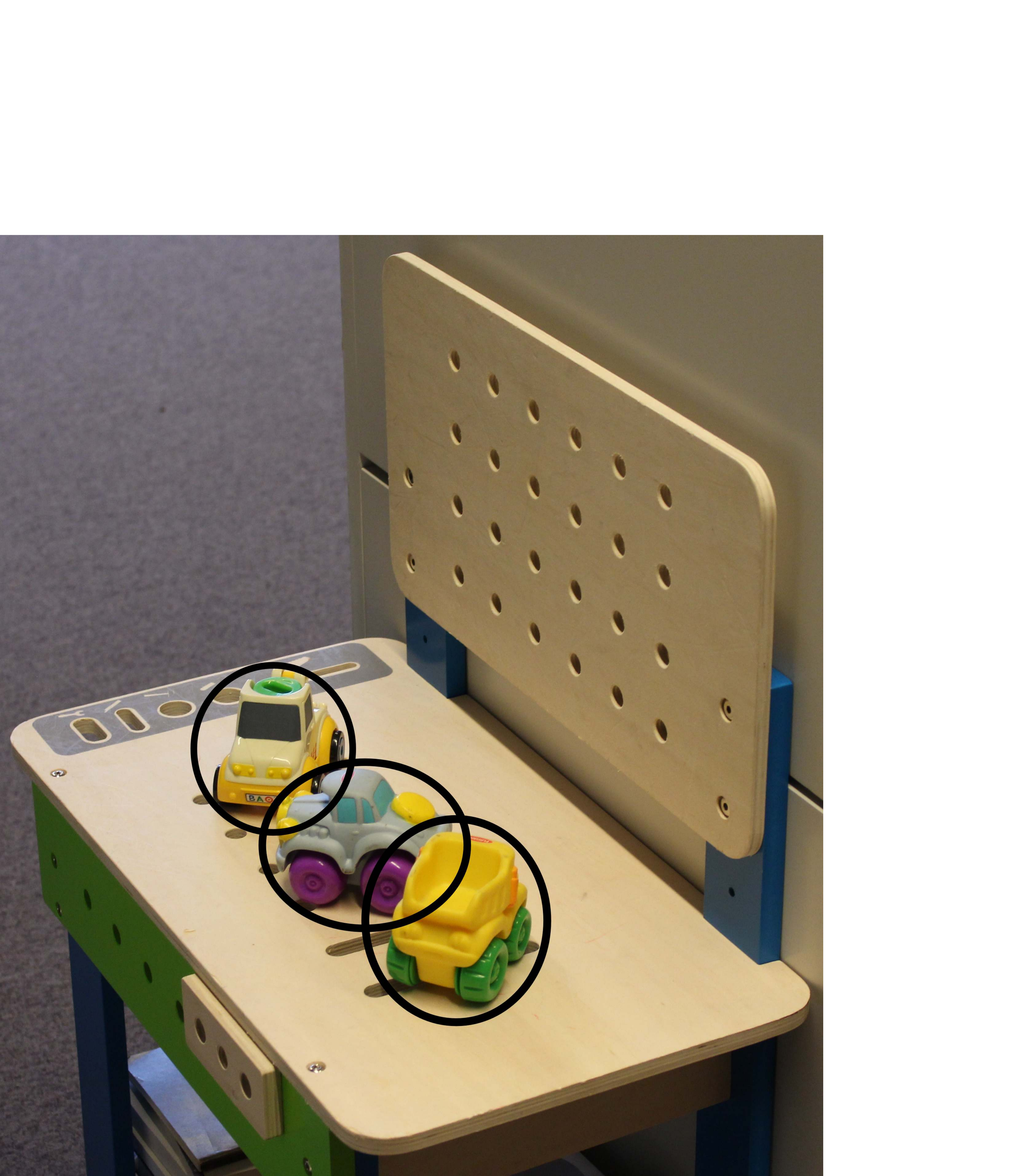}
} 
\subfloat[\textbf{Workbench2:} Toy workbench with fixed object on it and three moveable toy cars]{\label{fig:real_setup1}
\includegraphics[height=60mm,width=.45\linewidth,keepaspectratio]{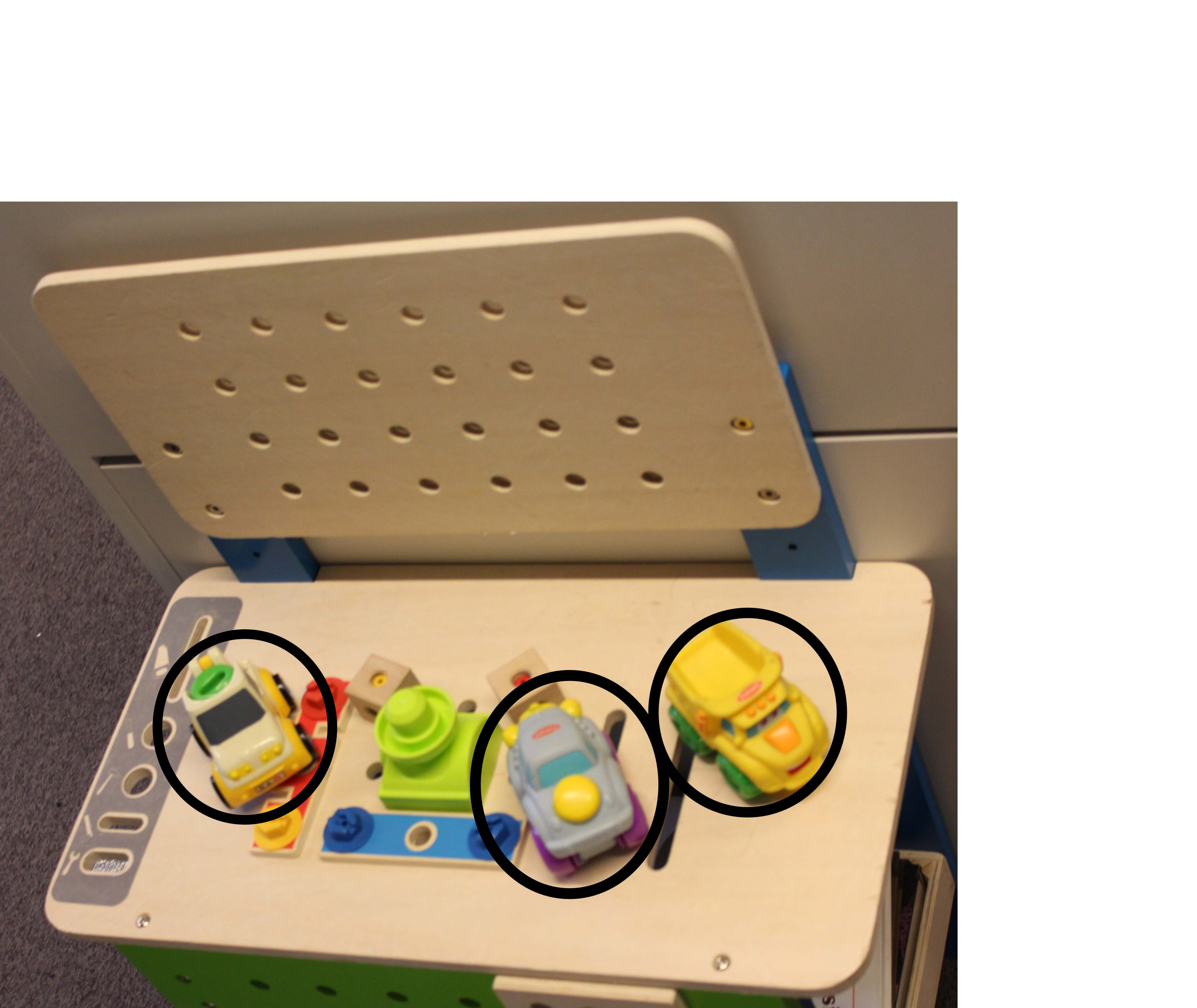}
} \\
\caption{Experimental setups with the real robot ordered by increasing difficulty. The moveable objects are marked by black circles.}
\label{fig:real_setups}
\end{figure}

\paragraph*{Real world setup}
Experiments are also conducted with a real robot to evaluate the method in a realistic scenario. The PR2 robot is used with a Kinect version 2 sensor. In figure \ref{fig:real_setups}, the setups used for the experiments are depicted. It is based on a modular workbench toy in two different configurations. These environments are colorful and have complex shapes that allow us to test the method on a complex and realistic setup. In these experiments, the robot interacts with the environment and the classifier learns from the label produces by these interactions.

\subsection{Classification Quality Measures}

\paragraph*{Precision, Recall, and Accuracy}

To measure the performance of the method, precision, recall, and accuracy are used. These are classical measures used in computer vision and more generally in classification tasks. In particular, these measures are used in most studies on SOD \citep{Borji2015}. The following equations define precision, recall, and accuracy, as used in this study:

\begin{equation}\label{eq:pra}
\begin{split}
precision & =  \frac{tp}{tp + fp} \\
recall & =  \frac{tp}{tp + fn} \\
accuracy & = \frac{1}{2}(\frac{tp}{G_{obj}} + \frac{tn}{G_{back}})
\end{split}
\end{equation}
Where $tp$ is the number of true positives and $tn$ is the number of true negatives (i.e. supervoxels well classified as part of moveable objects or as part of the background, respectively);  $fp$ are false positives, i.e. supervoxels misclassify as moveable, and $fn$ are false negatives, i.e. supervoxels misclassified as non-moveable; and $G_{obj}$ is the ground truth for parts of the environment that are objects and $G_{back}$ is the ground truth for parts of the environment that are fixed. Their definitions, for N supervoxels extracted on a pointcloud, is the following:

\begin{equation} \label{eq:tptn}
\begin{split}
tp & = \sum_i^N{P(L = 1 | W, \Theta, x_i)*(1 - \delta_i)} \\
tn & = \sum_i^N{P(L = 0 | W, \Theta, x_i)*\delta_i} \\
fp & = \sum_i^N{P(L = 1 | W, \Theta, x_i)*\delta_i} \\
fn & = \sum_i^N{P(L = 0 | W, \Theta, x_i)*(1 - \delta_i)} \\
G_{obj} & = \sum_i^N{1 - \delta_i} \\
G_{back} & = \sum_i^N{\delta_i}
\end{split}
\end{equation}
Where $\delta_i$ is the Kronecker symbol equal to $1$ if the i$^{th}$ supervoxel is part of the background, and otherwise equal to $0$; $x_i$ represents the features of the i$^{th}$ supervoxel.

\paragraph*{Measure of the Exploration Dynamic}
We also measure the number of samples gathered during the exploration of each category. This allows us to assess how the exploration is conducted according to the knowledge available. Also, the number of components is monitored to determine whether it increases with the complexity of an environment. For the experiments with the real robot, the number of mislabeled samples, which corresponds to failed interactions, is counted.

\section{Results}\label{sec:res}

\begin{figure}[h]
\centering
\subfloat[Plot for setup \ref{fig:setup0}]{\label{fig:praid0}
\includegraphics[height=60mm,width=.45\linewidth,keepaspectratio]{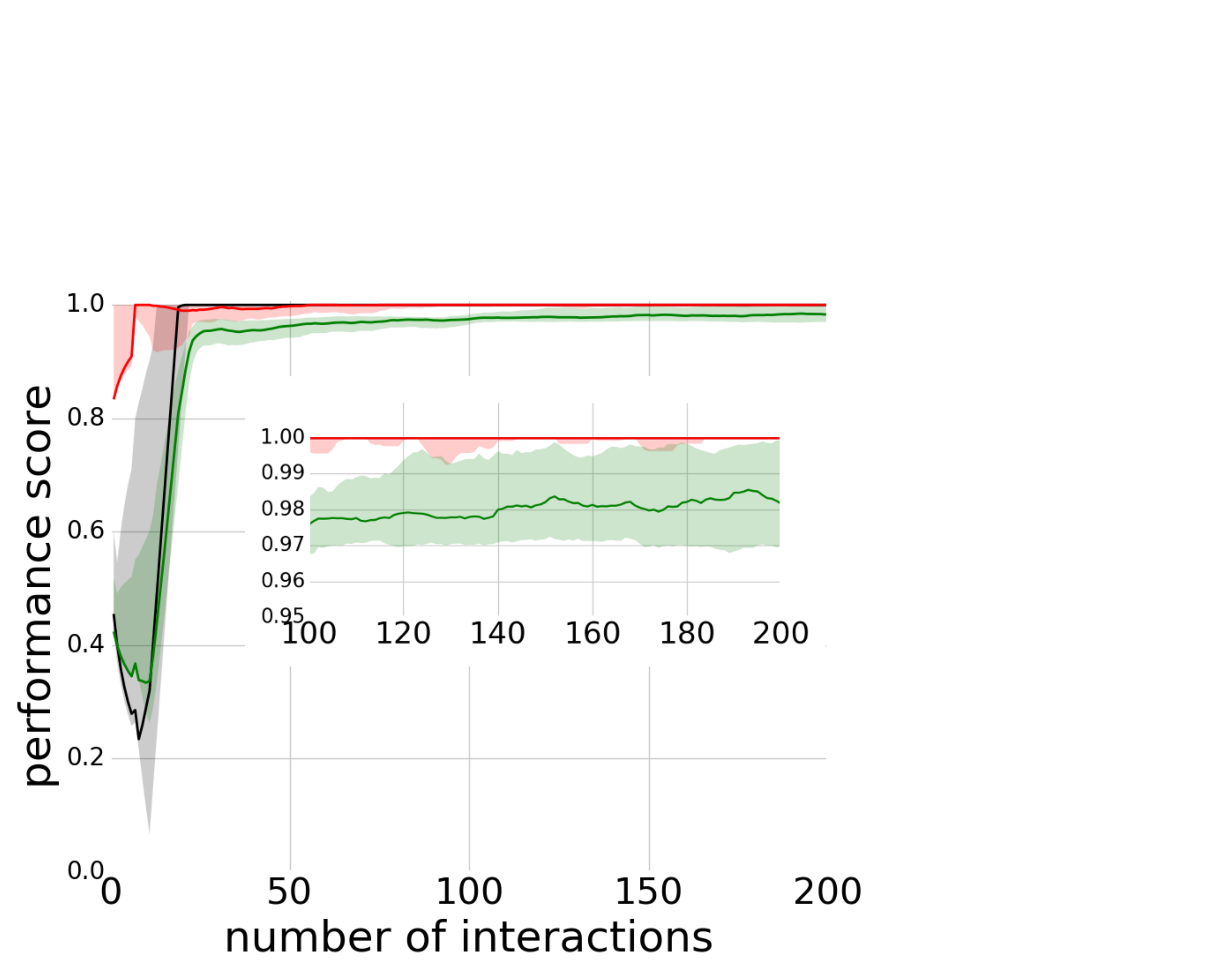}
} 
\subfloat[Plot for setup \ref{fig:setup1}]{\label{fig:praid1}
\includegraphics[height=60mm,width=.45\linewidth,keepaspectratio]{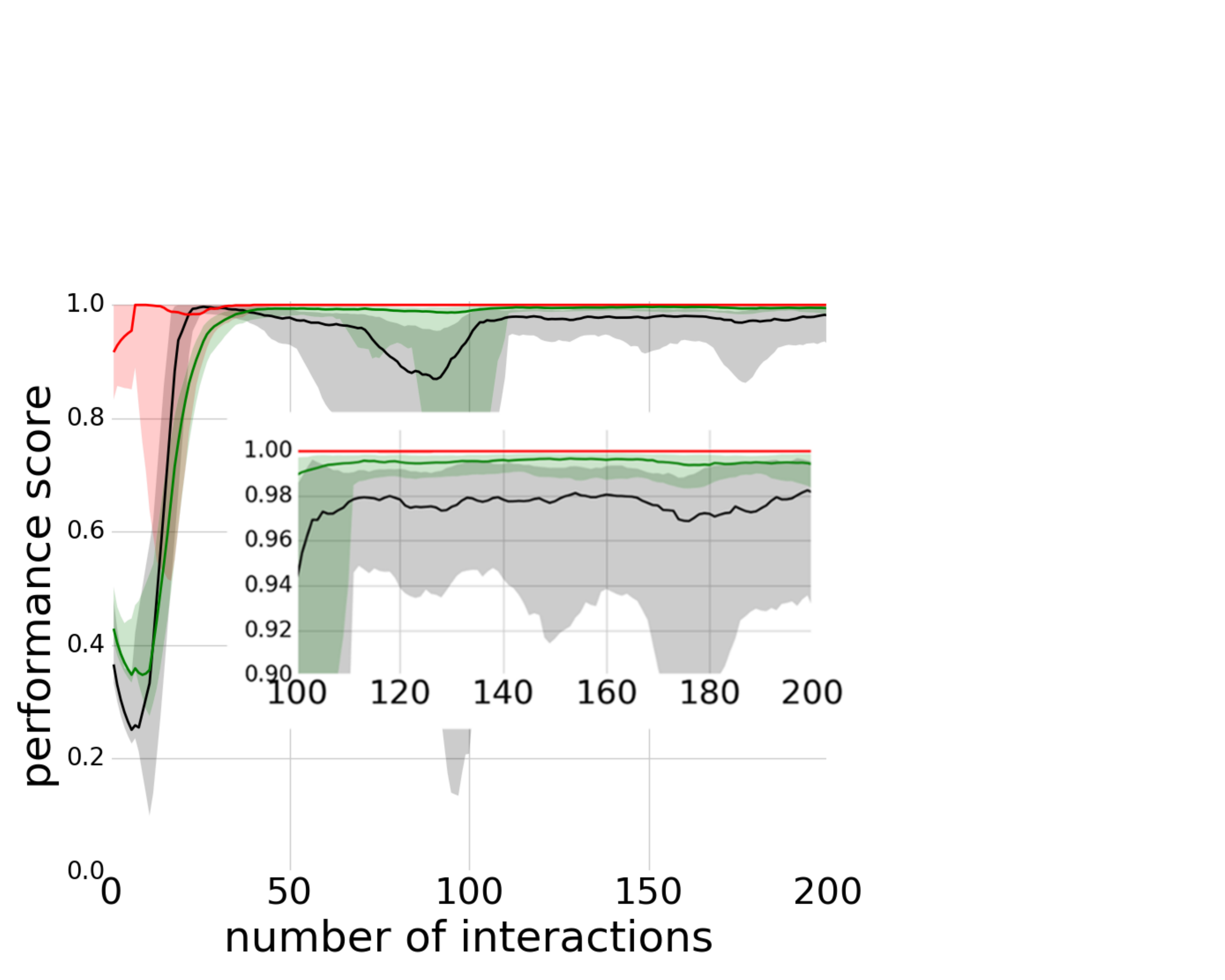}
}\\
\subfloat[Plot for setup \ref{fig:setup3}]{\label{fig:praid3}
\includegraphics[height=60mm,width=.45\linewidth,keepaspectratio]{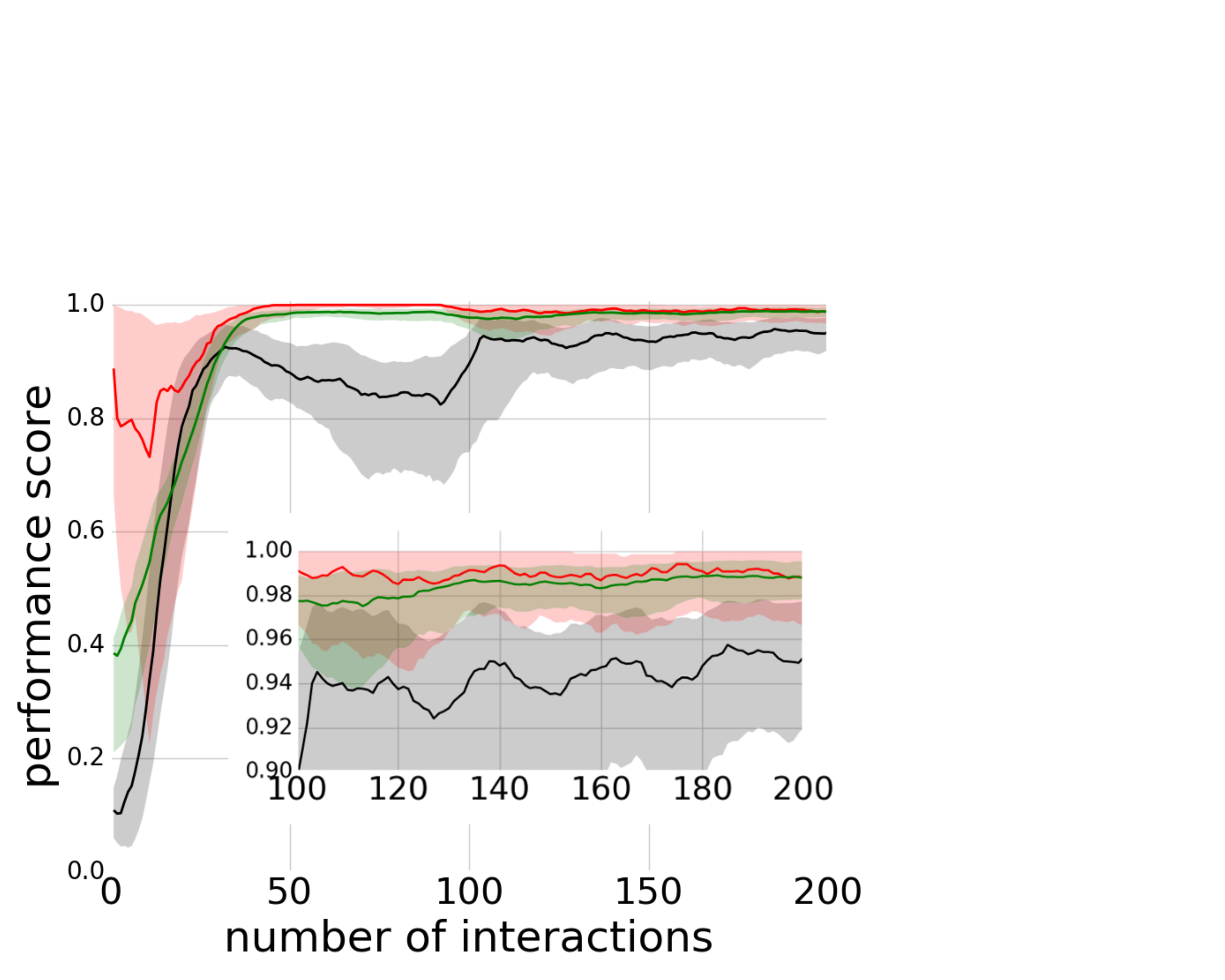}
} 
\subfloat[Plot for setup \ref{fig:setup4}]{\label{fig:praid4}
\includegraphics[height=60mm,width=.45\linewidth,keepaspectratio]{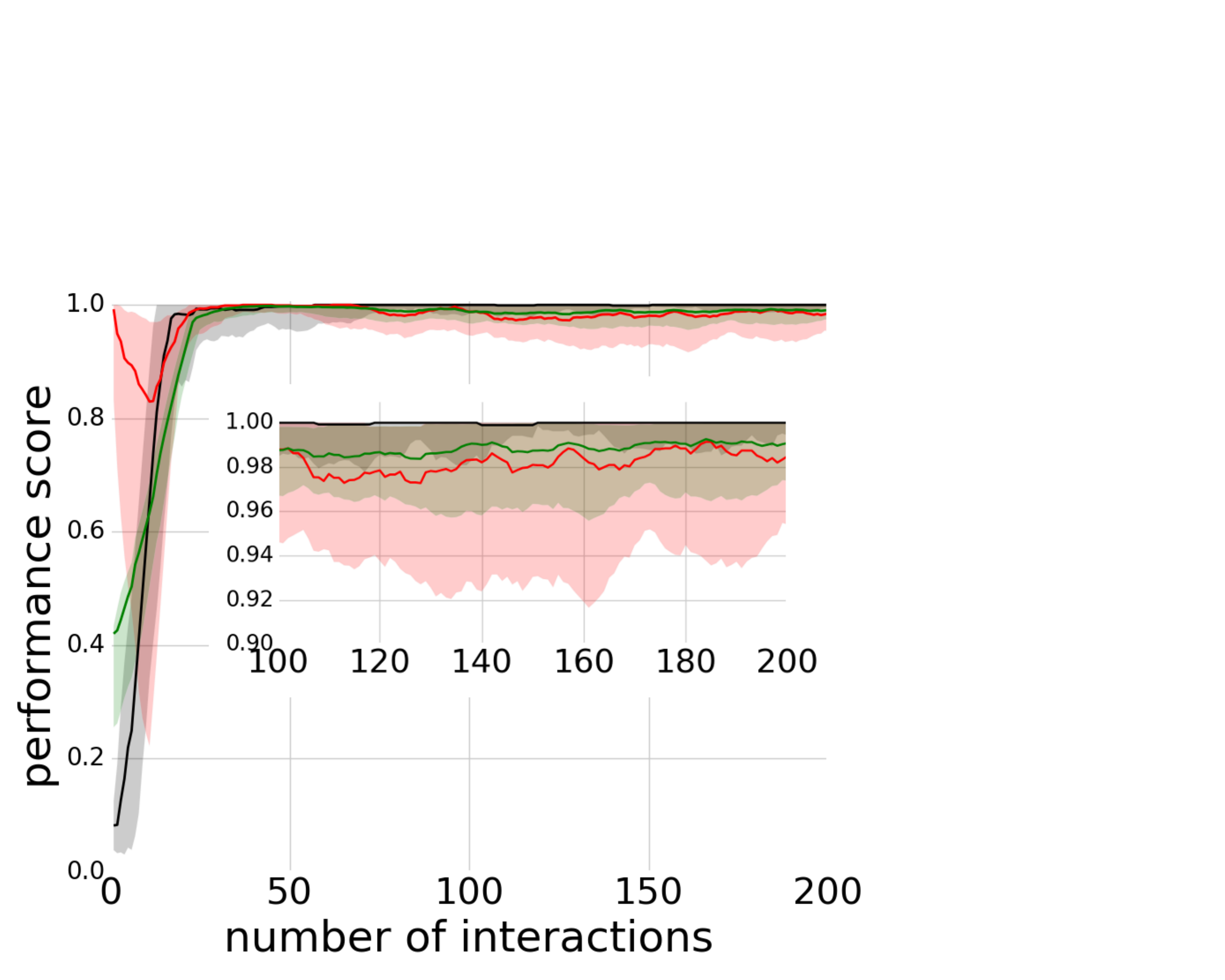}
} \\
\subfloat[Plot for setup \ref{fig:simkitchen}]{\label{fig:simkitchenpra}
\includegraphics[height=60mm,width=.9\linewidth,keepaspectratio]{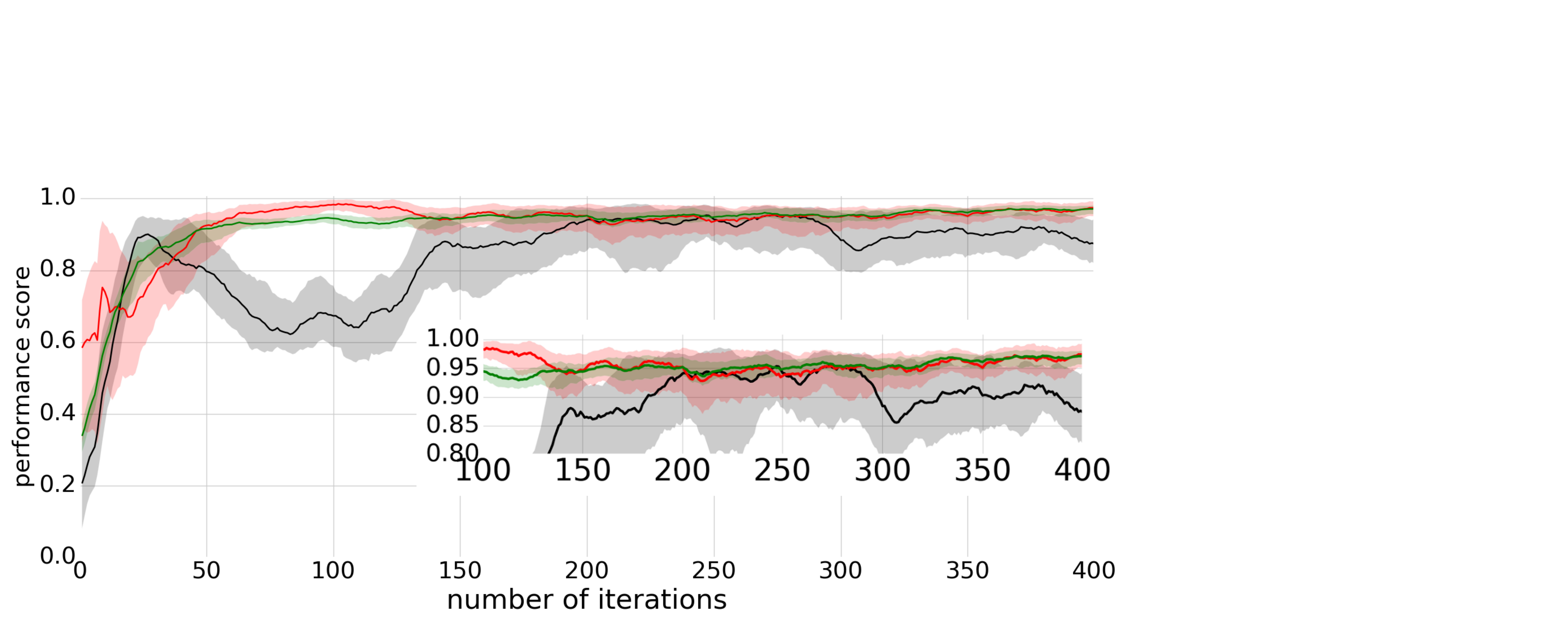}
} 
 \\
\subfloat[Plot for setup \ref{fig:setup5}]{\label{fig:praid5}
\includegraphics[height=60mm,width=.9\linewidth,keepaspectratio]{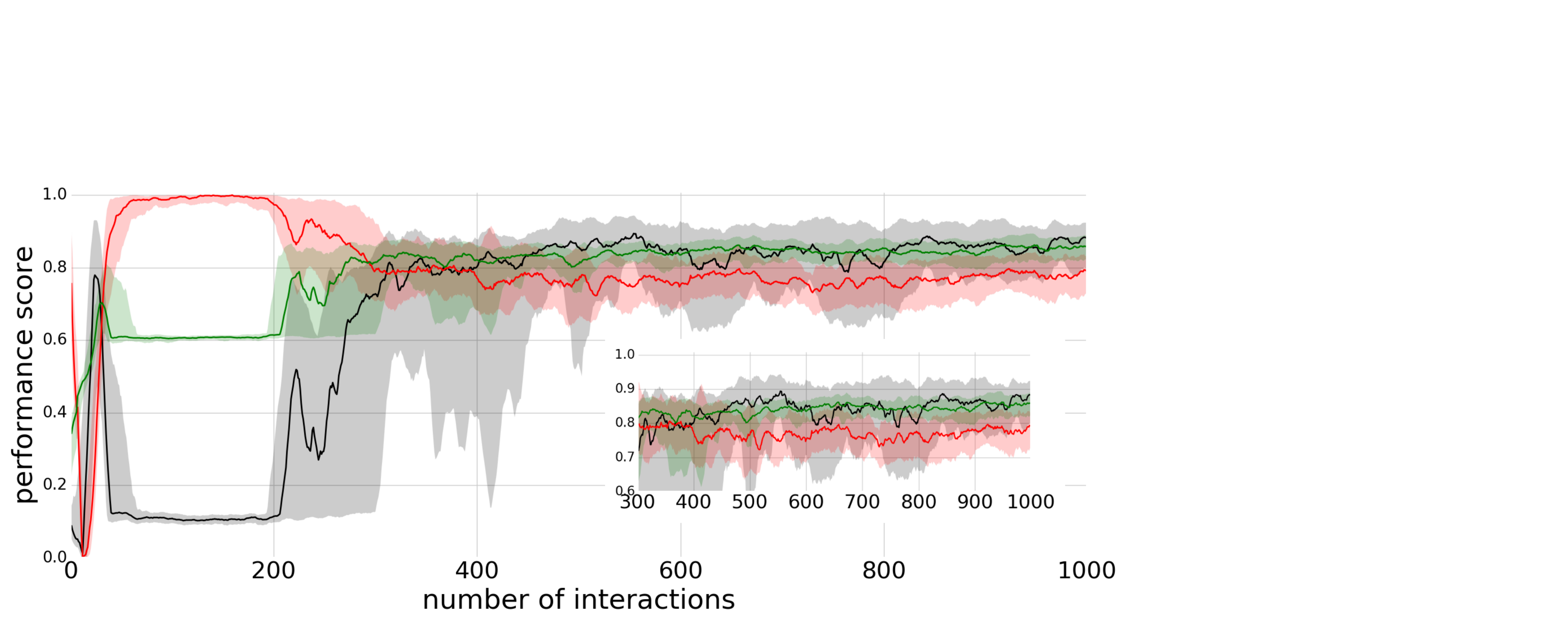}
}\\
\subfloat{
\includegraphics[height=60mm,width=.9\linewidth,keepaspectratio]{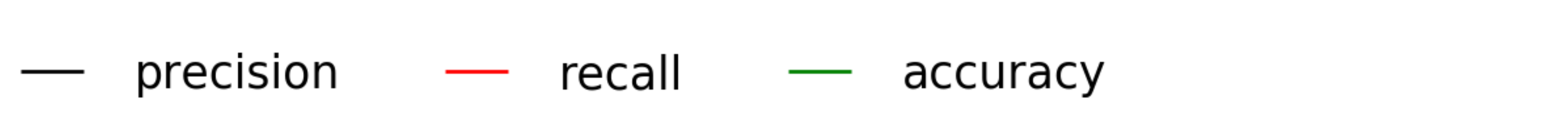}
} 
\caption{Plots of precision, recall and accuracy for each setup presented in figure \ref{fig:setups}}
\label{fig:praid}
\end{figure}

\subsection{Simplified setup}

\begin{figure}[h]
\centering
\subfloat[Plot for setup \ref{fig:setup0}]{\label{fig:posneg0}
\includegraphics[height=60mm,width=.45\linewidth,keepaspectratio]{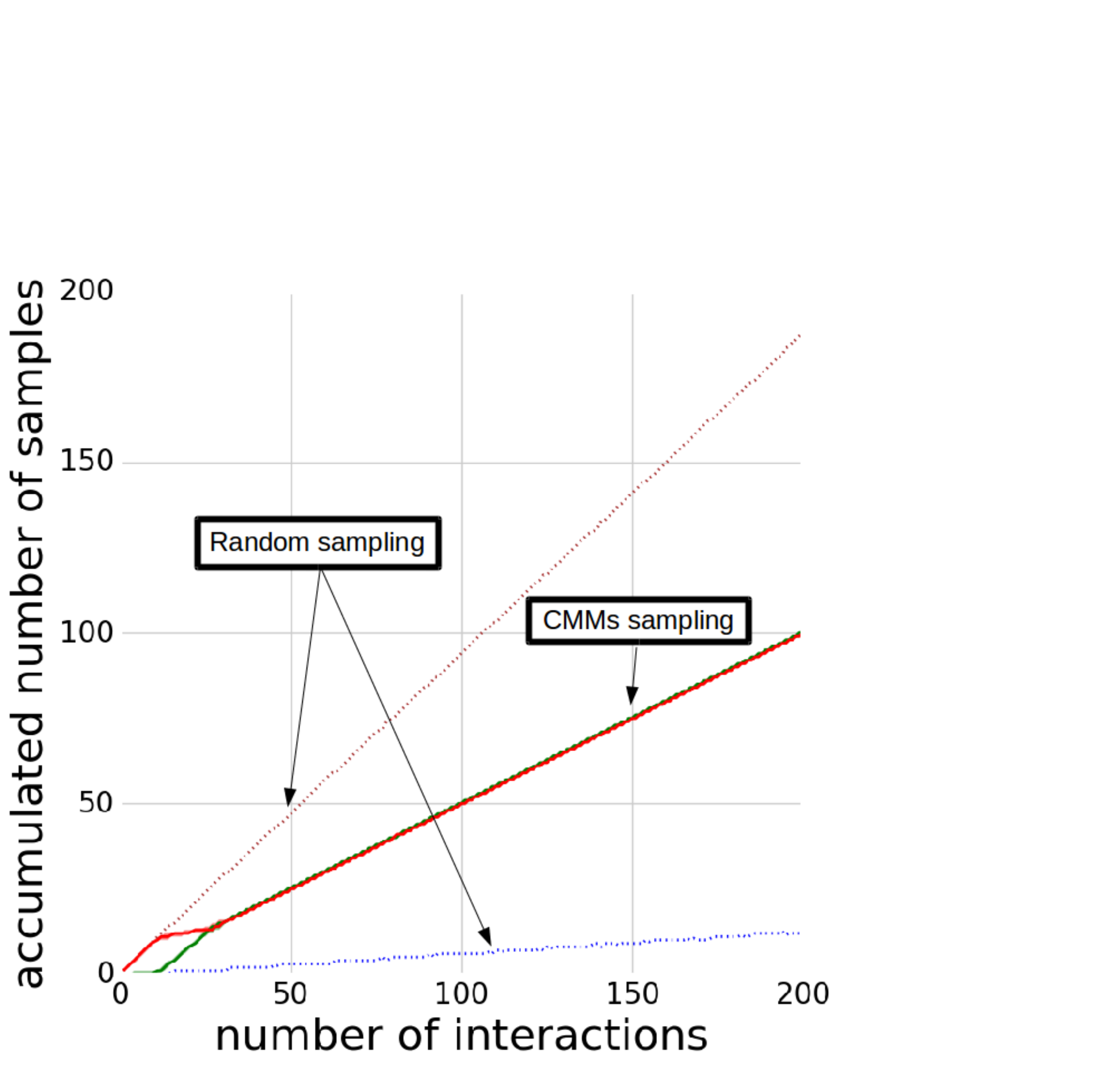}
} 
\subfloat[Plot for setup \ref{fig:setup1}]{\label{fig:posneg1}
\includegraphics[height=60mm,width=.45\linewidth,keepaspectratio]{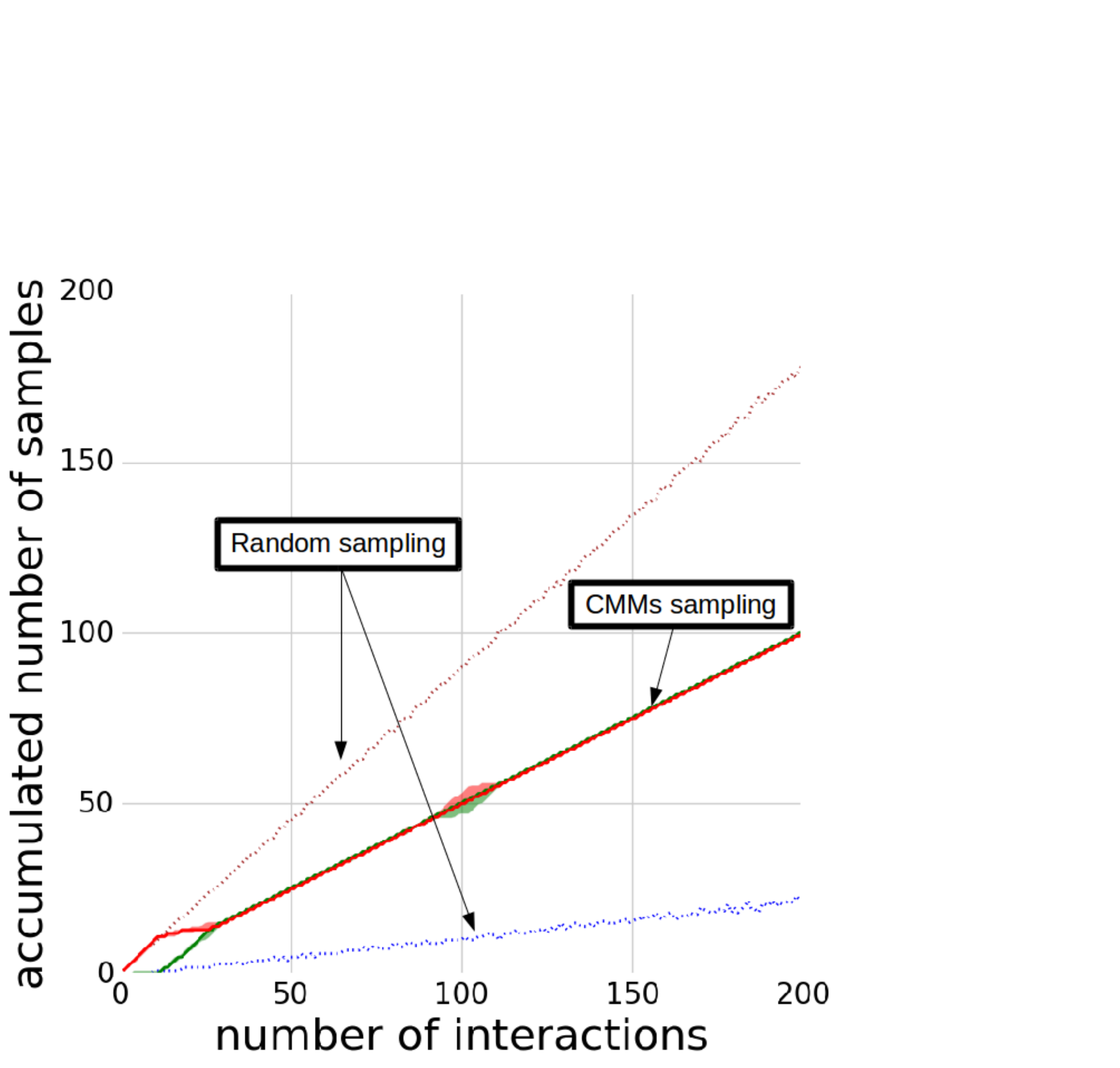}
}\\
\subfloat[Plot for setup \ref{fig:setup3}]{\label{fig:posneg3}
\includegraphics[height=60mm,width=.45\linewidth,keepaspectratio]{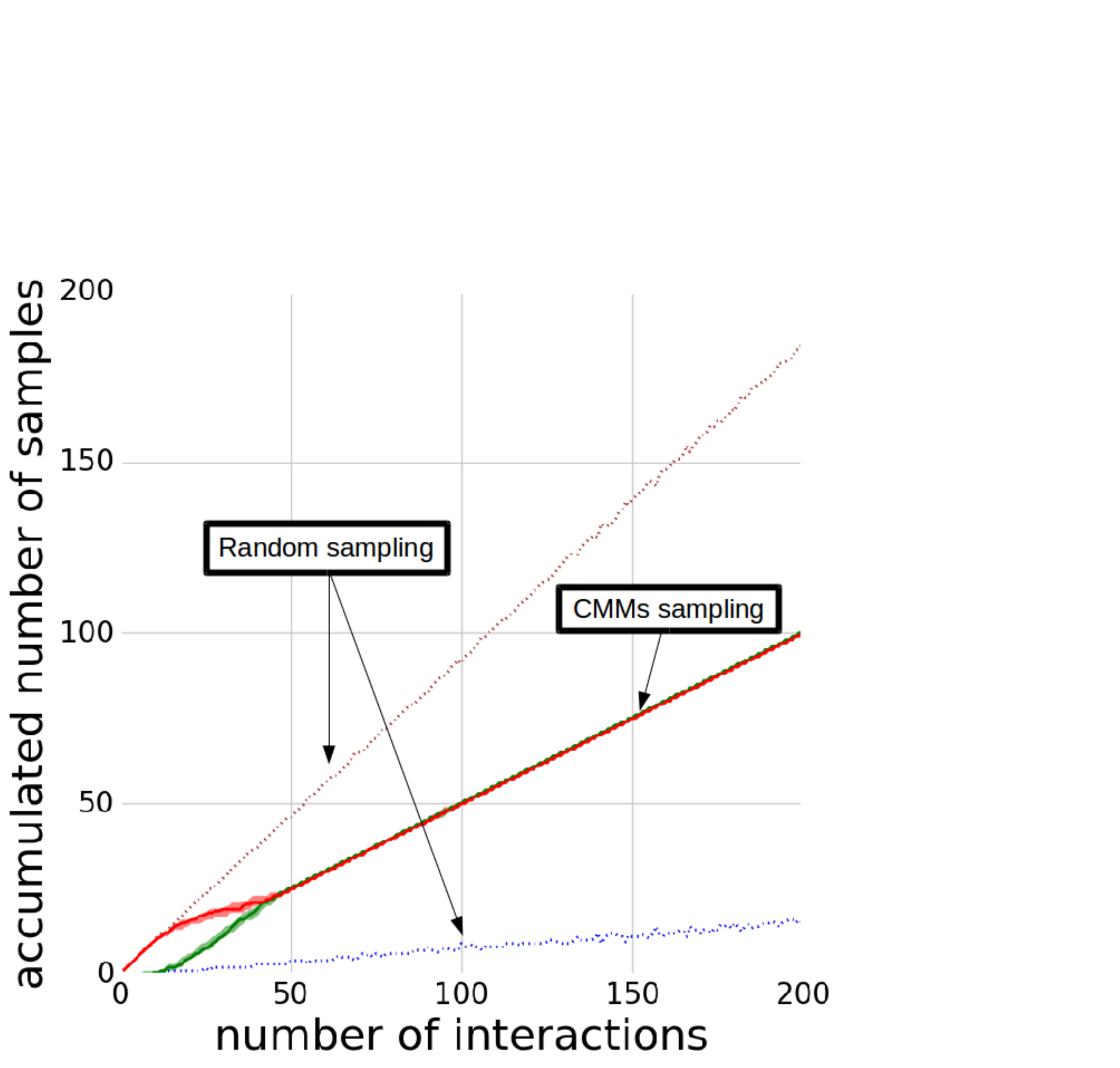}
} 
\subfloat[Plot for setup \ref{fig:setup4}]{\label{fig:posneg4}
\includegraphics[height=60mm,width=.45\linewidth,keepaspectratio]{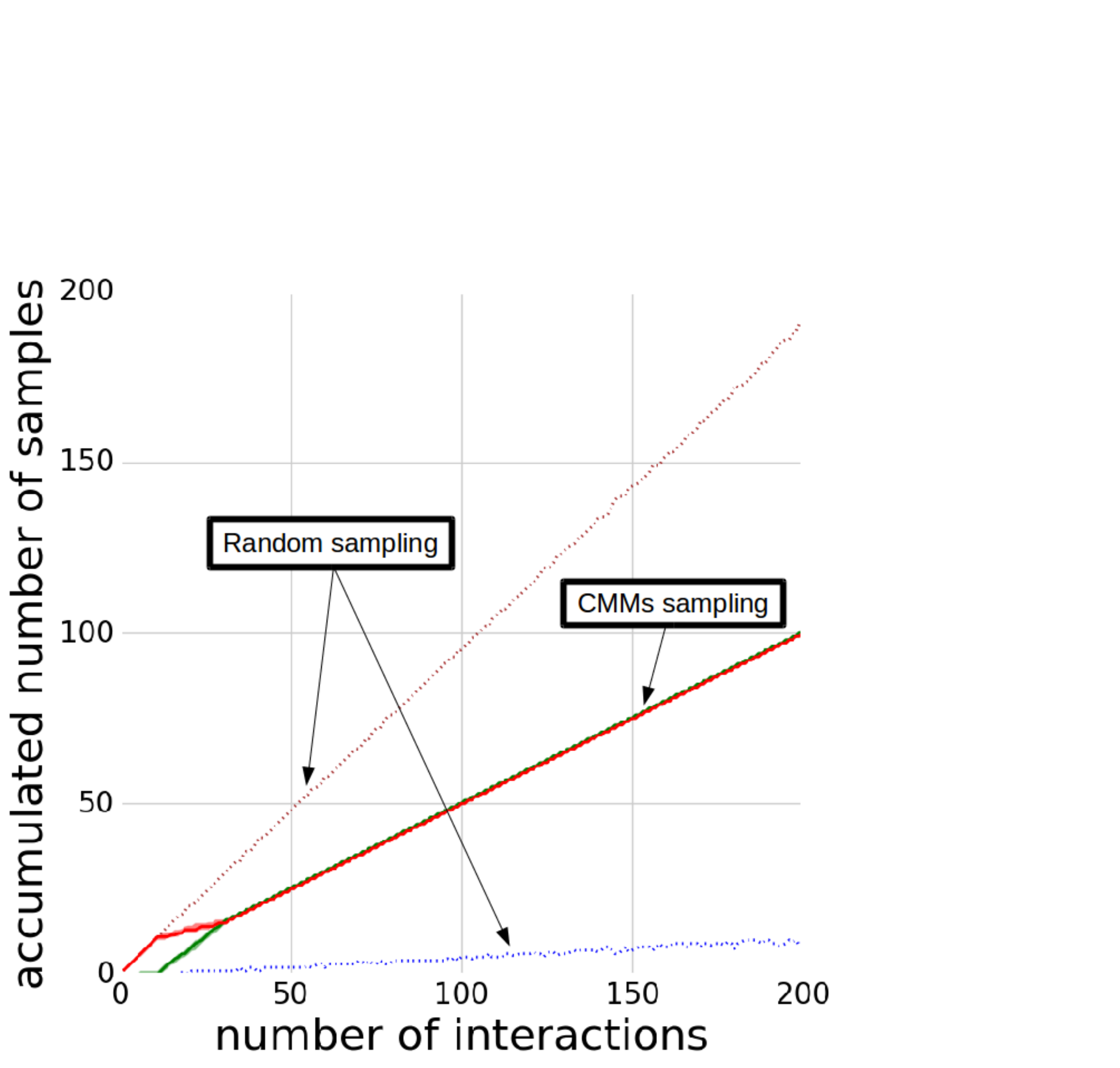}
} \\
\subfloat[Plot for setup \ref{fig:simkitchen}]{\label{fig:simkitchen_posneg}
\includegraphics[height=60mm,width=.45\linewidth,keepaspectratio]{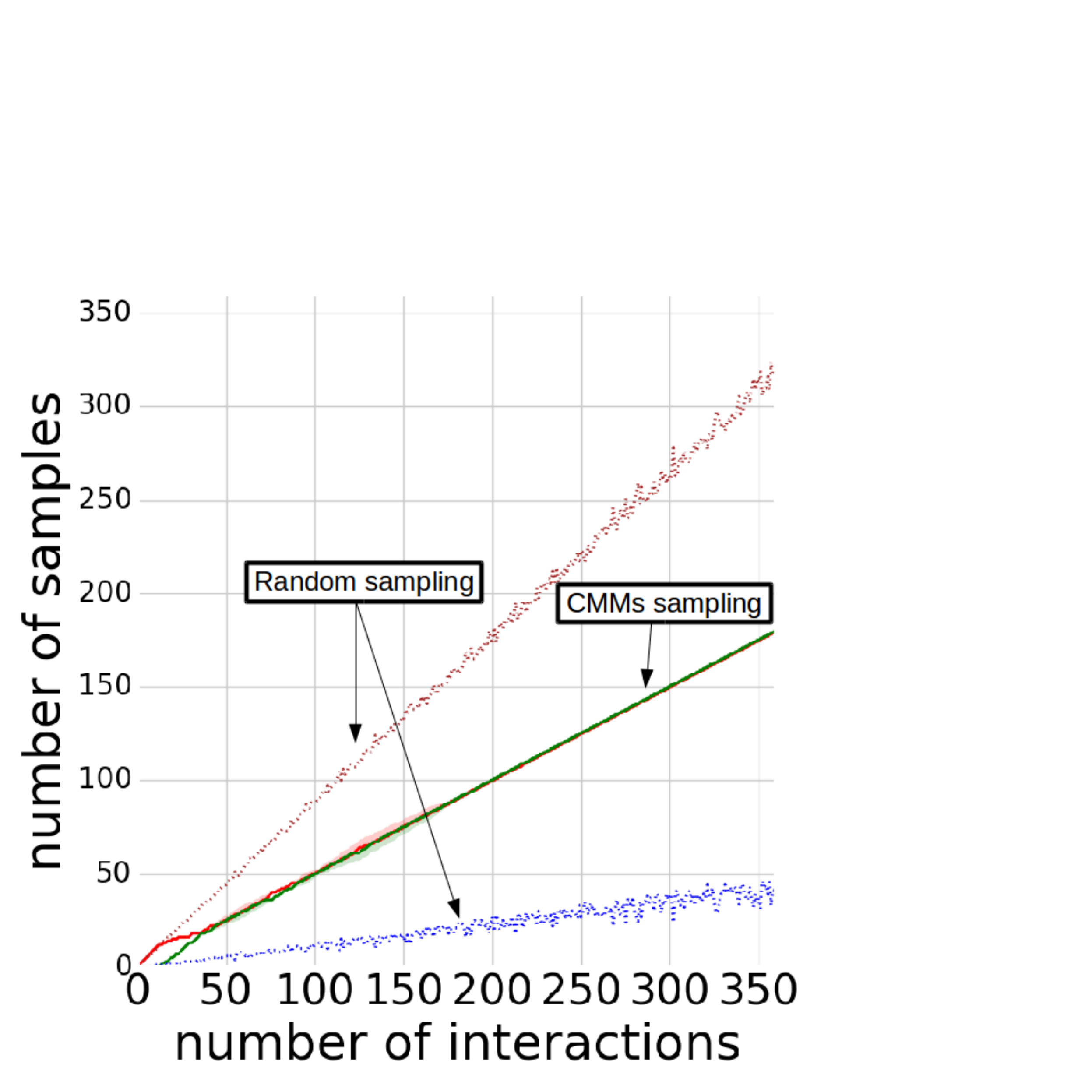}
}
\subfloat[Plot for setup \ref{fig:setup5}]{\label{fig:posneg5}
\includegraphics[height=60mm,width=.45\linewidth,keepaspectratio]{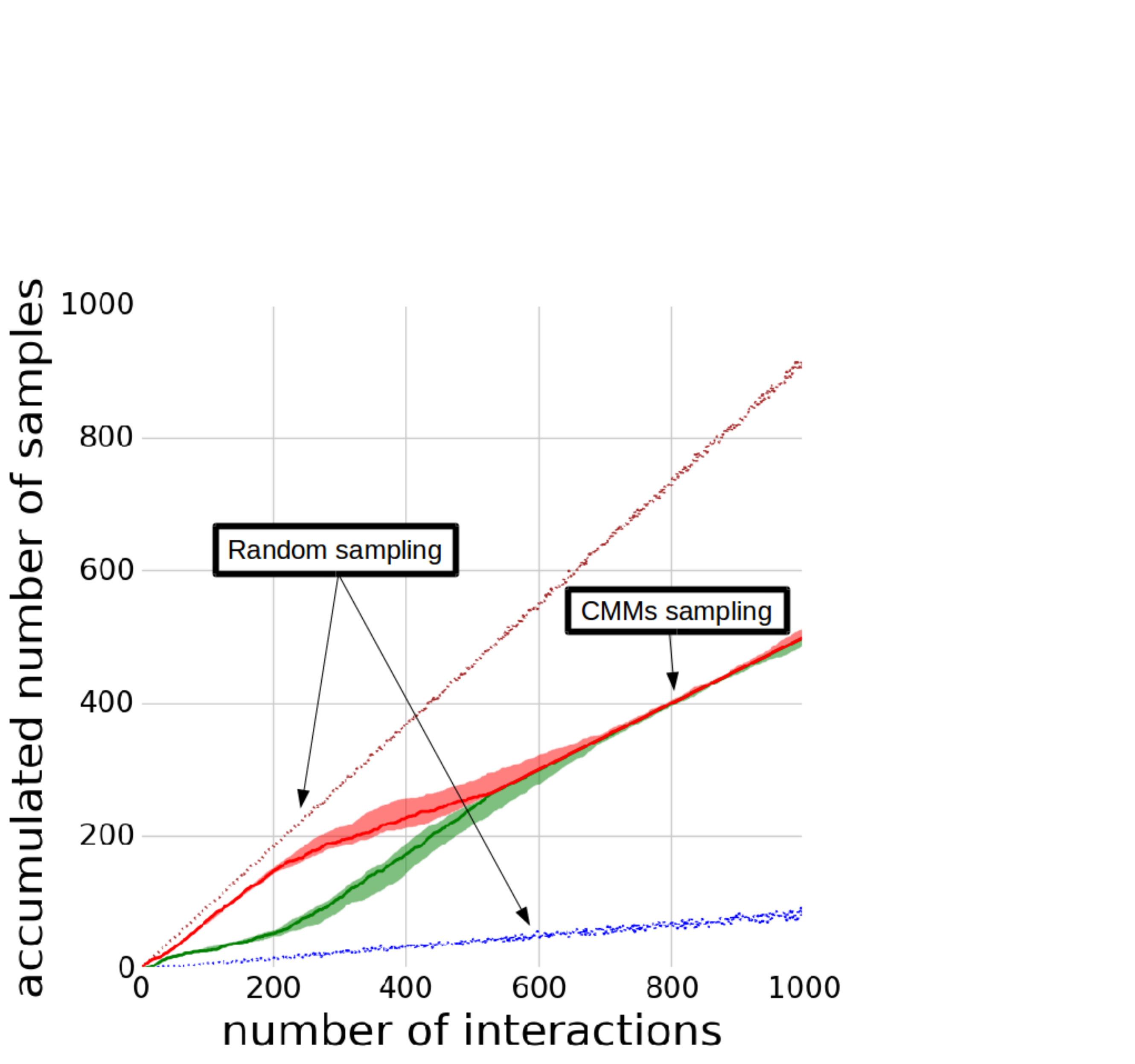}
}  \\
\subfloat{
\includegraphics[height=60mm,width=.9\linewidth,keepaspectratio]{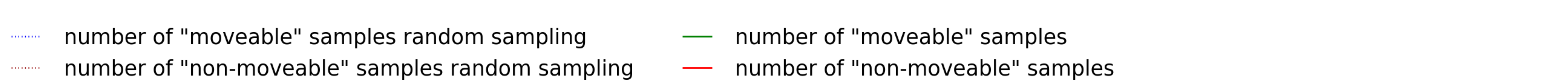}
}
\caption{Plots of the number of samples gathered for each class at each iteration during the experiments for each simplified setup presented in figure \ref{fig:setups}.}
\label{fig:posneg}
\end{figure}

As expected, the classification reaches scores of almost 1 for \textbf{MoveableBalls1} setup (\ref{fig:setup0})(see figure \ref{fig:praid0}). There is nothing complex about this setup as the moveable objects share no features with the background. However, in \textbf{MoveableBricks1} (\ref{fig:setup1}) and \textbf{MoveableBricks2} (\ref{fig:setup3}) setups, the performance does not directly reach maximum performances as shown in figures \ref{fig:praid1} and \ref{fig:praid3}. This is due to the similarities between the table and the cubes in terms of color and shape. As the classifier takes more time to converge, the sampling process is less efficient at choosing suitable samples at each iteration, therefore, the system takes more time to gather a representative dataset as shown in figure \ref{fig:posneg}.

A minimum number of samples is needed to start splitting components because of the constraint introduced by the component intersection criterion (see equation \ref{eq:intercond}) which explains the decrease in performance until iteration 100.

\begin{figure}[h]
\centering
\subfloat[Plot for setup \ref{fig:setup1}]{\label{fig:praid_fix_comp0}
\includegraphics[height=60mm,width=.45\linewidth,keepaspectratio]{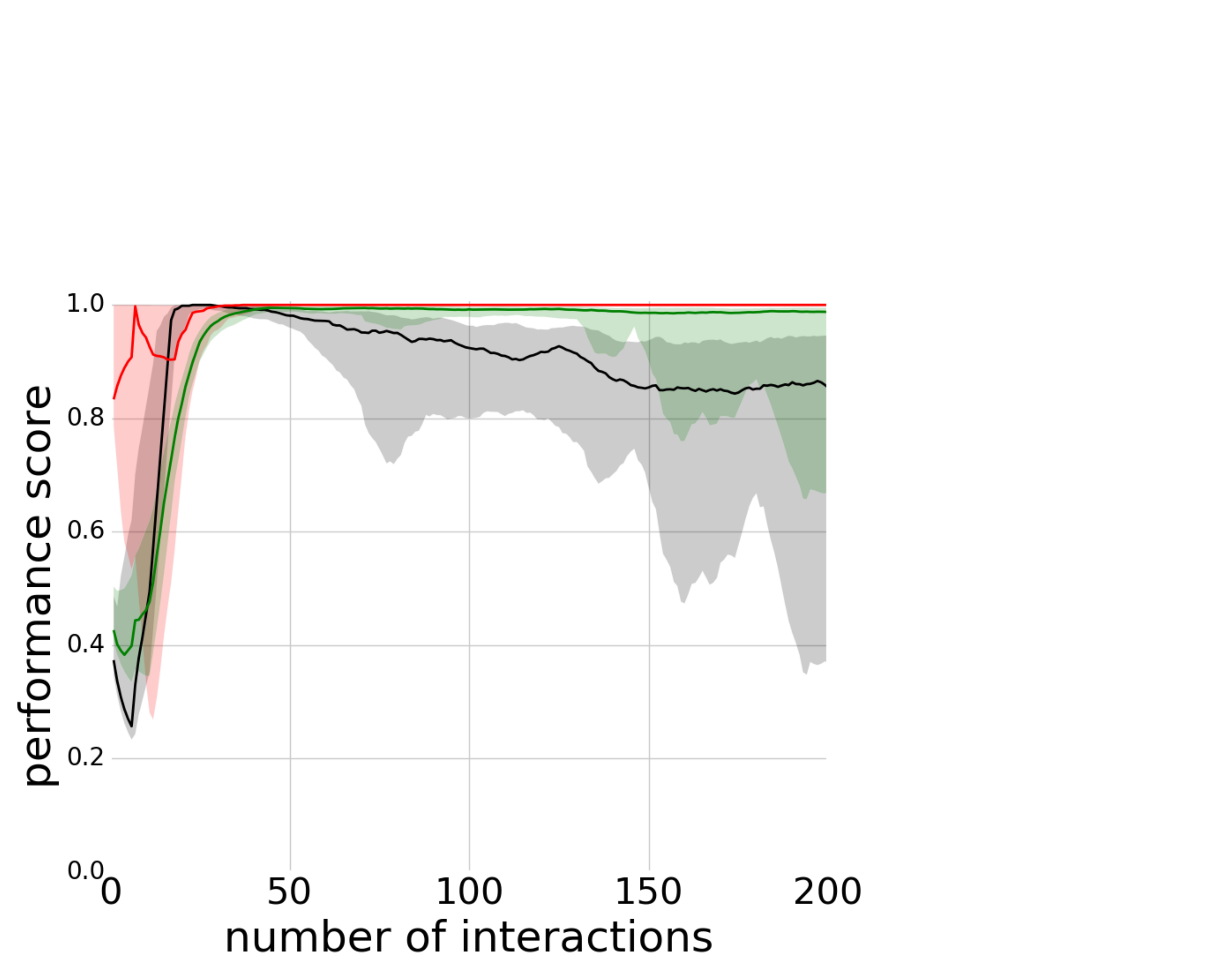}
} 
\subfloat[Plot for setup \ref{fig:setup3}]{\label{fig:praid_fix_comp1}
\includegraphics[height=60mm,width=.45\linewidth,keepaspectratio]{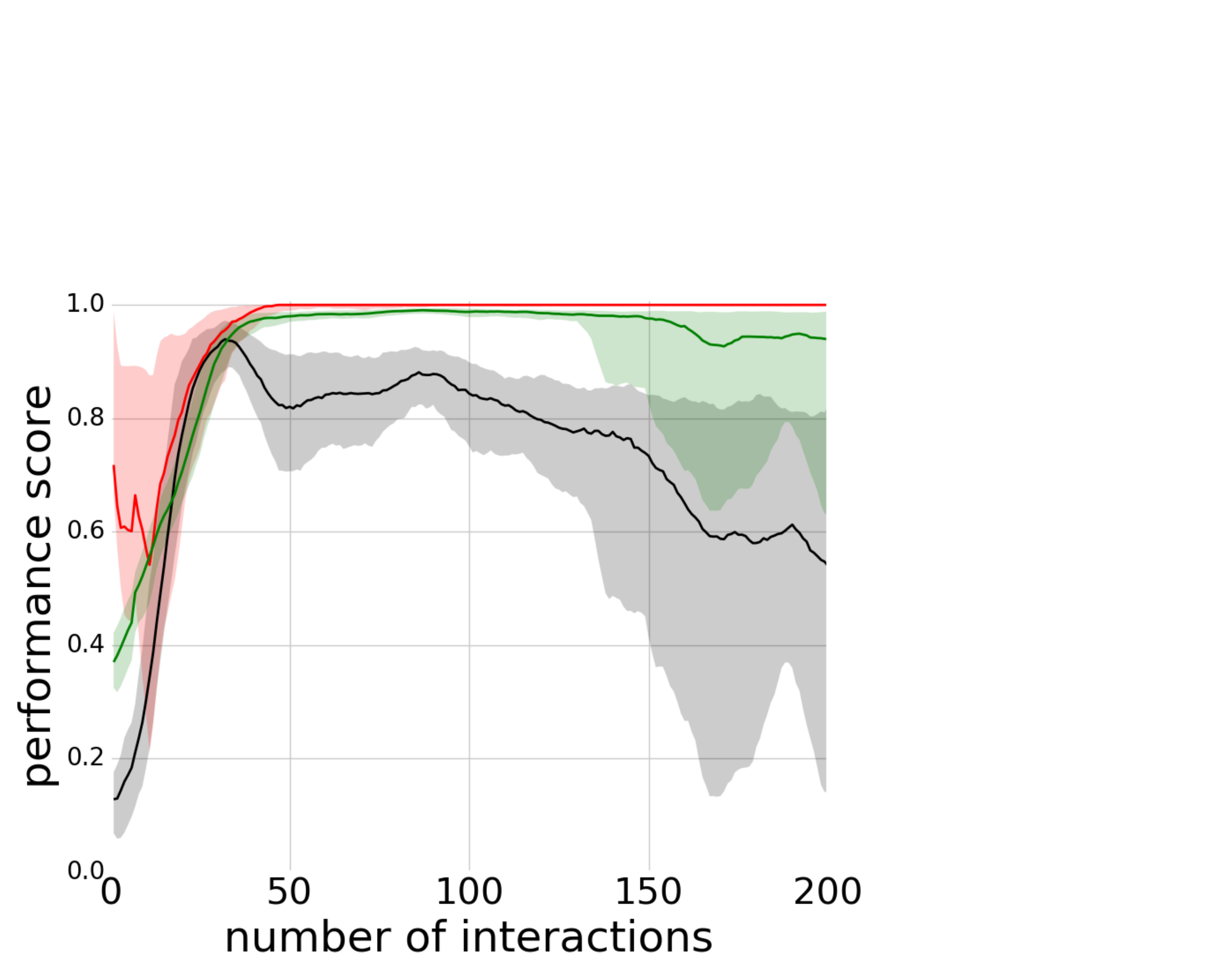}
}\\
\subfloat[Plot for setup \ref{fig:setup5}]{\label{fig:praid_fix_comp2}
\includegraphics[height=60mm,width=.9\linewidth,keepaspectratio]{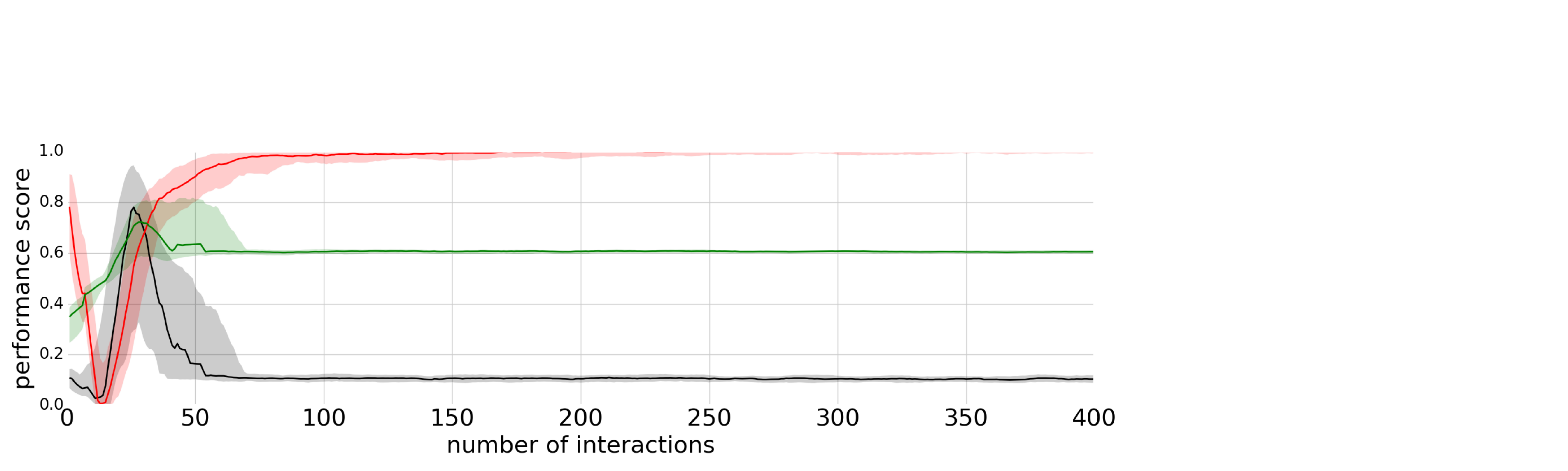}
} \\
\subfloat{
\includegraphics[height=60mm,width=.9\linewidth,keepaspectratio]{legend_pra.pdf}
} 
\caption{Plots of precision, recall, and accuracy for setups \ref{fig:setup1}, \ref{fig:setup3} and \ref{fig:setup5} for experiments conducted with one component per model. In these experiments, no split or merge operations were applied. These experiments are made to control the contribution of split and merge operations.} 
\label{fig:praid_fix_comp}
\end{figure}

The shape feature is sufficient for setup \textbf{WhiteMoveableBalls} (\ref{fig:setup4}) as the classification reaches scores of almost 1 (see figure \ref{fig:praid4}). For setup \textbf{WhiteMoveableBricks}(\ref{fig:setup5}), the classification does not reach maximum performance (see figure \ref{fig:praid5}), but this is expected as the background and moveable objects have similar shapes. Even in this setup, the accuracy converges to a value above 0.8 and this seems sufficient to perform a relevant segmentation, as can be seen in figure \ref{fig:final_rm0} which depicts a relevance map obtained using setup \textbf{WhiteMoveableBricks}(\ref{fig:setup5}). The segmentation is not perfect, but it is accurate enough to identify object hypotheses that can be validated in the next step of the robot developmental process (see section \ref{sec:ip}). 
On setup \textbf{SimKitchen} (\ref{fig:simkitchen}), the performance reaches a value above 0.8 after 150 interactions (see figure \ref{fig:simkitchenpra}). It is better than the results for setup \textbf{WhiteMoveableBricks}, probably because the simulated kitchen is richer in shapes and colors.

The selection process is far from a uniform random sampling, which would explore a majority of samples in the background. The comparison of the numbers of samples in each class can be considered as a convergence criterion: if the dataset converges to the same number of samples in both classes, it means that  the classifier is able to distinguish the two classes with sufficient precision. As shown in figure \ref{fig:posneg}, all explorations reach the same number of samples in each class with no variability over the replications, indicating the stability of the selection process.

According to these results, setup \textbf{MoveableBricks2}  (\ref{fig:setup3}) is actually harder for the classifier to deal with than setup \textbf{WhiteMoveableBalls} (\ref{fig:setup4}).

\begin{figure}[h]
\centering
\subfloat[Plot for setup \ref{fig:setup0}]{\label{fig:nbcomp0}
\includegraphics[height=60mm,width=.45\linewidth,keepaspectratio]{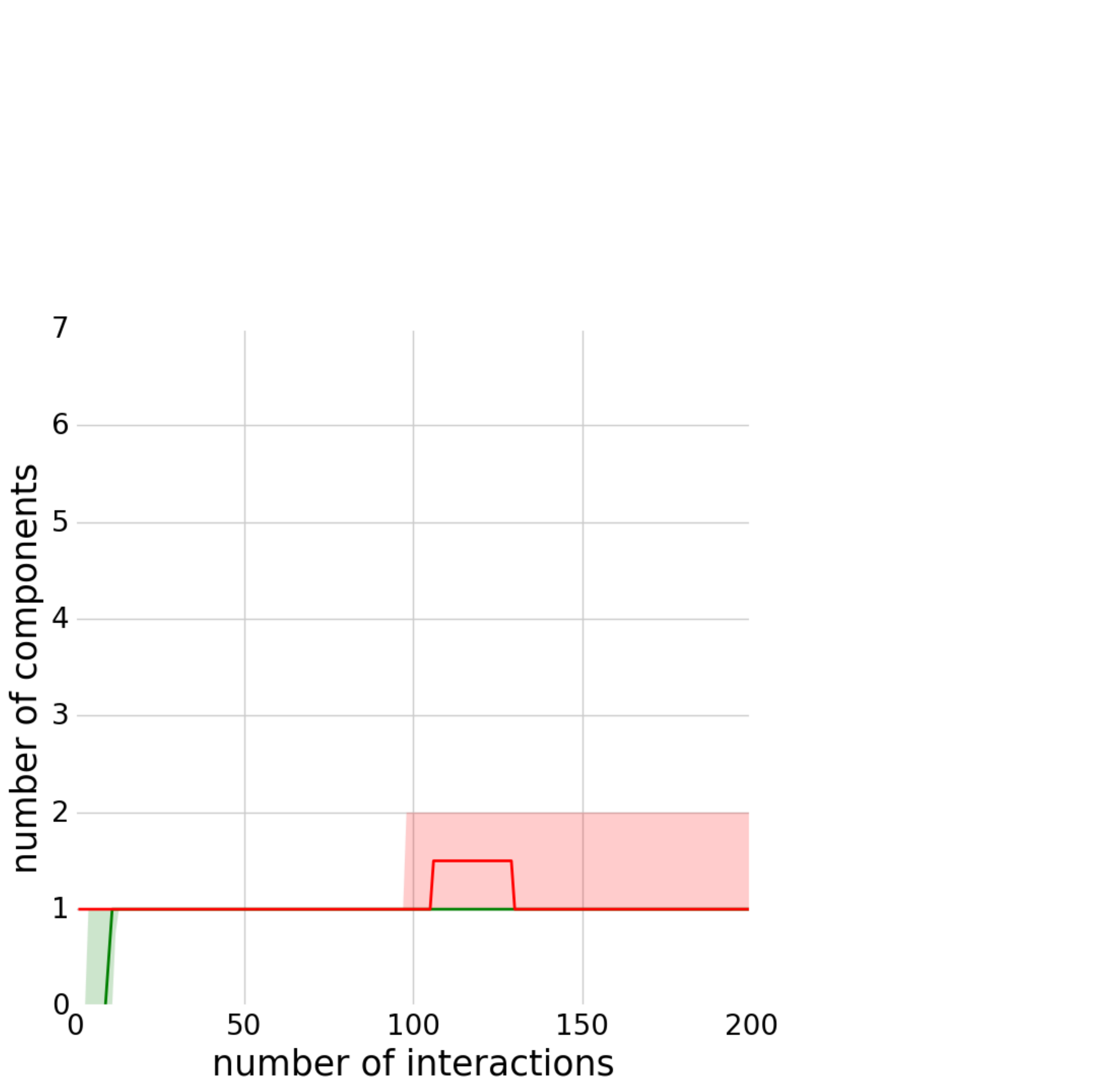}
} 
\subfloat[Plot for setup \ref{fig:setup1}]{\label{fig:nbcomp1}
\includegraphics[height=60mm,width=.45\linewidth,keepaspectratio]{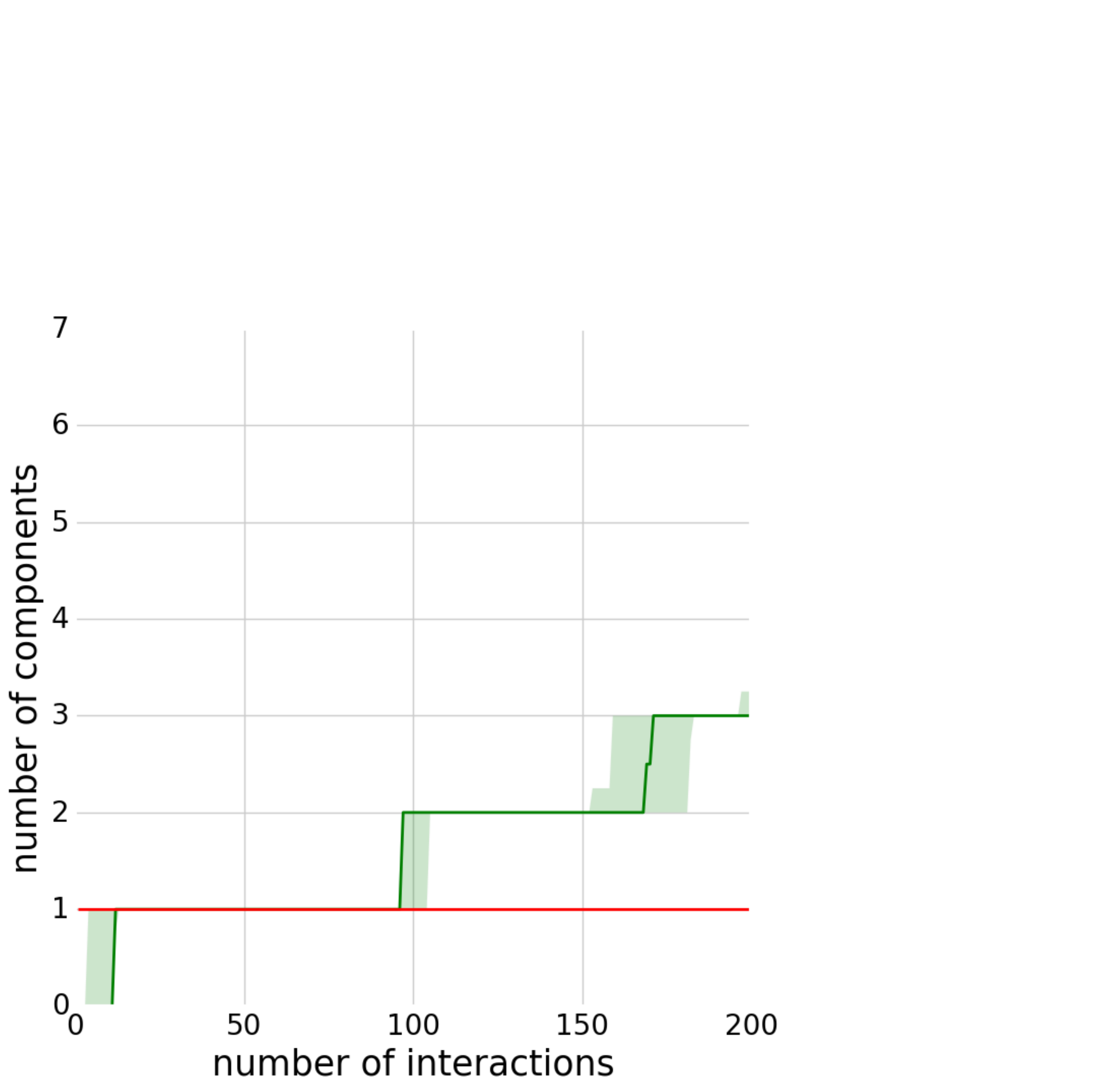}
}\\
\subfloat[Plot for setup \ref{fig:setup3}]{\label{fig:nbcomp3}
\includegraphics[height=60mm,width=.45\linewidth,keepaspectratio]{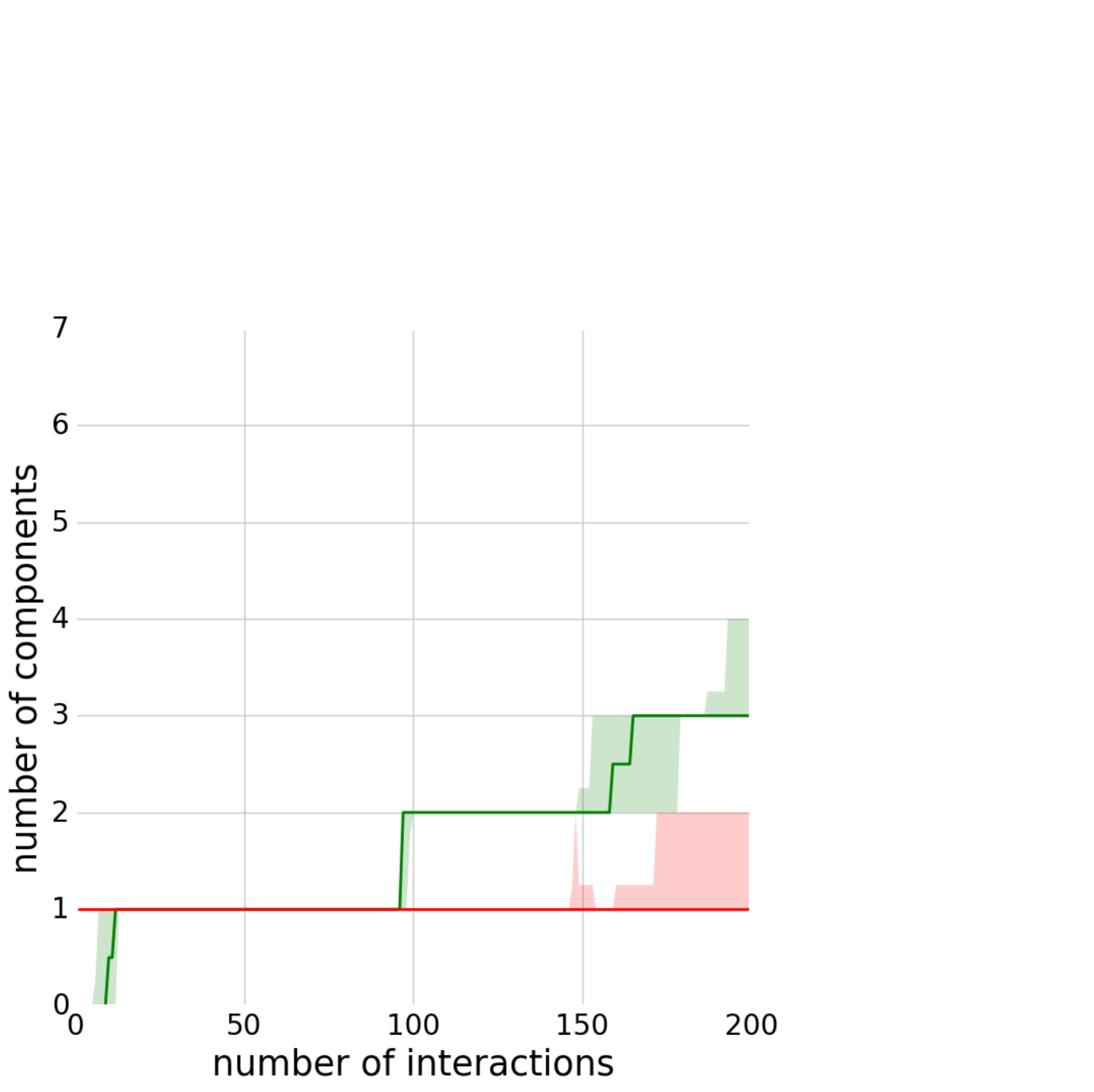}
} 
\subfloat[Plot for setup \ref{fig:setup4}]{\label{fig:nbcomp4}
\includegraphics[height=60mm,width=.45\linewidth,keepaspectratio]{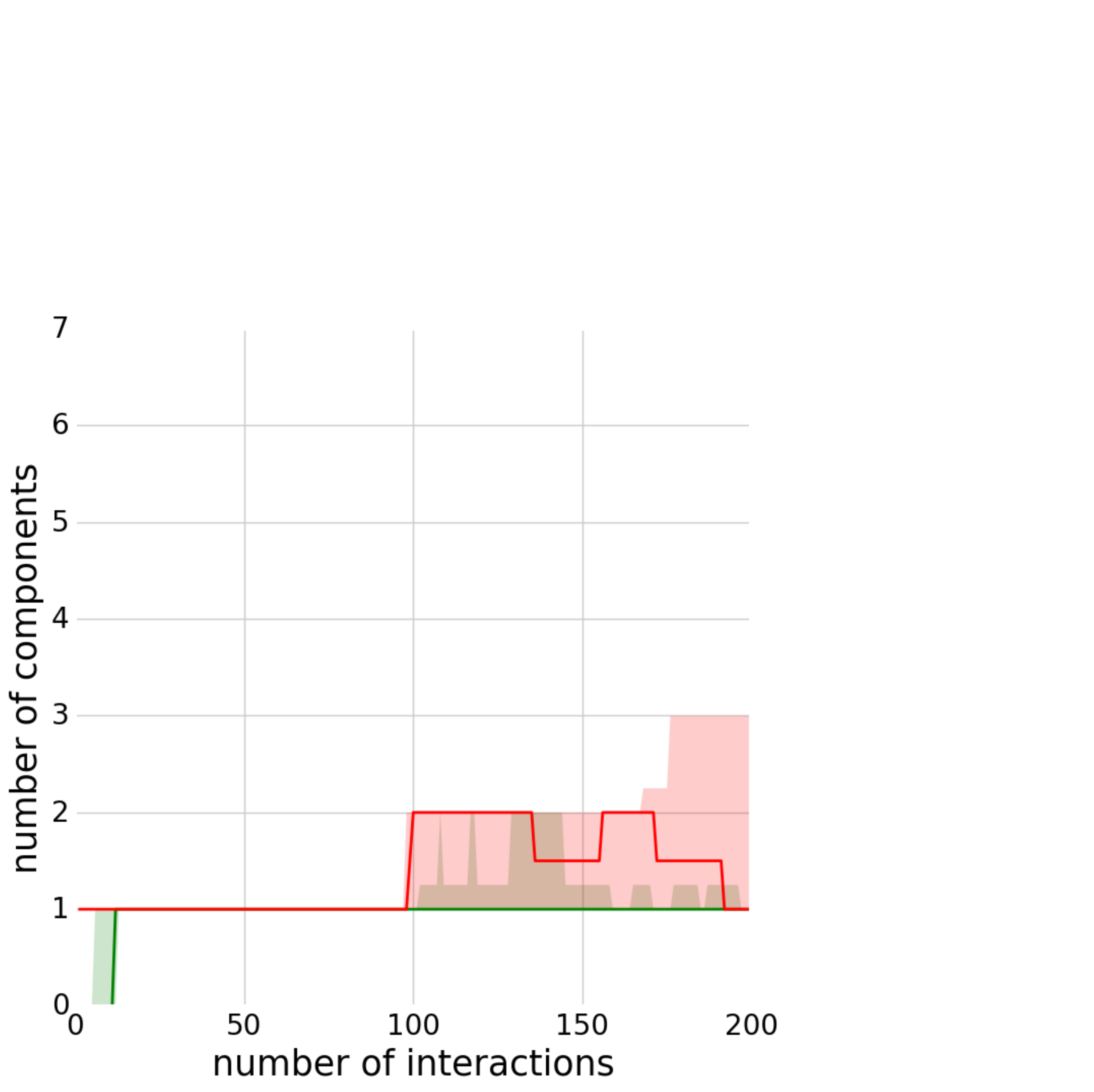}
}\\
\subfloat[Plot for setup \ref{fig:simkitchen}]{\label{fig:kitchencomp}
\includegraphics[height=60mm,width=.45\linewidth,keepaspectratio]{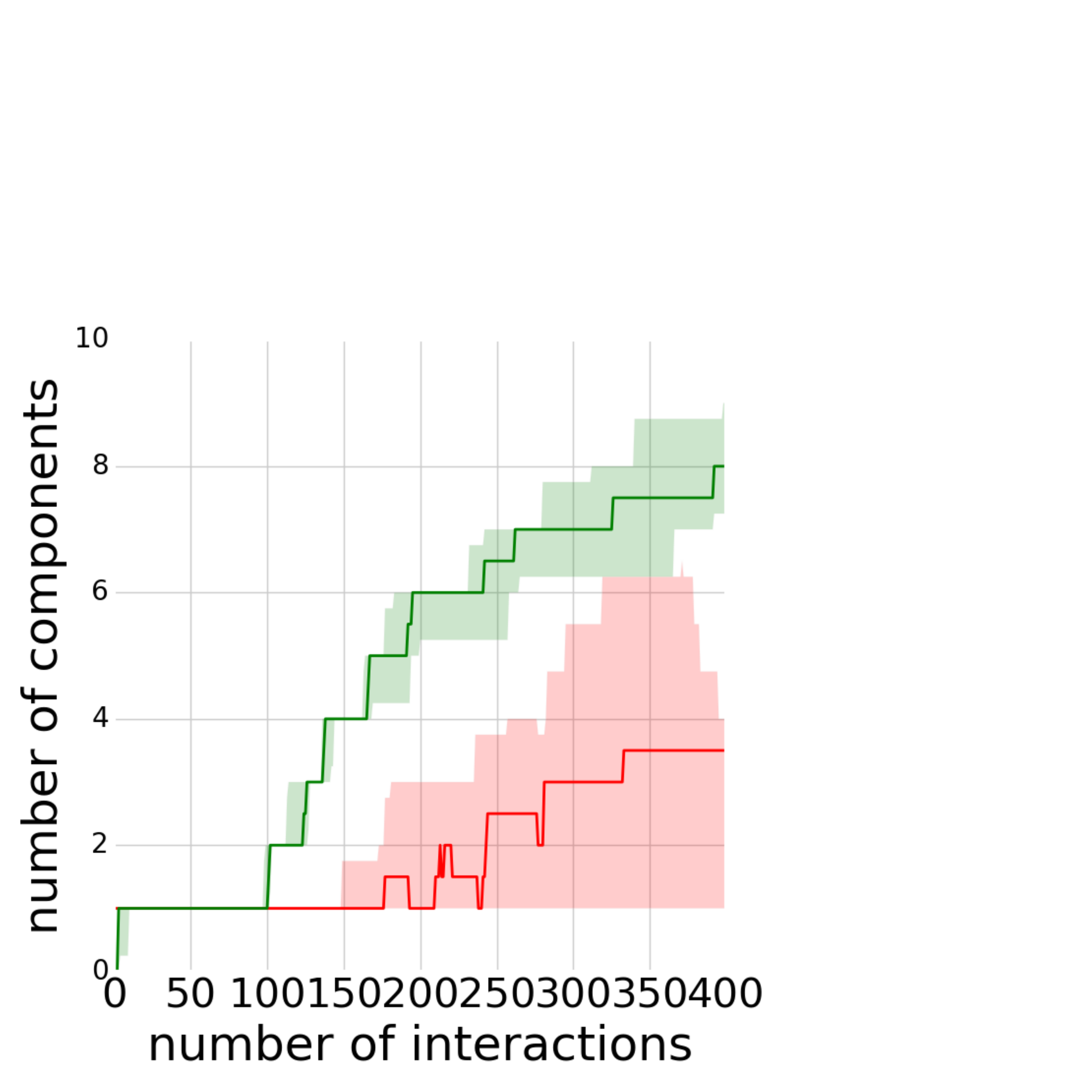}
} 
\subfloat[Plot for setup \ref{fig:setup5}]{\label{fig:nbcomp5}
\includegraphics[height=60mm,width=.45\linewidth,keepaspectratio]{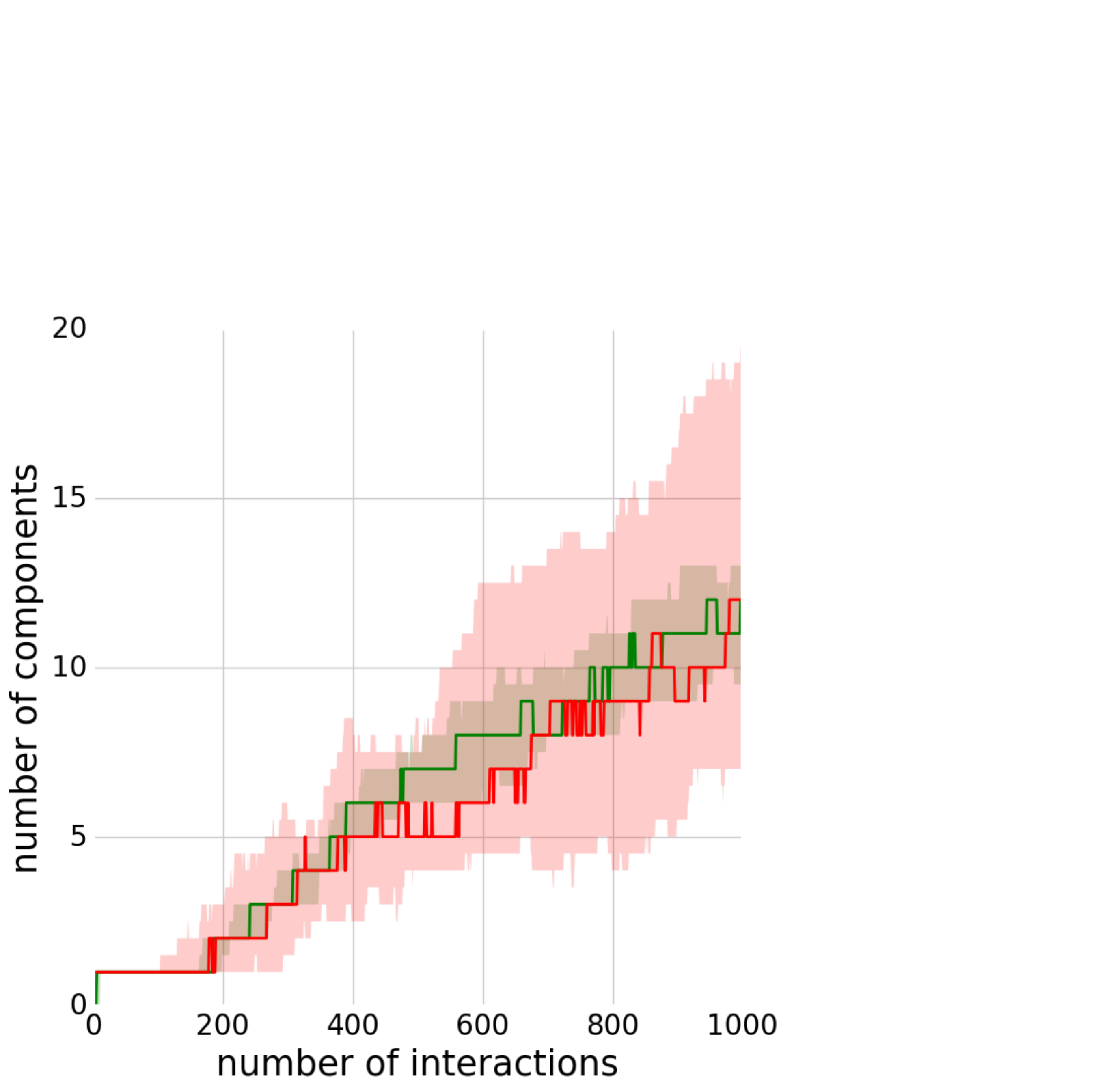}
}  \\
\subfloat{
\includegraphics[height=60mm,width=.45\linewidth,keepaspectratio]{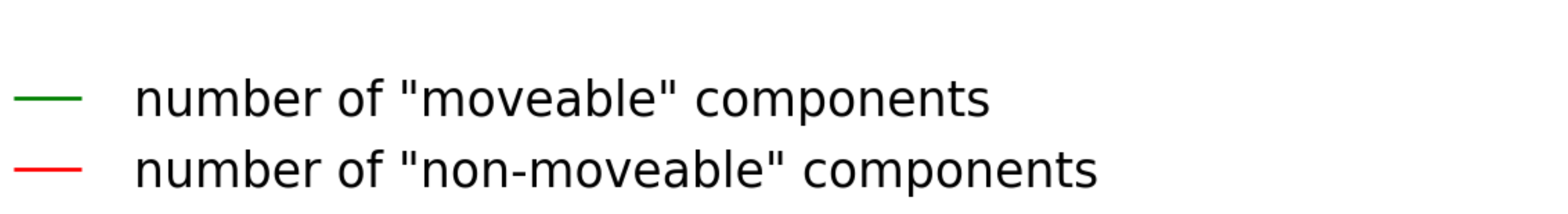}
}
\caption{Plots of the number of components of each class at each iteration during the experiments for each simplified setup presented in figure \ref{fig:setups}.}
\label{fig:nbcomp}
\end{figure}

Figure \ref{fig:nbcomp} shows the number of components for each class at each iteration. 
The increasing number of components when the setup becomes more complex is worth to look at. Most likely this corresponds to cases when both classes share common features. Setup \textbf{WhiteMoveableBricks} (\ref{fig:setup5}), which led to the worst results (see figure \ref{fig:praid5}), has a rapidly increasing number of components and does not reach a plateau within 1000 iterations (see figure \ref{fig:nbcomp5}). Also, for all setups except setup \ref{fig:setup5}, new components are created only after iteration 100; this is a consequence of the constraint introduced by the intersection criterion (see equation \ref{eq:intercond}).  For the setup on the simulated kitchen (\ref{fig:simkitchen}), the number of components are also increasing but seems to slow down (see figure \ref{fig:kitchencomp}). Moreover, the number of components for the non-moveable category is much lower than for the moveable category. This suggests that the background is less complex than the objects, which seems to be the case (see figure \ref{fig:simkitchen}). 

 Figure \ref{fig:praid_fix_comp} shows the performance of experiments conducted with only one component per model, i.e. when no split or merge operations are applied. In this case, performances are poorer than those including split and merge operations. This indicates that setups \textbf{MoveableBricks1} (\ref{fig:setup1}), \textbf{MoveableBricks2} (\ref{fig:setup3}) and \textbf{WhiteMoveableBricks} (\ref{fig:setup5}) involve nonconvex classes and nonlinearly separable datasets and justifies the need for the proposed split and merge operations.

\begin{figure}[h]
\centering
\subfloat[Accuracy]{\label{fig:alphas_acc}
\includegraphics[height=60mm,width=.9\linewidth,keepaspectratio]{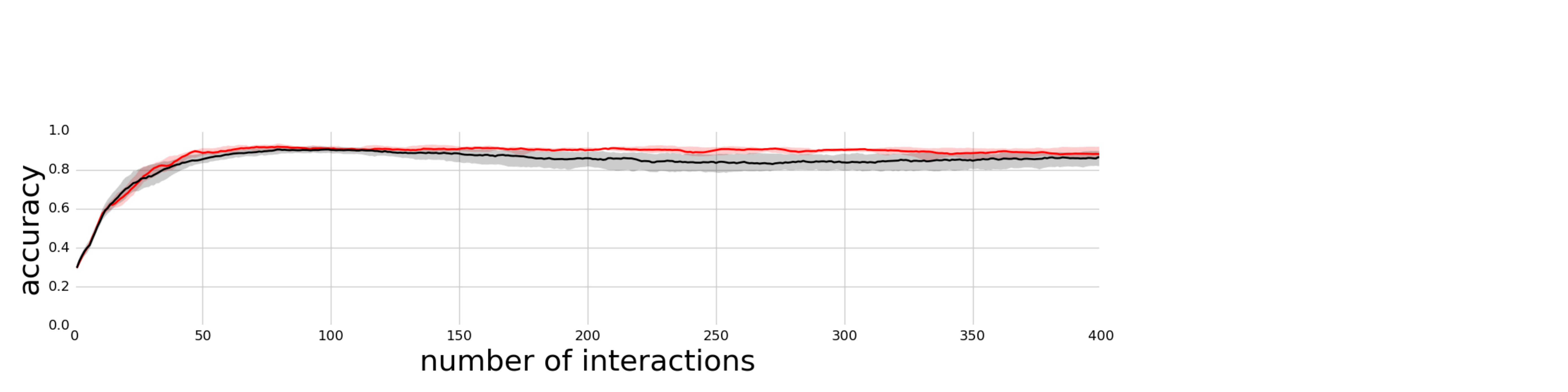}
} \\
\subfloat[Precision]{\label{fig:alphas_pre}
\includegraphics[height=60mm,width=.9\linewidth,keepaspectratio]{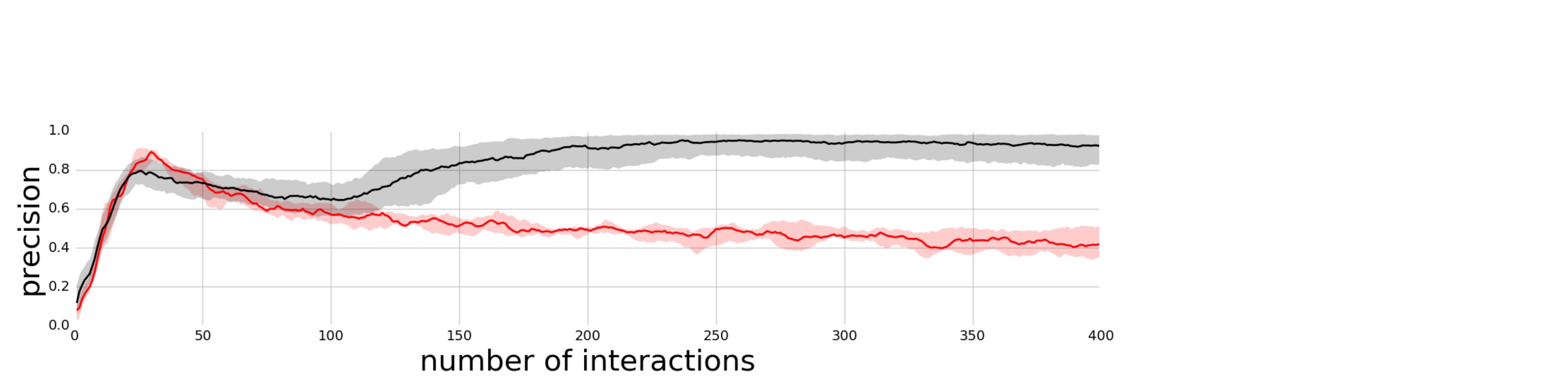}
}\\
\subfloat[Recall]{\label{fig:alphas_rec}
\includegraphics[height=60mm,width=.9\linewidth,keepaspectratio]{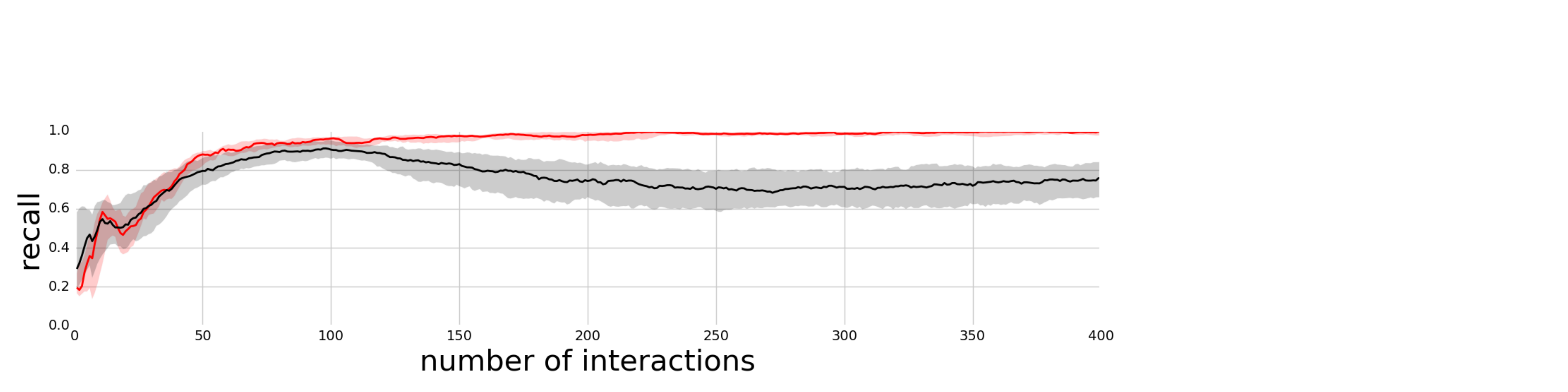}
} \\
\subfloat{
\includegraphics[height=60mm,width=.2\linewidth,keepaspectratio]{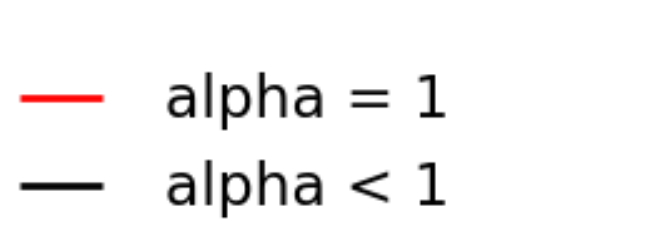}
}
\caption{Results of the experiments conducted on the simulated kitchen with $\alpha$ varying between 0 and 1 with a step of 0.1. The replication with $\alpha$ strictly less than 1 are grouped into the black curve and equal to one into the red curve}
\label{fig:alphas}
\end{figure}

Figure \ref{fig:alphas} represents the results of experiments conducted on the simulated kitchen (\ref{fig:simkitchen}) with $\alpha$ varying between 0 and 1 with a step of 0.1. For better clarity, the replications with alphas between 0.9 and 0 are grouped in the black curve, thus, the black curve gather 100 replications and the red one 10 replications. The replications are grouped like that because there is no splits and merges only for $\alpha$ equal to 1. The accuracy (see figure \ref{fig:alphas_acc}) is approximately the same for any value of alpha. For $\alpha$ strictly less than 1, the precision converged to a value around 0.9 (see figure \ref{fig:alphas_pre}) and for recall to a value around 0.8 (see figure \ref{fig:alphas_rec}), while, for $\alpha$ equal to 1, the precision keeps decreasing and the recall converged quickly to a value close from 1. Also, for $\alpha$ strictly less than 1, recalls and precisions have a low variability. The results seem to indicate that with split and merge, for any value of $\alpha$ (strictly less than 1), the classification reaches a sufficient quality with a bit low recall, while without split and merge ($\alpha$ equal to 1), the classification has a very low precision despite a high accuracy and recall. 

Finally, the $\alpha$ parameters could be fixed to any value between 0.9 and 0. The split and merge operations allow the system to have a better precision in classification but they lower the recall.

\subsection{Real world setup}

\begin{figure}[h]
\centering
\subfloat[Plot for setup \ref{fig:real_setup0}]{\label{fig:prareal0}
\includegraphics[height=60mm,width=.95\linewidth,keepaspectratio]{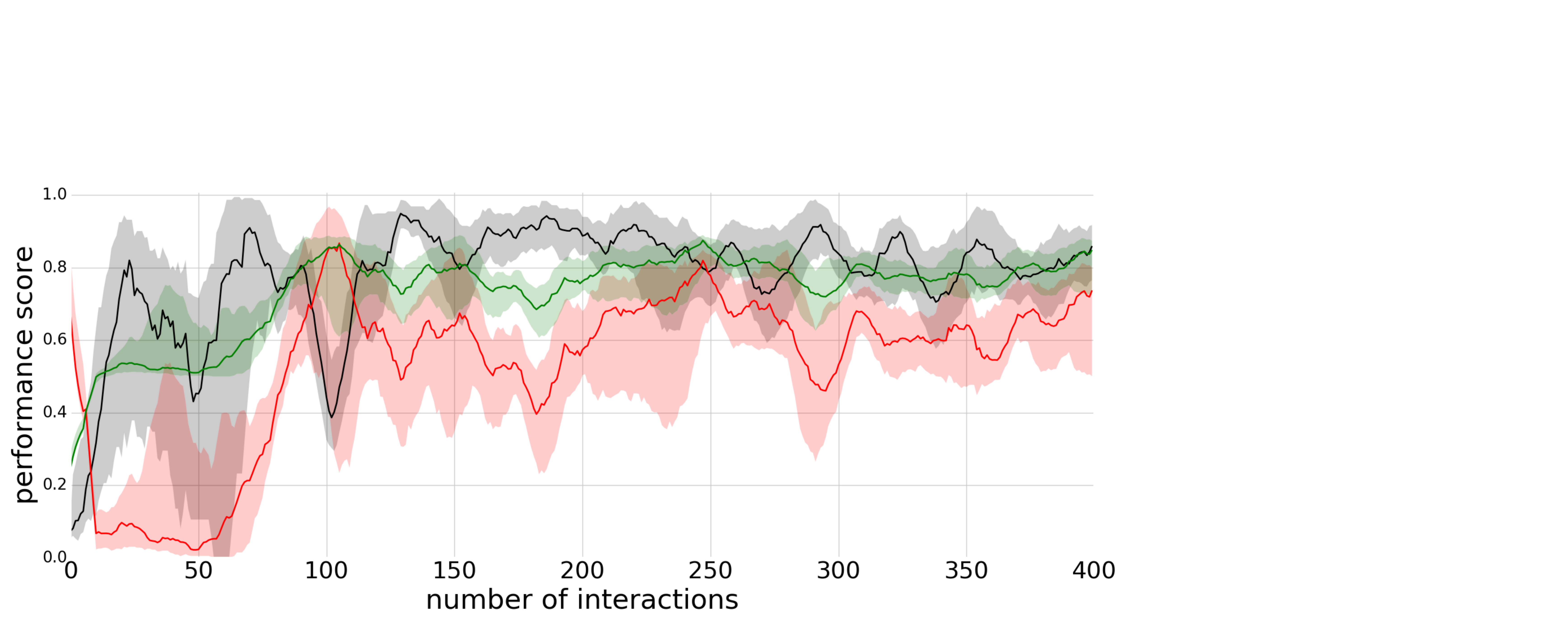}
} \\
\subfloat[Plot for setup \ref{fig:real_setup1}]{\label{fig:prareal1}
\includegraphics[height=60mm,width=.95\linewidth,keepaspectratio]{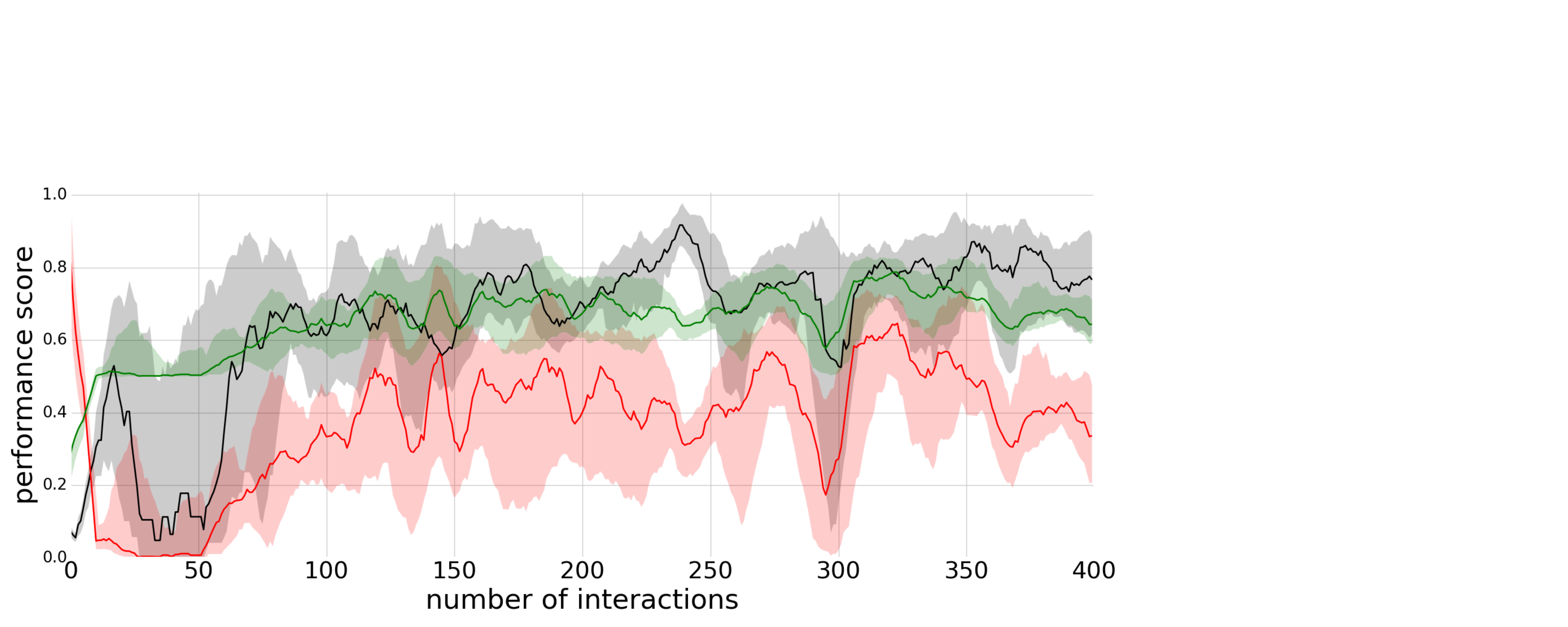}
} \\
\subfloat{
\includegraphics[height=60mm,width=.9\linewidth,keepaspectratio]{legend_pra.pdf}
} 
\caption{Plots of precision, recall, and accuracy for each real setup presented in figure \ref{fig:real_setups} }
\label{fig:prareal}
\end{figure}

\begin{figure}[h]  
\centering
\subfloat[Plot for setup \ref{fig:setup5}]{\label{fig:final_rm0}
\includegraphics[height=60mm,width=.95\linewidth,keepaspectratio]{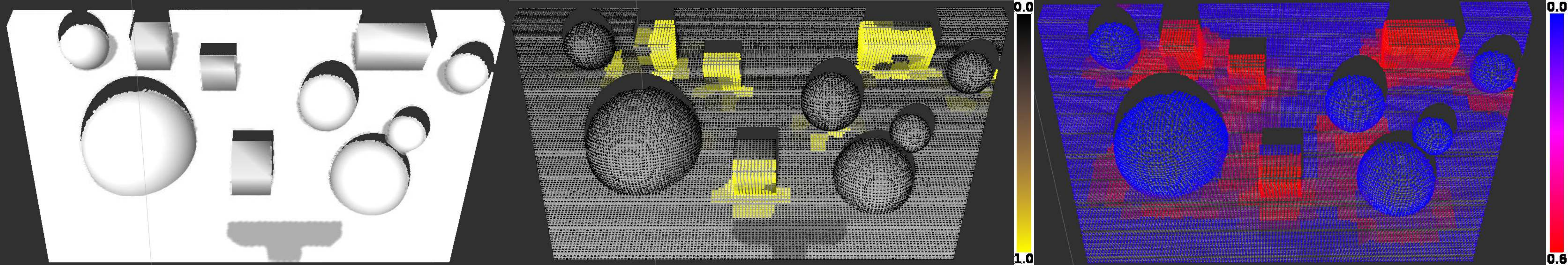}
} \\
\subfloat[Plot for setup \ref{fig:real_setup0}]{\label{fig:final_rm1}
\includegraphics[height=60mm,width=.95\linewidth,keepaspectratio]{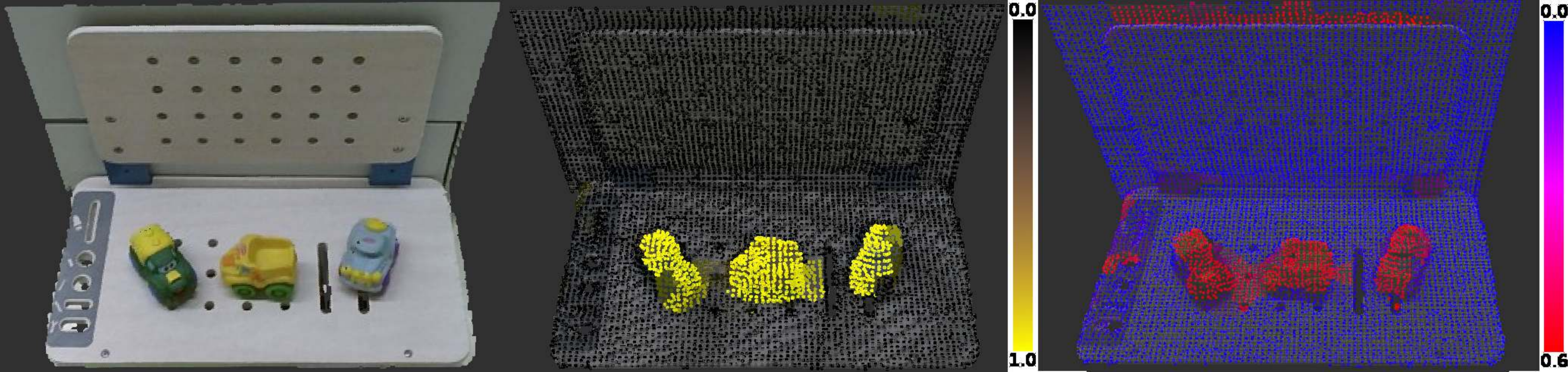}
} \\
\subfloat[Plot for setup \ref{fig:real_setup1}]{\label{fig:final_rm2}
\includegraphics[height=60mm,width=.95\linewidth,keepaspectratio]{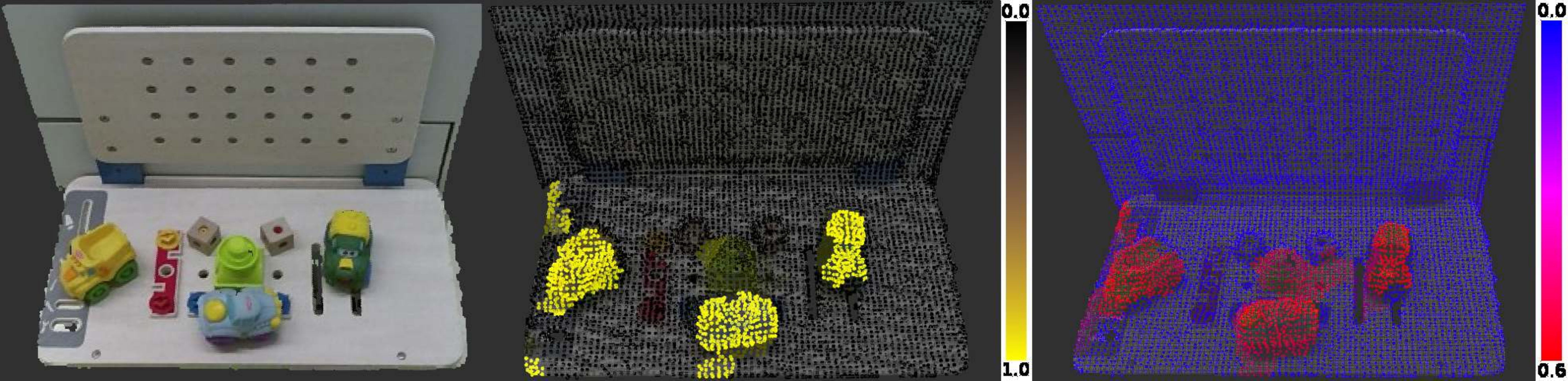}
}    
\caption{From right to left : Cloud pictures of both real setups (\ref{fig:real_setups}) and setup \textbf{WhiteMoveableBricks} (\ref{fig:setup5}); Relevance map with in yellow highest probability to be moveable and in black lowest (from 0.0 to 1.0); AveraThe selection process is far from random samplingge choice distribution map over an exploration with in red most explored areas and blue least explored areas (from 0.0 to 0.6).} 
\label{fig:final_rm}
\end{figure}

In this setup, the data is obtained after the application of the push movement primitive and the detection of an eventual change in the targeted area.
As expected, the performance in the real environment does not reach the maximum level, but it does reach 0.8 for accuracy (see figure \ref{fig:prareal}). This is enough to produce a useful segmentation of the scene, as shown in figures \ref{fig:final_rm1} and \ref{fig:final_rm2}. The classification is less efficient for setup \textbf{Workbench2} (\ref{fig:real_setup1}) than for setup \textbf{Workbench1} (\ref{fig:real_setup0}), and it is less stable over the iterations. This is expected as setup \textbf{Workbench2} (\ref{fig:real_setup1}) is more complex than setup \textbf{Workbench1} (\ref{fig:real_setup0}).
 
\begin{figure}[h]
\centering
\subfloat[Plot for setup \ref{fig:real_setup0}]{\label{fig:posnegreal0}
\includegraphics[height=60mm,width=.45\linewidth,keepaspectratio]{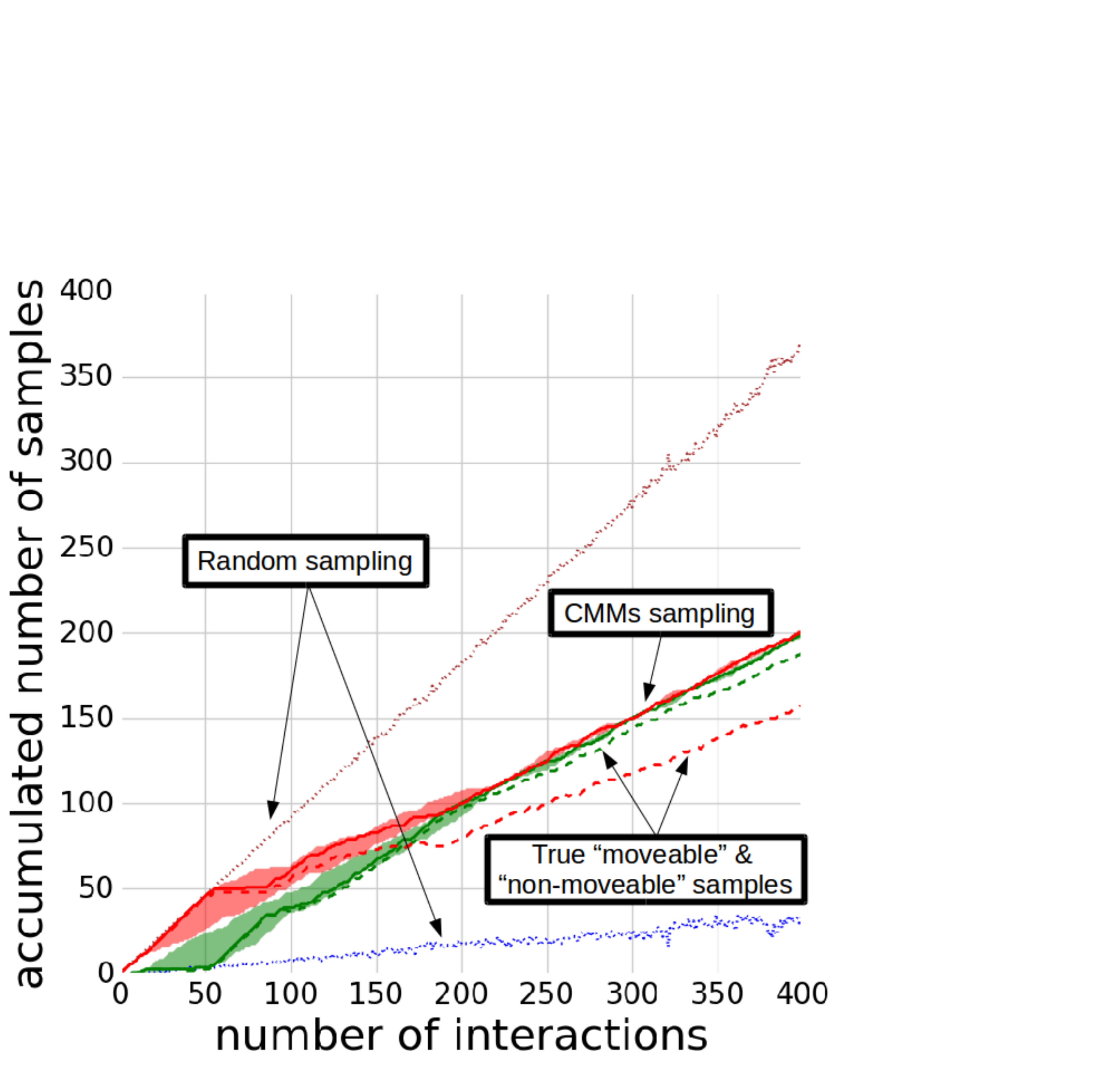}
}
\subfloat[Plot for setup \ref{fig:real_setup1}]{\label{fig:posnegreal1}
\includegraphics[height=60mm,width=.45\linewidth,keepaspectratio]{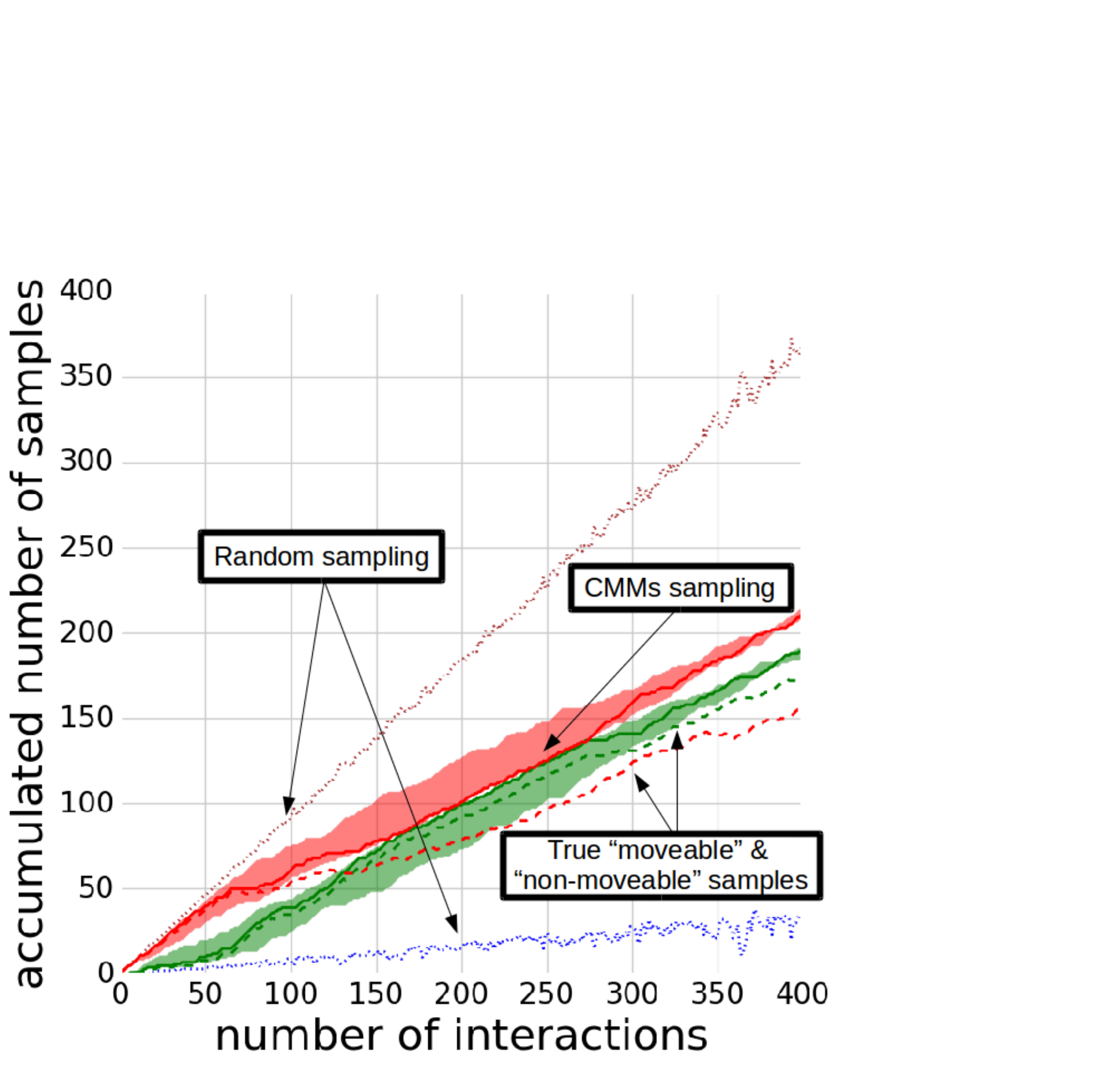}
} \\
\subfloat{
\includegraphics[height=60mm,width=.45\linewidth,keepaspectratio]{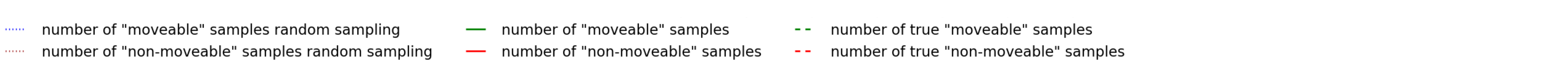}
} 
\caption{Plots of the number of samples gathered for each class at each iteration during the experiments for each real setup presented in figure \ref{fig:real_setups}. }
\label{fig:posnegreal}
\end{figure}

A total of 400 iterations is enough to achieve almost the same amount of samples in both classes, as shown in figure \ref{fig:posnegreal}. Compared with the simulation, the exploration produces mislabeled samples owing to failed interactions and failures in the detection of motion. As the classifier is online, the training is more sensitive to mislabeled samples which introduce instability over the iterations and variability over the replications.

\begin{figure}[h]
\centering
\subfloat[Plot for setup \ref{fig:real_setup0}]{\label{fig:nbrcompreal0}
\includegraphics[height=60mm,width=.45\linewidth,keepaspectratio]{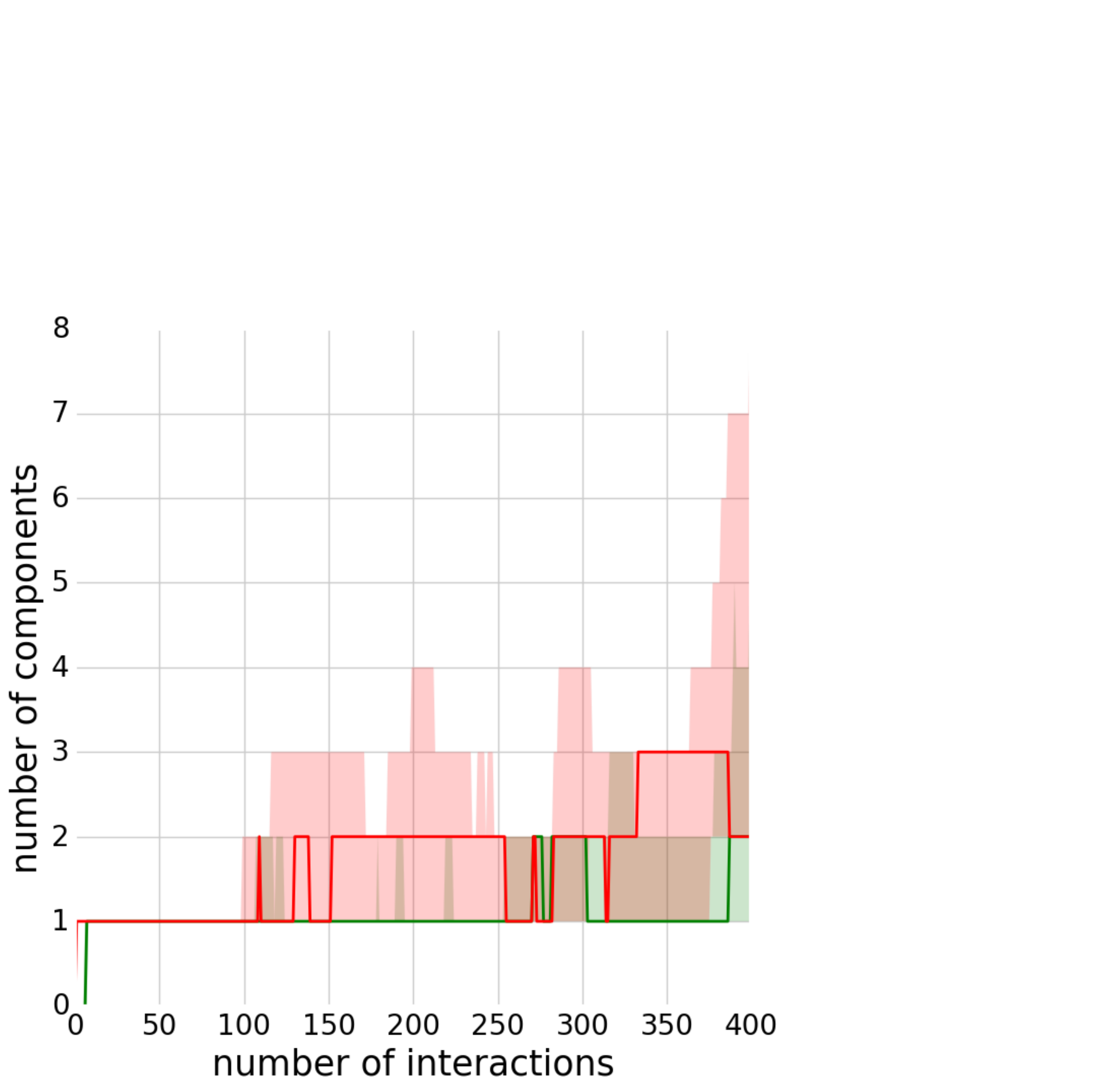}
}
\subfloat[Plot for setup \ref{fig:real_setup1}]{\label{fig:nbrcompreal1}
\includegraphics[height=60mm,width=.45\linewidth,keepaspectratio]{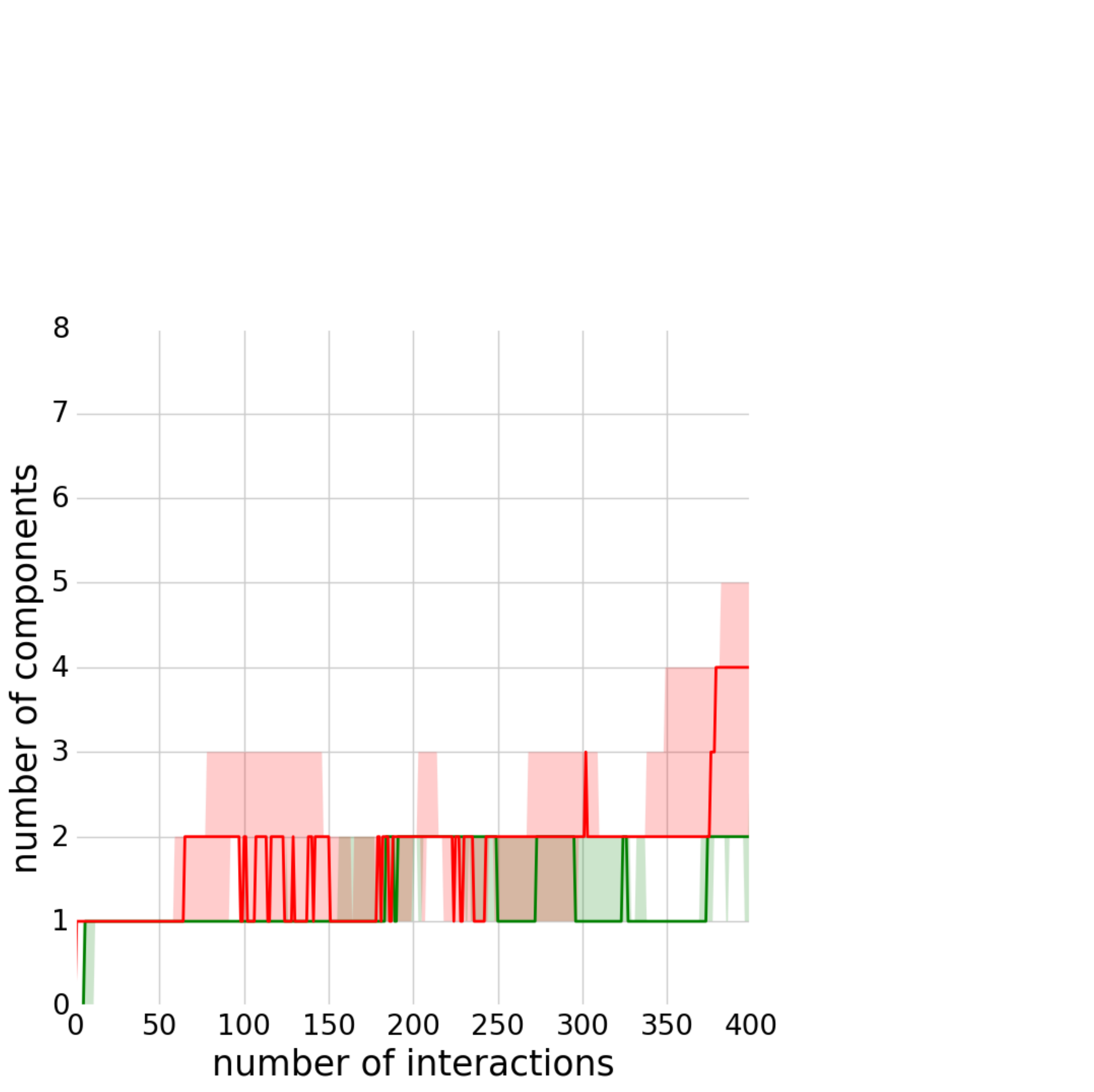}
} \\
\subfloat{
\includegraphics[height=60mm,width=.7\linewidth,keepaspectratio]{legend_nbr_comp.pdf}
} 
\caption{Plots of the number of components of each class at each iteration during the experiments for each real setup presented in figure \ref{fig:real_setups}.}
\label{fig:nbrcompreal}
\end{figure}

The increasing number of components seen in the simulation is related to the complexity of the environment; with the real robot, it is also related to mislabeled samples. Indeed, mislabeled samples introduce a higher complexity in the distribution of samples in the feature space, therefore components split is more likely to occur. This explains the large variability in the number of components on figure  \ref{fig:nbrcompreal0}. 

\begin{figure*}[t!]
\centering
\subfloat[Relevance map after 1 interaction]{\label{fig:rm_it1}
\includegraphics[height=60mm,width=.19\linewidth,keepaspectratio]{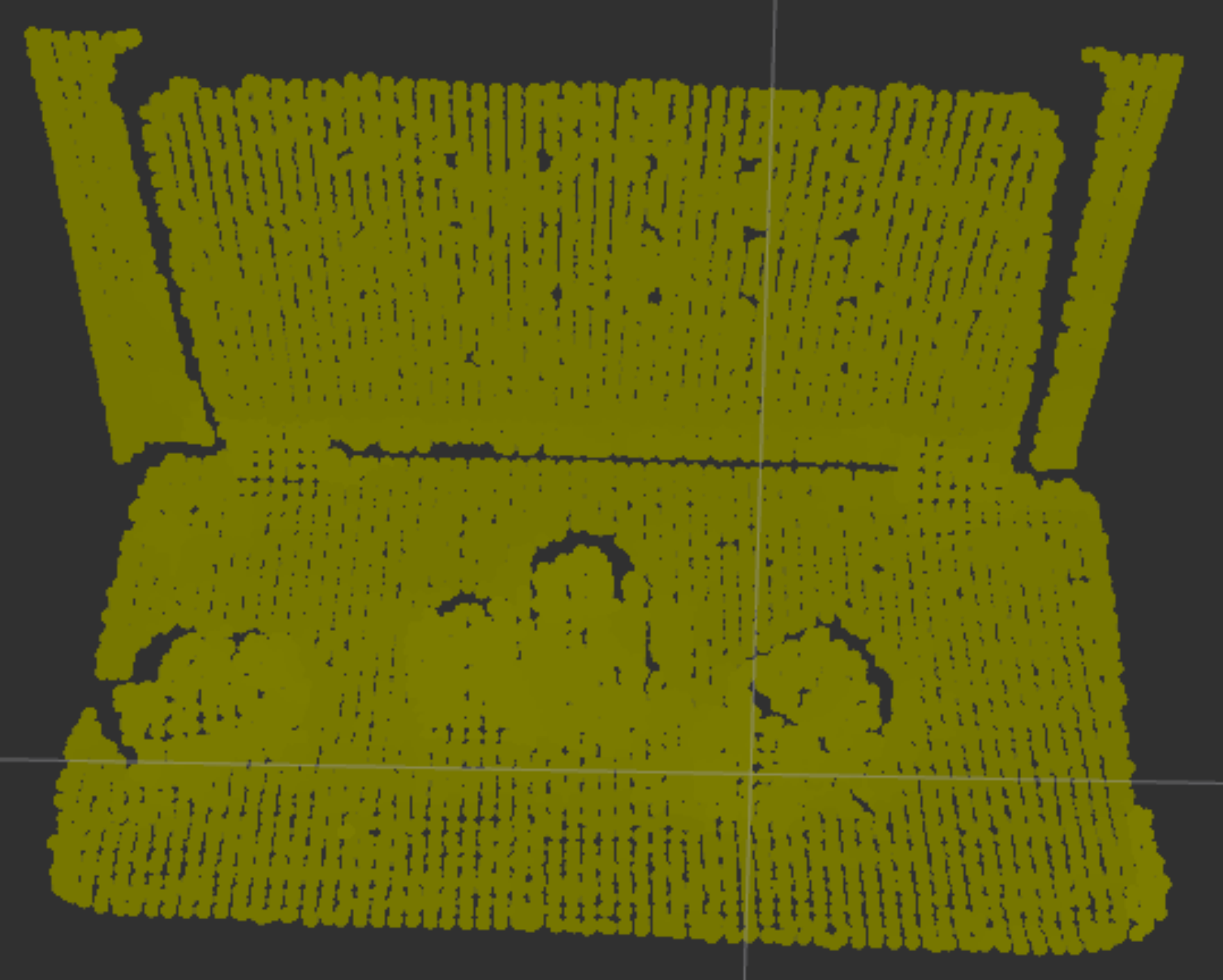}
} 
\subfloat[Relevance map after 10 interactions]{\label{fig:rm_it10}
\includegraphics[height=60mm,width=.19\linewidth,keepaspectratio]{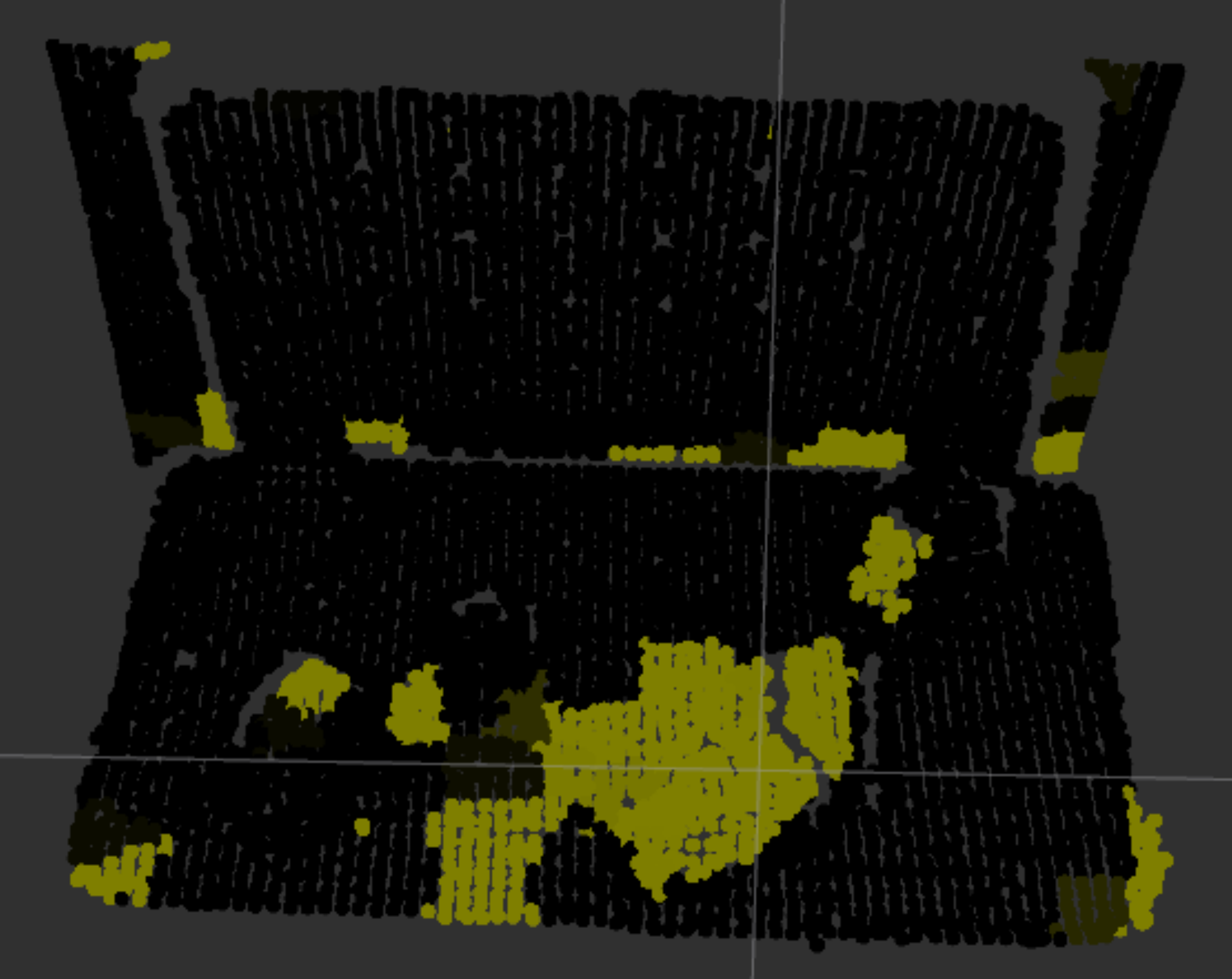}
}
\subfloat[Relevance map after 50 interactions]{\label{fig:rm_it50}
\includegraphics[height=60mm,width=.19\linewidth,keepaspectratio]{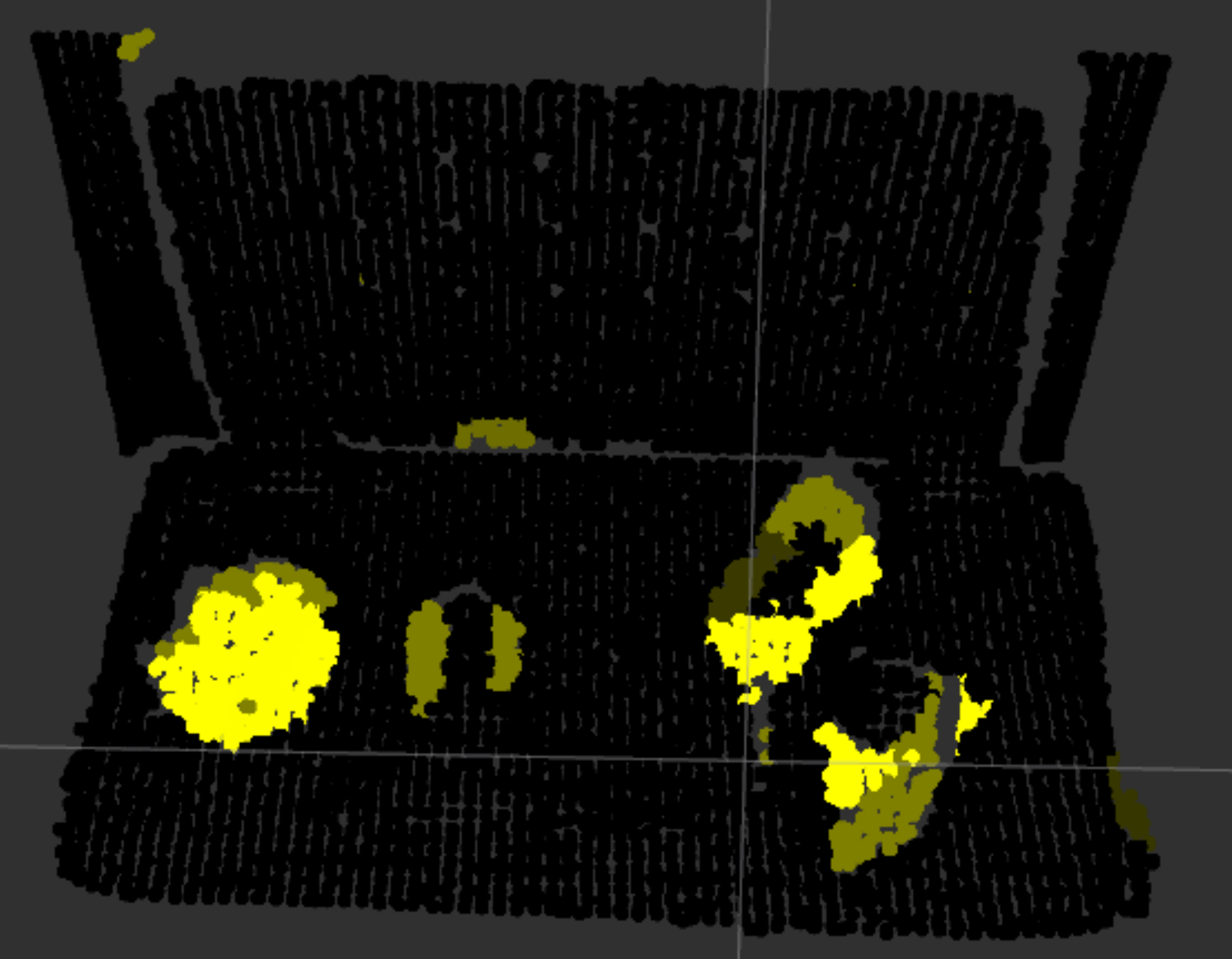}
} 
\subfloat[Relevance map after 100 interactions]{\label{fig:rm_it100}
\includegraphics[height=60mm,width=.19\linewidth,keepaspectratio]{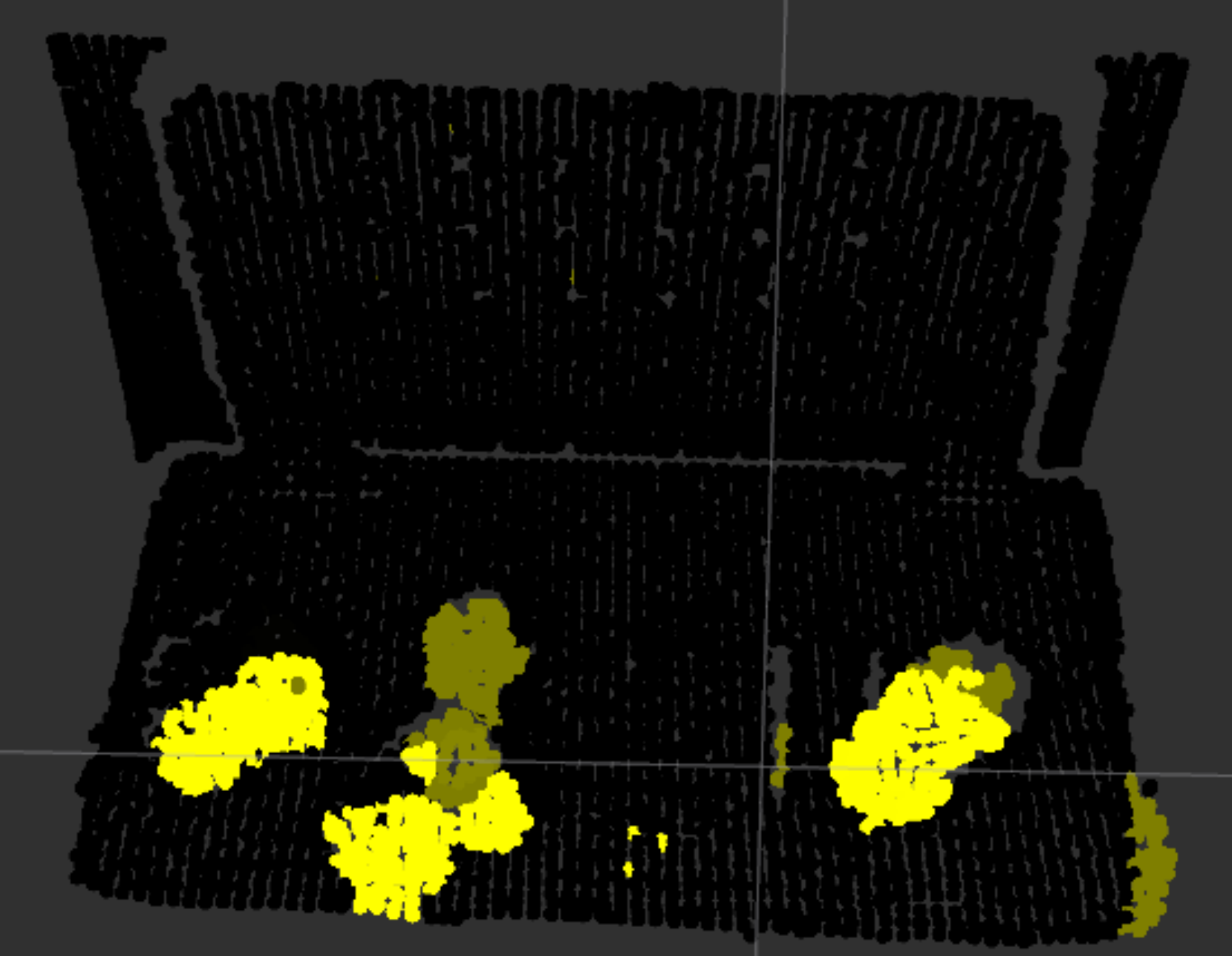}
} 
\subfloat[Relevance map after 400 interactions]{\label{fig:rm_it400}
\includegraphics[height=60mm,width=.19\linewidth,keepaspectratio]{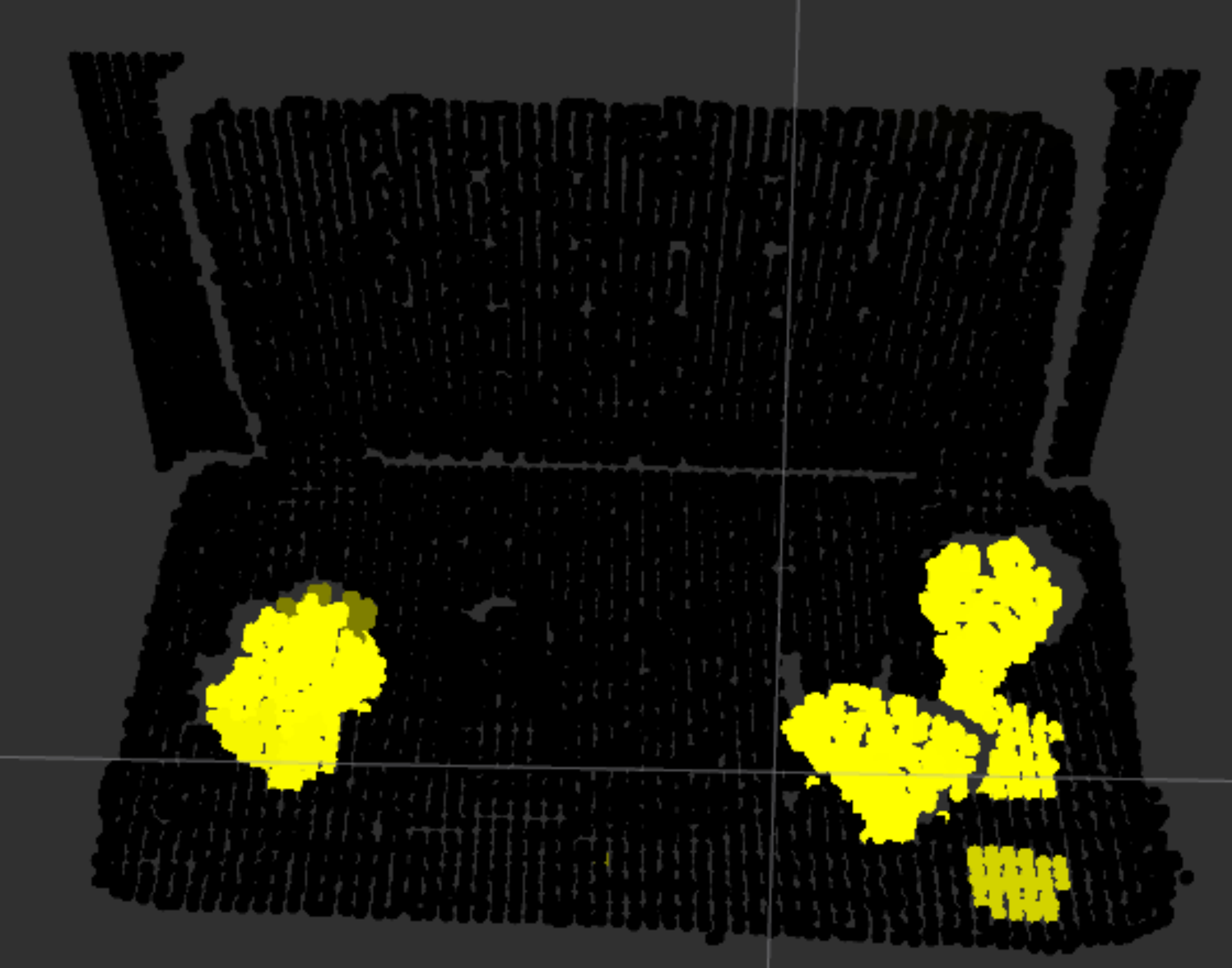}
} 
\subfloat{
\includegraphics[height=60mm,width=.013\linewidth,keepaspectratio]{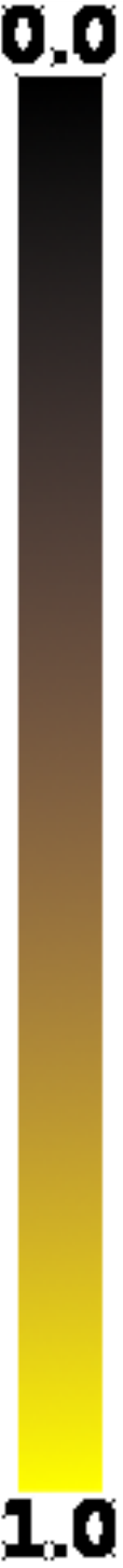}
} \\
\subfloat[Pointcloud used to generate above image]{\label{fig:cloud_it1}
\includegraphics[height=60mm,width=.19\linewidth,keepaspectratio]{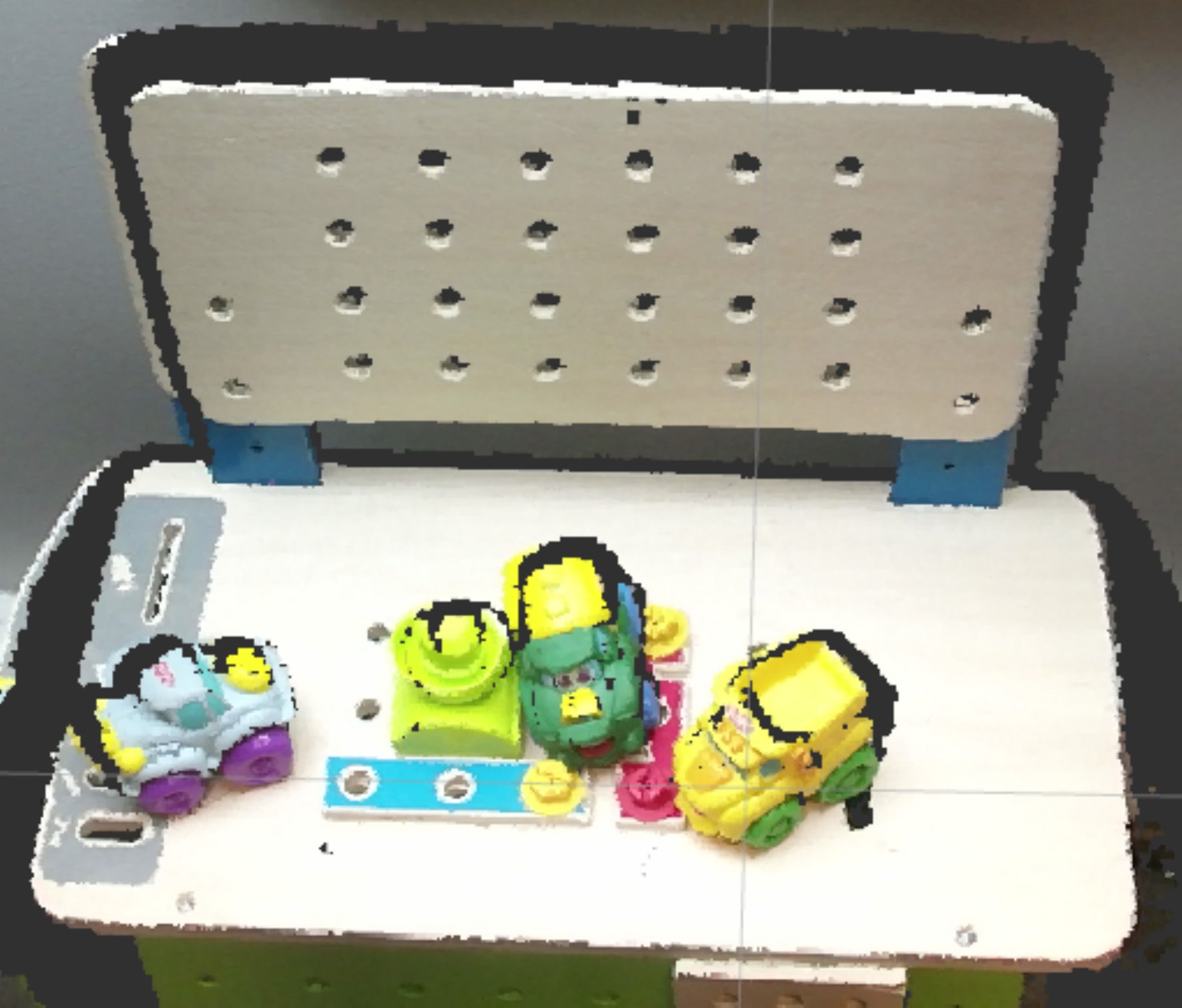}
} 
\subfloat[Pointcloud used to generate above image]{\label{fig:cloud_it10}
\includegraphics[height=60mm,width=.19\linewidth,keepaspectratio]{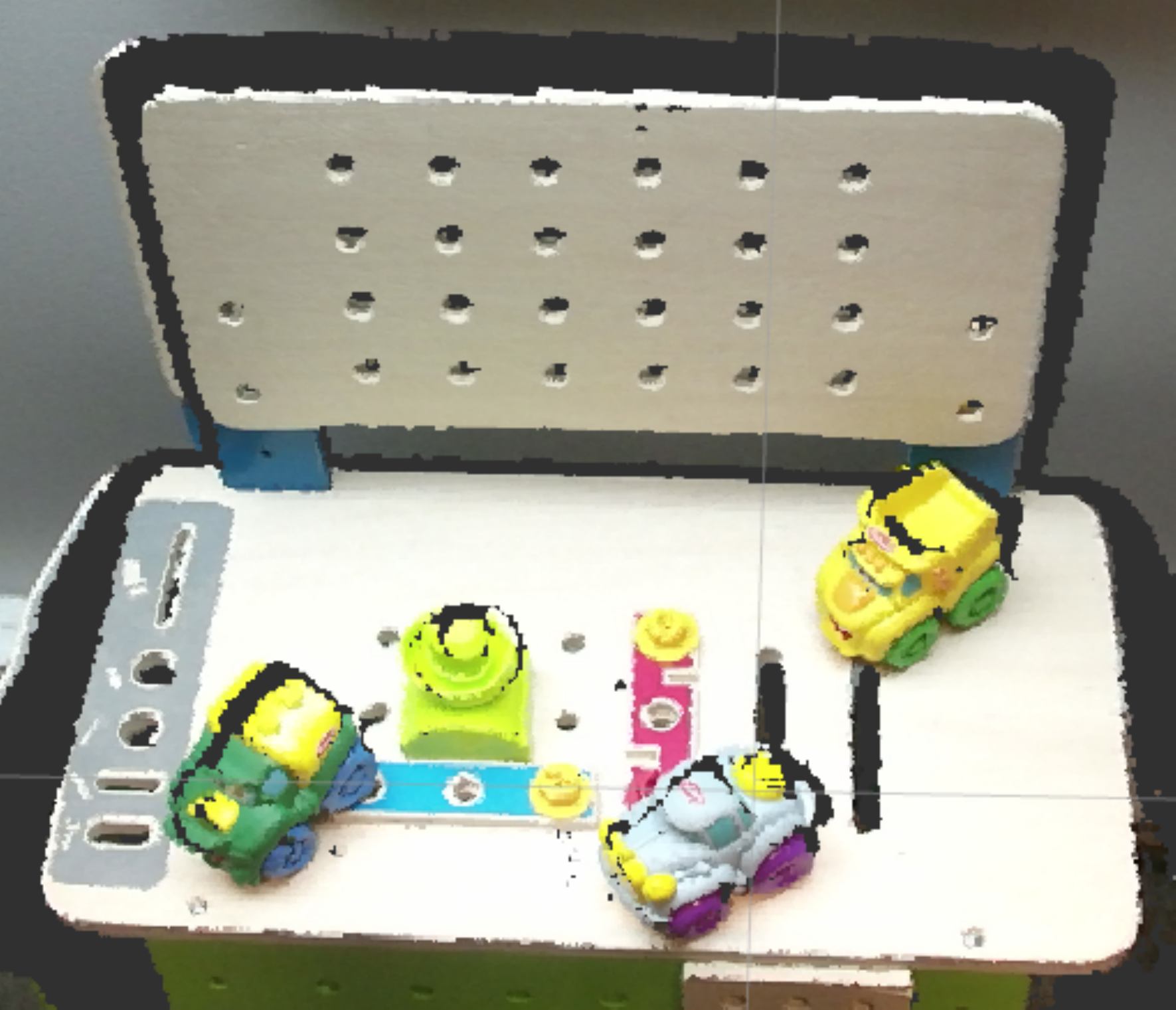}
}
\subfloat[Pointcloud used to generate above image]{\label{fig:cloud_it50}
\includegraphics[height=60mm,width=.19\linewidth,keepaspectratio]{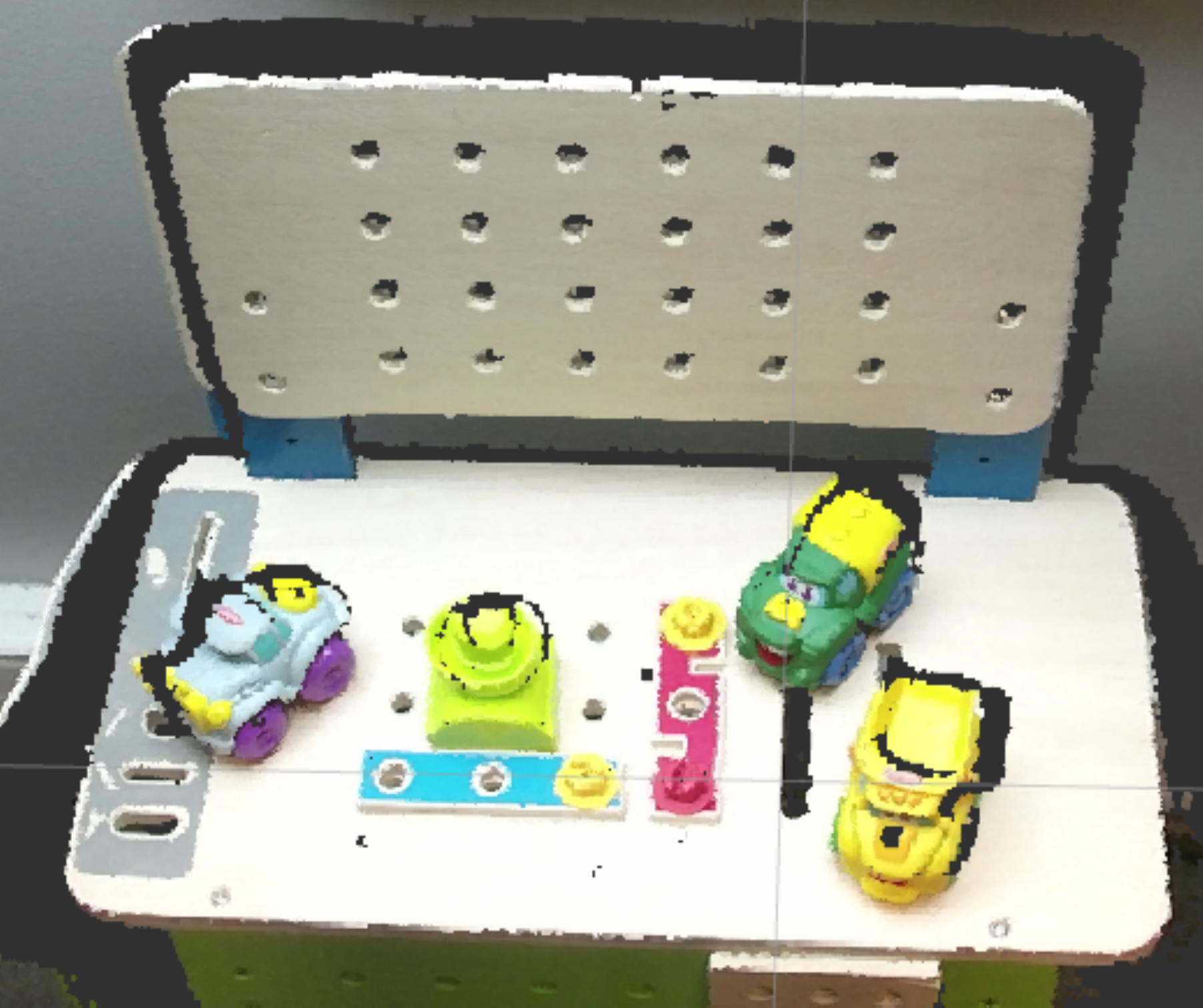}
} 
\subfloat[Pointcloud used to generate above image]{\label{fig:cloud_it100}
\includegraphics[height=60mm,width=.19\linewidth,keepaspectratio]{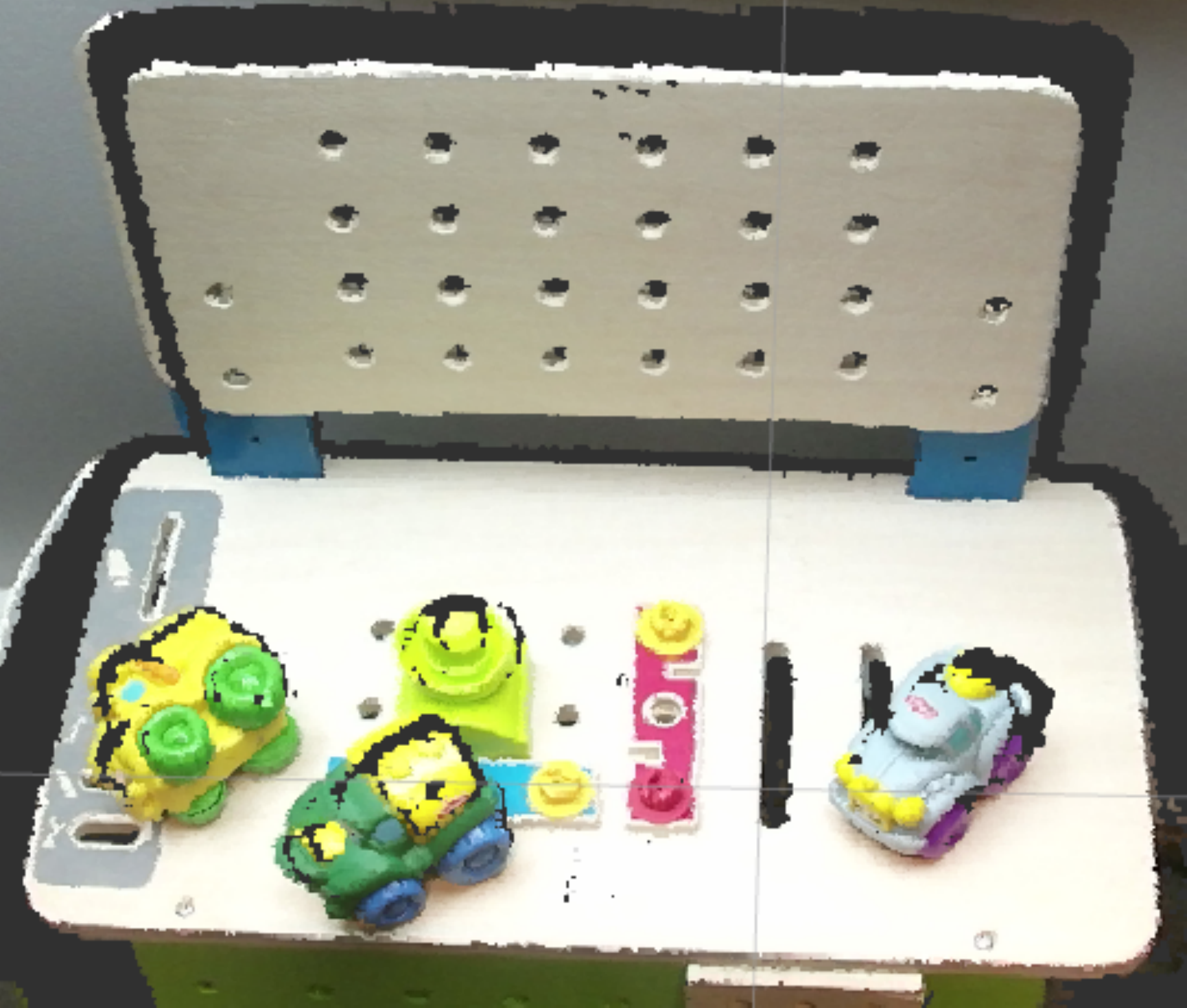}
} 
\subfloat[Pointcloud used to generate above image]{\label{fig:cloud_it400}
\includegraphics[height=60mm,width=.19\linewidth,keepaspectratio]{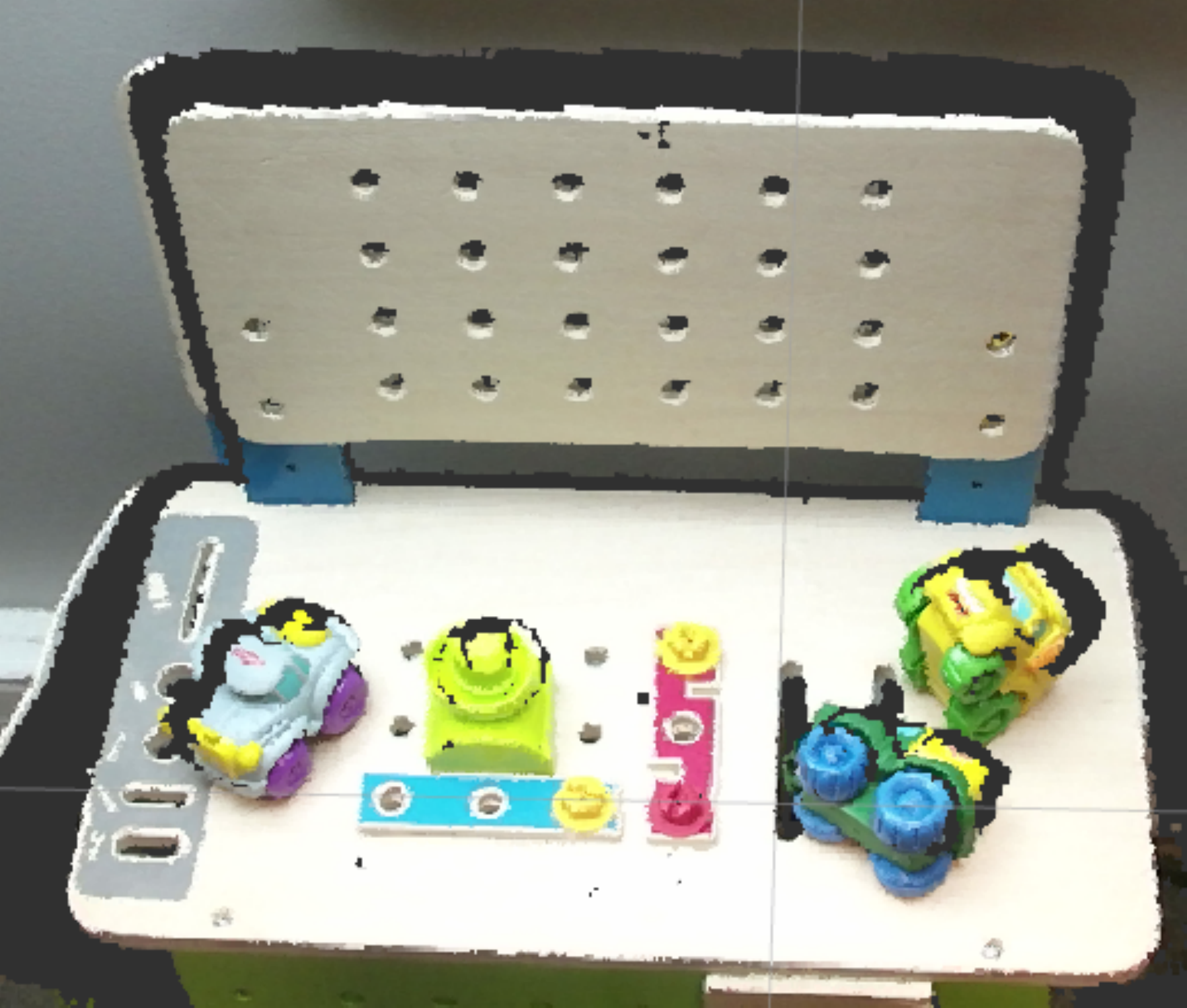}
} 
\caption{Sequence of pointclouds representing a relevance map at different point of during the exploration. This images have been generated after an exploration.}
\label{fig:rm_seq}
\end{figure*}

Figure \ref{fig:final_rm} presents the best performing relevance map for both real setups and for the most complex simulated setup, \textbf{WhiteMoveableBricks} (\ref{fig:setup5}). It also shows an average choice distribution map over all iterations, which represents the most probable exploration areas. This map provides an insight into which parts of the environment are most considered during exploration.  For the three setups, the exploration is more focused on complex areas such as moveable or fixed objects (that are then part of the background). With an exploration driven by uncertainty, this is an expected feature as the complex areas are the slowest to decrease their uncertainty. 
Finally, figure \ref{fig:rm_seq} shows a sequence of relevance maps generated from classifiers taken at different moments of the exploration on the setup \textbf{Workbench2} (\ref{fig:real_setup1}). The first relevance map after 1 interaction is totally neutral, showing that all the environment is considered as moveable with a probability of $\frac{1}{2}$, while the last relevance map attributes a high probability to almost only the cars.

All the source code used to produce these results can be found on github\endnote{CMMs source code: \url{https://github.com/LeniLeGoff/IAGMM_Lib/tree/rm-pub-18}; Experiments source code: \url{https://github.com/robotsthatdream/babbling_experiments/}}.

\section{Discussion and Future work}\label{sec:disc}

The exploration process is the most crucial component of this method. On the one hand, the classifier requires a representative dataset of the scene; on the other hand, it requires a suitable sample at each iteration to learn efficiently. With only a uniform random exploration, the dataset would be composed of an overwhelming majority of background samples; therefore, the classifier would have difficulty converging (see figures \ref{fig:posneg} and \ref{fig:posnegreal}). In the proposed approach, the sampling process is based on an uncertainty reduction. Uncertainty is measured based on the probability of membership to each class being close to 0.5, i.e. at the border of both Gaussian mixture models. This focuses exploration on unknown or poorly known areas. The confidence of the classification is used to focus the exploration on areas in which the dataset has a low density. Confidence draws inspiration from entropy while being less costly to compute. The entropy of models is a measure of information quantity. The selection process thus increases the representativeness of the dataset. Combining uncertainty and confidence allows the exploration process to focus on unknown and informative areas, as shown in the left picture of figure \ref{fig:final_rm}.  Finally, to balance the dataset between the two classes, priority is given to sampling the less represented class of the dataset. 

The quality of the relevance map depends on the precision of the change detector. In this work, the change detector is a simple frames comparison between before and after the execution of the push primitive. This component of the framework could be easily enhanced by adding haptic sensors on the robots end-effector or real-time tracking system for motion detection. The other components are mostly independent of the change detector, thus, it can be changed without major issues. 

The online training of CMMs does not offer a precise measure of convergence. In batch learning, test steps give a measure of overfitting and the generalization error, which is not achievable in online learning. In online learning, establishing a test dataset is complicated as the test dataset must be sufficiently different from the training dataset to detect overfitting and to compute generalization error. Within the budget fixed for the experiment, the classifier converges, as shown in figures \ref{fig:praid} and \ref{fig:prareal}, in which precision, recall, and accuracy converge to a mean value. In addition, the exploration always reaches a balanced dataset, as shown in figures \ref{fig:posneg} and \ref{fig:posnegreal}, suggesting the convergence of the classifier. But, for the most complex setup \ref{fig:real_setup1}, the performance is unstable and decrease several times. Thus, accumulating more samples does not guarantee an increase in performance.
Loglikelihood is a promising path to explore for a measure of convergence as it is used, for example, in the M-step of the EM algorithm. 

The classifier used in this paper is designed to be non-specific to a particular kind of environments. In particular, the hyperparameters of the model should be the same for all environments. CMMs have one critical hyperparameter : the tolerance ellipse size ($\alpha$). It determines the sensitivity of the \textit{merge} and \textit{split} operations. This parameter could be tuned to have best classification efficiency but the experiments show a low variability on the results when changing $\alpha$ (see \ref{fig:alphas}). In this study, the value of $\alpha$ was fixed to the same value for all the experiments ($\alpha = 0.25$) (see sections \ref{sec:alg} and \ref{sec:proto}). Thus, a compromise for a large set of environments can be found for the value of this hyperparameter. Moreover, varying the value of $\alpha$ between 0.9 and 0 does not introduce a high variability in classification quality.

The quality of the classification is conditioned by the features used. A complex environment would require the use of features that can capture this complexity to generate an efficient segmentation. However, as the feature extractor is designed prior to exploration, it could potentially reduce the kinds of environment to which the robot can adapt. The results of setups \textbf{WhiteMoveableBalls} (\ref{fig:setup4}) and \textbf{WhiteMoveableBricks} (\ref{fig:setup5}) show that the feature space can be more complex than required in practice (in these setups, the color descriptor is ignored by the method). As in the work of \citet{jiang2013discri}, the feature space is 93-dimensional, which is more than what is actually required for a lot of problems but who can do more, can do less.  In the proposed approach, the feature used is 48-dimensional, which is already large and integrates rich shape and color information. It allows the method to be more adaptive. Thus, to deal with any situation the robot may encounter, the use of a descriptor as rich as possible is recommended. One option is to use a designed visual features (as in this work); another option is to use the features built by a pretrained convolutional neural network \citep{hariharan2015hypercolumns}.

The interactive perception paradigm has a strong link with affordances. An affordance is a relational property which emerges from the agent-environment system . This concept allows to formalize a representation of the environment through the action of the robot \citep{gibson1979}. As the relevance map is built thanks to the interaction of the robot with a push primitive, it represents the probability of environment parts to be "pushable", in other words, it represents areas of the environment that afford the push action for the robot. As for the change detector, the primitive is an independent component of the framework, so, it is possible to change the primitive with a minimum of efforts. Other experiments with primitives like lifting, pulling, or grasping could be conducted with the proposed methods. Of course, a change detector adapted to the new primitive needs to be provided. In these cases, the relevance map would represent affordances like "liftable", "pullable" or "graspable".

\section{Conclusion}\label{sec:con}

A method has been introduced that allows a robot to segment a visual scene into two different classes: regions that belong to moveable areas and regions that belong to non-moveable areas. The method relies on interactive perception. It includes a classifier that is trained online and a sampling process that selects the regions upon which to focus. The classifier is called collaborative mixture models (CMMs). It has been designed to exhibit five properties: (1) ability to handle non-convex/non-linearly separable data, (2) ability to estimate classification uncertainty, (3) environment agnostic parameter tuning, (4) supervision, and (5) online training. The robot interacts with areas selected via a sampling process that aims to reduce uncertainty in the classification and balancing of samples in each class. A change detector determines the class of the region with which the robot has interacted. The corresponding data are added to a dataset used to train the classifier. The approach generates a relevance map segmenting potential objects from the background. This information can be used to bootstrap an object discovery method, reducing the assumptions on the structure of the environment and thus paving the way to approaches that can adapt to a wider range of environments. The approach has been tested on setups of increasing complexity using simulations and a real PR2 robot.

\section*{Acknowledgements}
This work is supported by the DREAM project \endnote{\url{http://www.robotsthatdream.eu}} through the European Unions Horizon 2020 research and innovation program under grant agreement No 640891. This work has been partially sponsored by the French government research program Investissements d'avenir through the Robotex Equipment of Excellence (ANR-10-EQPX-44).

\theendnotes

\bibliographystyle{SageH}
\bibliography{bib.bib}

\end{document}